\documentclass[runningheads]{llncs}

\usepackage[year=2026]{eccv}

\usepackage{eccvabbrv}

\usepackage{graphicx}
\usepackage{booktabs}

\usepackage[accsupp]{axessibility}  
\usepackage{multirow}

\usepackage{hyperref}

\usepackage{orcidlink}

\usepackage{xcolor}

\usepackage{bm}
\newcommand{\secref}[1]{Sec.~\ref{#1}}

\begin{document}

\title{PAS3R: Pose-Adaptive Streaming 3D Reconstruction for Long Video Sequences} 

\titlerunning{PAS3R}

\author{Lanbo Xu\inst{1} \and
Liang Guo\inst{1} \and
Caigui Jiang\inst{1}$^{\dagger}$ \and
Cheng Wang\inst{2}$^{\dagger}$}

\authorrunning{L.~Xu et al.}

\institute{State Key Laboratory of Human-Machine Hybrid Augmented Intelligence, Institute of Artificial Intelligence and Robotics, Xi'an Jiaotong University, China \and
University of East Anglia, Norwich, NR47TJ, United Kingdom
\email{Cheng.C.Wang@uea.ac.uk}\\
\small $^{\dagger}$Corresponding authors}

\maketitle

\begin{abstract}
Online monocular 3D reconstruction enables dense scene recovery from streaming video but remains fundamentally limited by the stability–adaptation dilemma: the reconstruction model must rapidly incorporate novel viewpoints while preserving previously accumulated scene structure. Existing streaming approaches rely on uniform or attention-based update mechanisms that often fail to account for abrupt viewpoint transitions, leading to trajectory drift and geometric inconsistencies over long sequences.
We introduce PAS3R, a pose-adaptive streaming reconstruction framework that dynamically modulates state updates according to camera motion and scene structure. Our key insight is that frames contributing significant geometric novelty should exert stronger influence on the reconstruction state, while frames with minor viewpoint variation should prioritize preserving historical context. PAS3R operationalizes this principle through a motion-aware update mechanism that jointly leverages inter-frame pose variation and image frequency cues to estimate frame importance.
To further stabilize long-horizon reconstruction, we introduce trajectory-consistent training objectives that incorporate relative pose constraints and acceleration regularization. A lightweight online stabilization module further suppresses high-frequency trajectory jitter and geometric artifacts without increasing memory consumption.
Extensive experiments across multiple benchmarks demonstrate that PAS3R significantly improves trajectory accuracy, depth estimation, and point cloud reconstruction quality in long video sequences while maintaining competitive performance on shorter sequences. The source
code is available at \url{https://pas-3r.github.io/PAS3R.io/}.
\end{abstract}

\section{Introduction}
\label{sec:intro}

3D reconstruction is a technique that extracts 3D reconstruction aims to recover geometric scene representations from image observations and has become a fundamental problem in computer vision. These representations can be broadly categorized into explicit and implicit forms. Typical explicit representations include point clouds \cite{qi2017pointnet, qi2017pointnet++}, meshes \cite{wang2018pixel2mesh, groueix2018papier}, and voxels \cite{choy20163d, wu2016learning, liu2020dlgan}. Recently, with the continuous advancement of rendering technologies, novel explicit representations represented by 3D Gaussian Splatting (3DGS) \cite{kerbl20233d} have emerged \cite{wu20244d, lu2024scaffold, guo2024tetsphere}. In contrast, implicit methods primarily define a continuous function to map points in space \cite{mildenhall2020nerf, park2019deepsdf, muller2022instant, wang2021neus}. Among these, point cloud–based representations remain widely used due to their flexibility and compatibility with downstream geometric processing tasks.

Methods based on point cloud representation typically employ calibrated multi-view cameras to recover scene point clouds \cite{schonberger2016structure, yao2018mvsnet, barnes2009patchmatch}. However, monocular camera-captured scenes are ubiquitous in daily life, and in most cases, camera calibration parameters are unavailable. Consequently, numerous studies \cite{leroy2024grounding, teed2021droid, wang2024dust3r, wang2025continuous} have focused on this challenging scenario. 
These methods are primarily categorized into offline \cite{leroy2024grounding, wang2024dust3r, wang2025vggt, yang2025fast3r} and online \cite{teed2021droid, wang2025continuous, zhuo2025streaming, zhang2025efficiently} frameworks. Offline methods typically leverage attention mechanisms \cite{vaswani2017attention} to directly predict camera parameters from image sequences, supporting both multi-view and monocular scenarios. However, due to their batch-processing nature, they require loading all images simultaneously, causing memory consumption to scale with the sequence length and thus limiting their practical application. In contrast, online methods maintain a fixed-size state feature vector to achieve constant memory usage, but they inevitably introduce the issue of "catastrophic forgetting". Since online models rely on a single global state vector that must be updated with information from each new viewpoint, the proportion of historical information from early frames diminishes as the sequence progresses, leading to a loss of context.

Recent work such as TTT3R \cite{anonymous2026tttr} attempts to mitigate this issue by adapting the learning rate through cross-attention between historical and incoming features. While this mechanism partially alleviates forgetting, it does not explicitly consider the magnitude of viewpoint change between frames. In practice, we observe that when the camera undergoes abrupt pose changes, the update intensity may remain insufficient, preventing the model from quickly adapting to new scene geometry. Conversely, when pose variation is minimal, overly strong updates can unnecessarily overwrite historical information. This imbalance often manifests as trajectory drift and geometric inconsistencies in long sequences.

To address this issue, we propose PAS3R (Pose-Adaptive Streaming 3D Reconstruction), a framework designed to improve the stability of online monocular reconstruction over long video sequences. The central idea of PAS3R is that the influence of each incoming frame should be determined by the degree of geometric novelty it introduces. Frames associated with substantial viewpoint change should contribute more strongly to the reconstruction state in order to capture new scene information, while frames with minimal pose variation should prioritize preserving previously accumulated geometry.
To operationalize this principle, PAS3R introduces a pose-adaptive state update mechanism that dynamically modulates update intensity according to inter-frame camera motion and scene frequency characteristics. This mechanism enables the model to rapidly adapt to novel viewpoints while maintaining long-term geometric consistency.
In addition, we introduce trajectory-consistent training objectives that incorporate relative pose constraints and acceleration regularization to improve temporal stability during reconstruction. Finally, a lightweight online stabilization module reduces residual trajectory jitter and geometric artifacts while preserving the efficiency required for streaming reconstruction. 
Compared with state-of-the-art (SOTA) online 3D reconstruction models, PAS3R demonstrates superior performance in long-sequence modeling while remaining highly competitive on short sequences.

Overall, the contributions of this work are summarized below:
\begin{itemize}
    \item We introduce PAS3R, a streaming 3D reconstruction framework that dynamically regulates state updates based on camera motion and scene structure, enabling stable long-horizon reconstruction from monocular video streams.
    \item We propose a pose-adaptive update mechanism that estimates frame importance using inter-frame camera displacement and image frequency cues, allowing the model to balance rapid adaptation to novel viewpoints with preservation of historical geometry.
    \item We introduce relative pose and acceleration regularization into the training process, improving temporal coherence and reducing trajectory drift in long video sequences.
    \item We design an efficient spatiotemporal stabilization strategy that suppresses trajectory jitter and geometric artifacts without increasing memory overhead.
\end{itemize}

\section{Related Work}

\subsection{Traditional 3D Reconstruction Methods}
Early research in 3D reconstruction primarily relied on multi-view geometry theory \cite{hartley2003multiple, faugeras1993three, szeliski2022computer}, where Structure-from-Motion (SfM) \cite{schonberger2016structure} and Multi-View Stereo (MVS) \cite{yao2018mvsnet} served as the cornerstones for offline reconstruction. Traditional pipelines, represented by COLMAP \cite{schonberger2016pixelwise, schonberger2016vote}, recover high-precision camera poses and dense point clouds from unordered image sets through feature extraction \cite{lowe2004distinctive, rublee2011orb, detone2018superpoint, sarlin2020superglue} and global Bundle Adjustment (BA) \cite{triggs1999bundle, agarwal2010bundle}. However, this paradigm's reliance on global optimization typically requires simultaneous access to all historical frames, with computational complexity growing cubically with the number of images, making it difficult to meet the real-time requirements of online processing. To address this, visual SLAM systems \cite{mur2017orb, engel2017direct, newcombe2011dtam, NEURIPS2023, cadena2017past, strasdat2011double, zhu2022nice, forster2016svo, brachmann2017dsac} introduce keyframe mechanisms and local optimization strategies. However, these methods still rely on SfM-style optimization within sliding windows, causing memory consumption and computational cost to increase with sequence length. In addition, their performance often degrades in low-texture environments or under rapid camera motion, where accumulated errors can significantly affect reconstruction quality. In contrast, our work focuses on enabling stable long-sequence reconstruction under constant memory constraints, which is essential for streaming reconstruction scenarios.

\subsection{End-to-End 3D Reconstruction Methods}
In contrast to traditional methods that decouple feature tracking, back-end optimization, and mapping into independent modules, end-to-end approaches seek to build a unified differentiable framework that directly maps image sequences to 3D geometric representations. As a pioneering model, DUSt3R \cite{wang2024dust3r} introduced a radically novel paradigm based on Transformers for dense and unconstrained stereo 3D reconstruction of arbitrary image collections, operating without prior camera calibration or viewpoint poses. However, DUSt3R was specifically designed for image pairs and lacks support for multi-view inputs. Although several follow-up works \cite{zhang2024monst3r, chen2025easi3r, cabon2025must3r, wang20243d} have extended it to broader scenarios, its fundamental limitations remain. To overcome this, VGGT \cite{wang2025vggt} and Fast3r \cite{yang2025fast3r} adopted feed-forward Transformer architectures with global attention, allowing for multi-view input in a single pass. Nevertheless, the high computational cost of global attention and the inability to handle incremental frames limit their application in online reconstruction. 

As an alternative, CUT3R \cite{wang2025continuous} utilizes recurrent networks to update memory states and employs local attention to extend end-to-end 3D reconstruction to the online domain. In addition, contemporary works \cite{wu2025point3r, shen2025mut3r} also achieves successful online 3D reconstruction by employing a similar memory-update approach. More recently, TTT3R \cite{anonymous2026tttr} introduced cross-attention learning rates \cite{su2024roformer} and state resets from the perspective of Test-Time Training (TTT) \cite{sun2020test, behrouz2024titans, gu2024mamba}, while InfiniteVGGT (IVGGT) \cite{yuan2026infinitevggt} utilized adaptive the key-value cache (KV-cache) \cite{zhang2023h2o} to further extend end-to-end online reconstruction to long sequences with promising results. Despite these advances, existing online reconstruction approaches still struggle to balance rapid adaptation to new viewpoints with preservation of previously accumulated geometry. In particular, update mechanisms based primarily on attention or feature similarity do not explicitly account for the magnitude of viewpoint change between frames. As a result, large pose transitions may not sufficiently influence state updates, while small viewpoint changes may unnecessarily overwrite historical information. Our work addresses this limitation by introducing pose-adaptive state modulation, which explicitly incorporates camera motion and image structural cues when determining the influence of each incoming frame.

\subsection{Spatial Geometric Consistency}

High-quality 3D reconstruction depends not only on accurate camera trajectory estimation but also on the spatial consistency \cite{weinzaepfel2022croco, Yin_2019_ICCV} and edge sharpness \cite{bochkovskii2410depth, kopf2021robust, wang2023neural} of dense geometric representations. Although end-to-end models can directly regress dense point clouds, the absence of explicit geometric constraints often leads to high-frequency artifacts, particularly near depth-discontinuous object boundaries and planar surfaces.
Traditional smoothing techniques \cite{gedraite2011investigation, perona2002scale, rudin1992nonlinear} can reduce noise but often blur geometric edges and degrade structural detail. To address this issue, edge-preserving filtering methods such as bilateral filtering \cite{tomasi1998bilateral, paris2006fast, digne2017bilateral, fleishman2003bilateral, barron2016fast, zhang2021bilateral} have been widely used in geometry processing. Inspired by these approaches, we incorporate an online bilateral spatial filtering strategy to refine reconstructed point clouds during inference. This lightweight refinement step suppresses artifacts in planar regions while preserving sharp object boundaries, improving geometric consistency without increasing model complexity.

\begin{figure}[t]
    \centering
    \includegraphics[width=\linewidth]{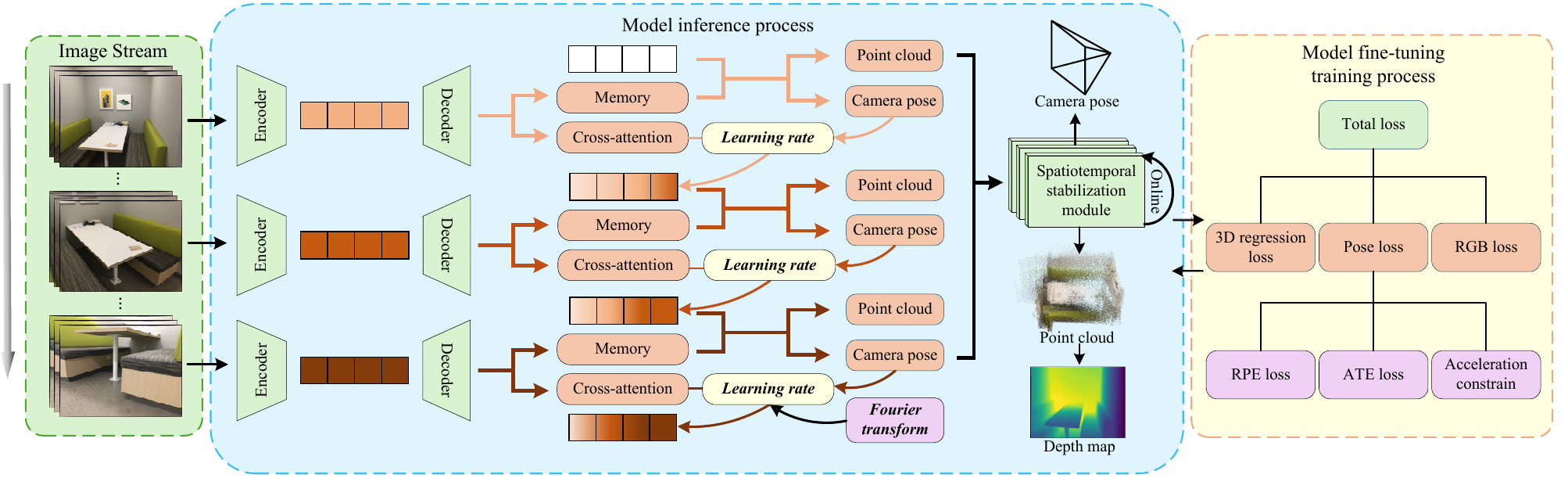}
    \vspace{-3mm}
    \caption{The overall pipeline. Given a streaming image sequence, the model maintains a persistent memory that is iteratively updated through cross-attention with incoming frames. PAS3R introduces pose-adaptive state update modulation, where the update intensity (learning rate) is dynamically adjusted based on camera motion and image structure (via Fourier analysis), allowing the model to balance adaptation to novel viewpoints and preservation of accumulated geometry. During inference, the model predicts camera poses and dense point clouds at each step. During training, we further enforce trajectory-consistent optimization using a novel pose loss with ATE, RPE, and acceleration constraints. A lightweight online spatiotemporal stabilization module is applied to reduce trajectory jitter and geometric artifacts. Together, these components enable stable and scalable long-horizon streaming 3D reconstruction.}
    \label{pipeline}
    \vspace{-5mm}
\end{figure}

\section{Approach}
This work focuses on improving the stability and reconstruction fidelity of online monocular 3D reconstruction over long video sequences. Streaming reconstruction models maintain a persistent internal state that must be updated sequentially as new frames arrive. The effectiveness of this paradigm critically depends on how strongly each incoming frame influences the state update. PAS3R addresses this challenge through \textit{pose-adaptive state modulation}, which dynamically regulates the update intensity of the reconstruction state according to camera motion and scene characteristics.
As illustrated in \cref{pipeline}, the proposed method consists of three components. First, pose-adaptive state update modulation (Section~\ref{ssec:pose-adaptive}); 
Second, trajectory-consistent model optimization (Section~\ref{ssec:fine-tune}); 
Third, lightweight online stabilization module (Section~\ref{ssec:st-stabilization}). 
Together these components enable stable long-horizon reconstruction while preserving the efficiency required for streaming inference.

\subsection{Pose-Adaptive State Update Modulation} \label{ssec:pose-adaptive}
Streaming reconstruction models maintain an internal state that encodes accumulated scene information. During sequential processing, this state must be updated as new frames arrive. In the Test-Time Training (TTT)\cite{sun2024learning} paradigm, this state can be interpreted as a set of fast weights that are updated dynamically while the model parameters
(i.e., slow weights) fixed during inference. Formally, let the internal state at time step $t-1$ be represented as $S_{t-1}\in R^{n\times c}$. When a new observation $X_t$ arrives, the state update can be expressed as in Eq. \eqref{state update}:
\begin{equation}
\label{state update}
    \mathrm{Update}(S_{t-1},X_t)=S_{t-1}-\beta\nabla (S_{t-1},X_t)
\end{equation}
where $\nabla(S_{t-1}, X_t)$ is a learned gradient function represents the association between the previous state $S_{t-1}$ and the current observation $X_t$, with $\beta$ controlling the update magnitude (learning rate). 
Intuitively, this online learning process encodes KV-cache from the current observation into a fixed-length memory the state as accurately as possible.

\begin{figure}[tb]
    \captionsetup[subfigure]{justification=centering}
    \centering
    \begin{subfigure}[b]{0.24\textwidth}
        \centering
        \includegraphics[width=\textwidth]{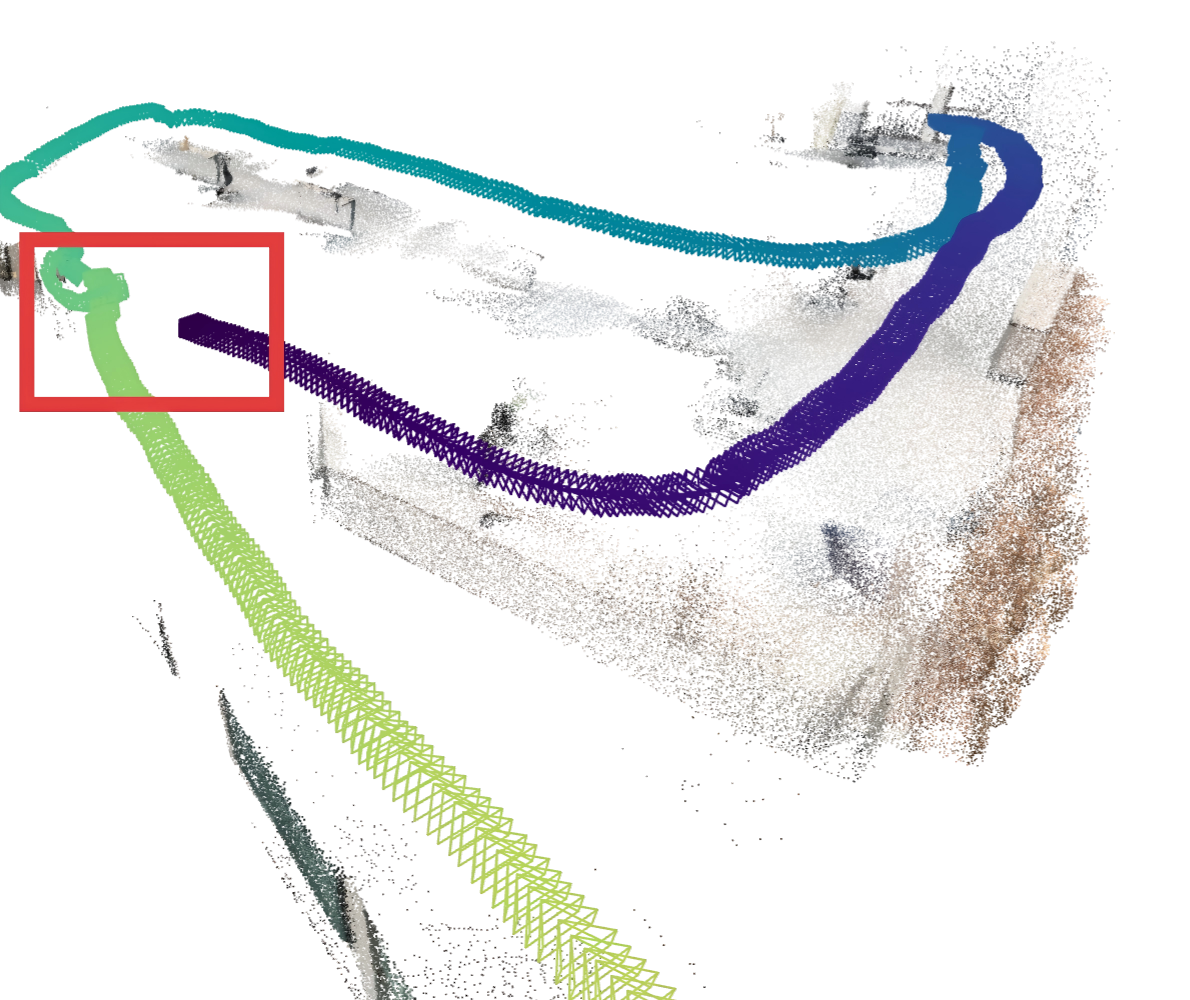}
        \caption{CUT3R}
        \label{weight-a}
    \end{subfigure}
    \hfill
    \begin{subfigure}[b]{0.24\textwidth}
        \centering
        \includegraphics[width=\textwidth]{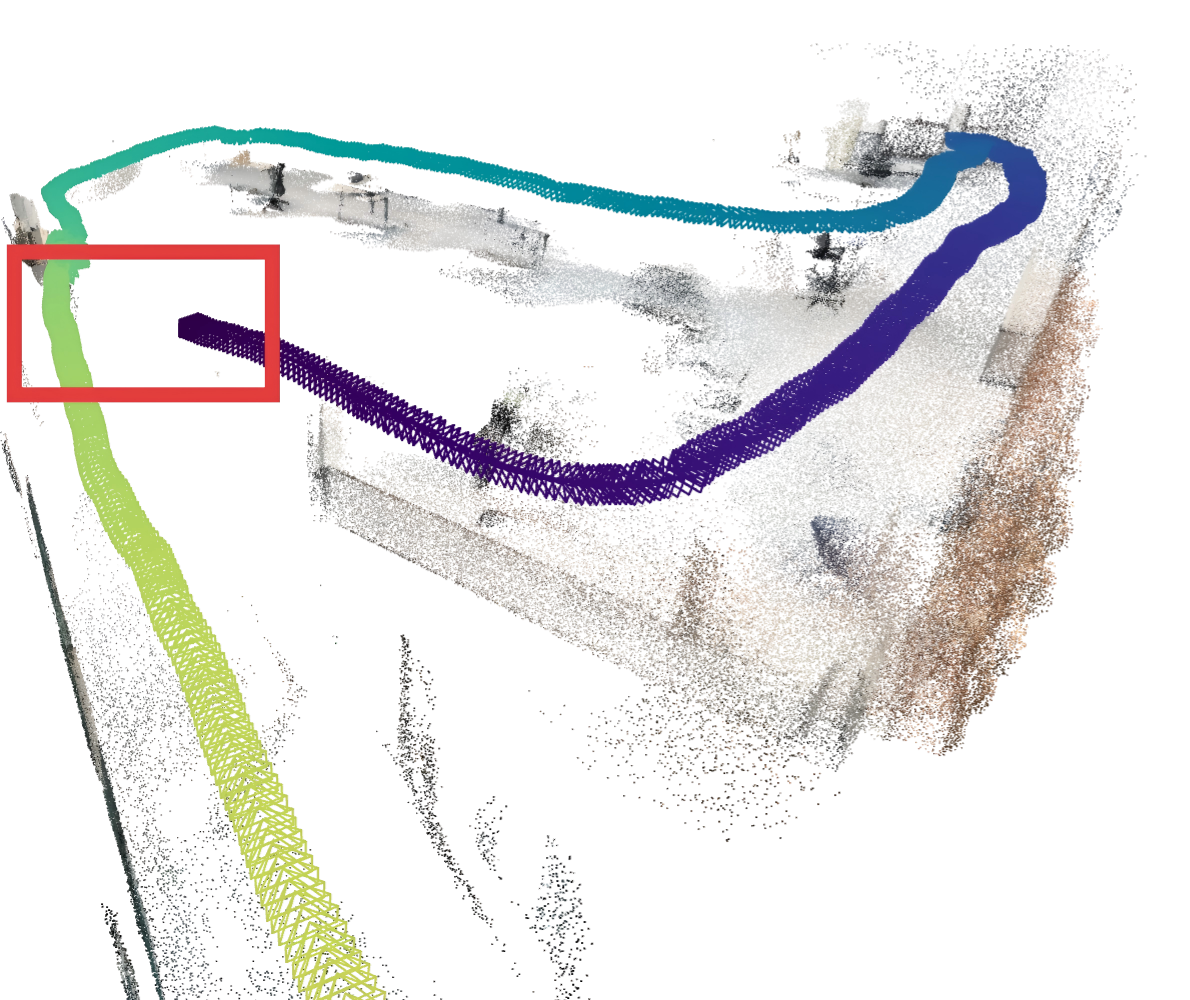}
        \caption{TTT3R}
        \label{weight-b}
    \end{subfigure}
    \begin{subfigure}[b]{0.24\textwidth}
        \centering
        \includegraphics[width=\textwidth]{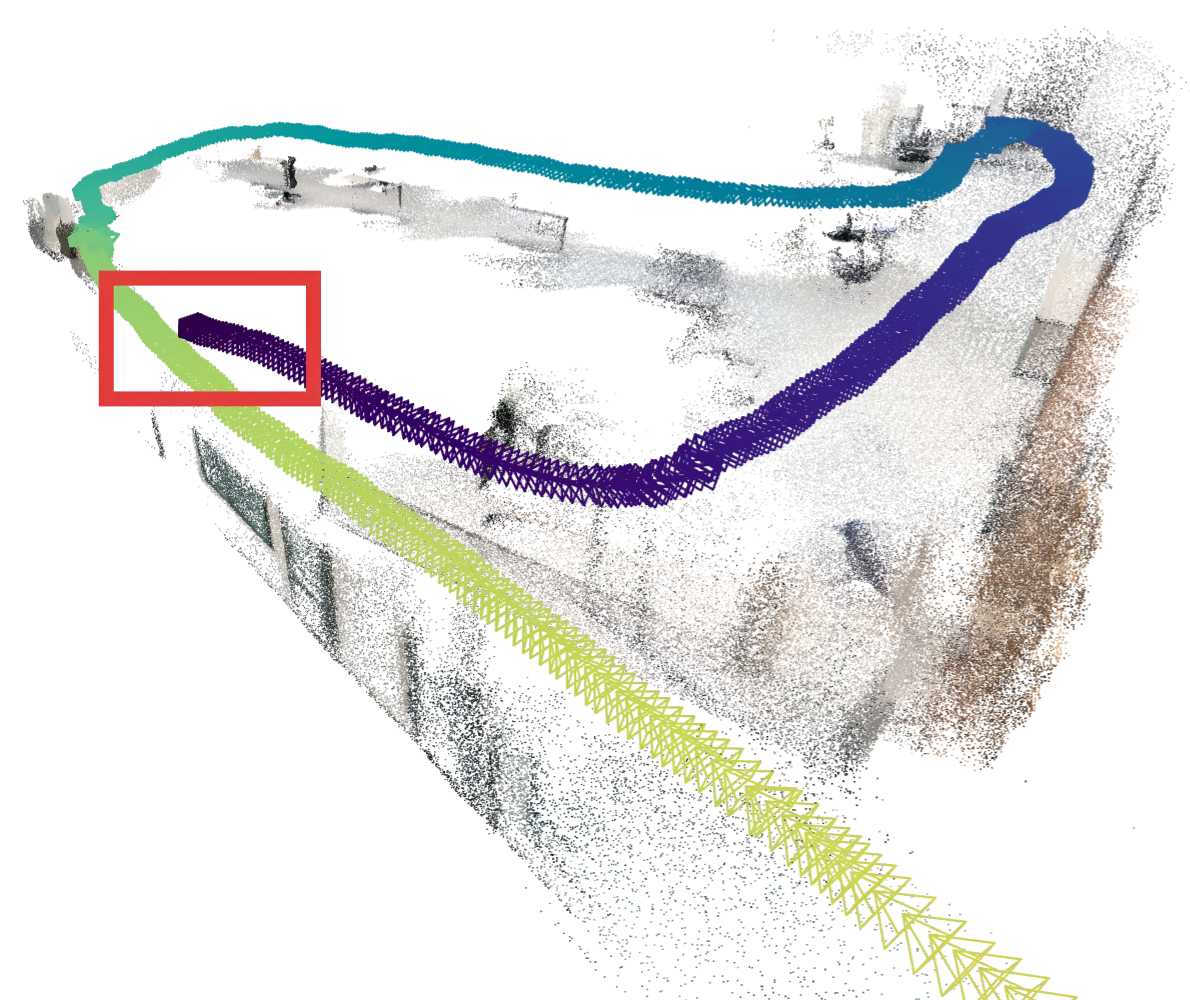}
        \caption{TTTWL}
        \label{weight-c}
    \end{subfigure}
    \hfill
    \begin{subfigure}[b]{0.24\textwidth}
        \centering
        \includegraphics[width=\textwidth]{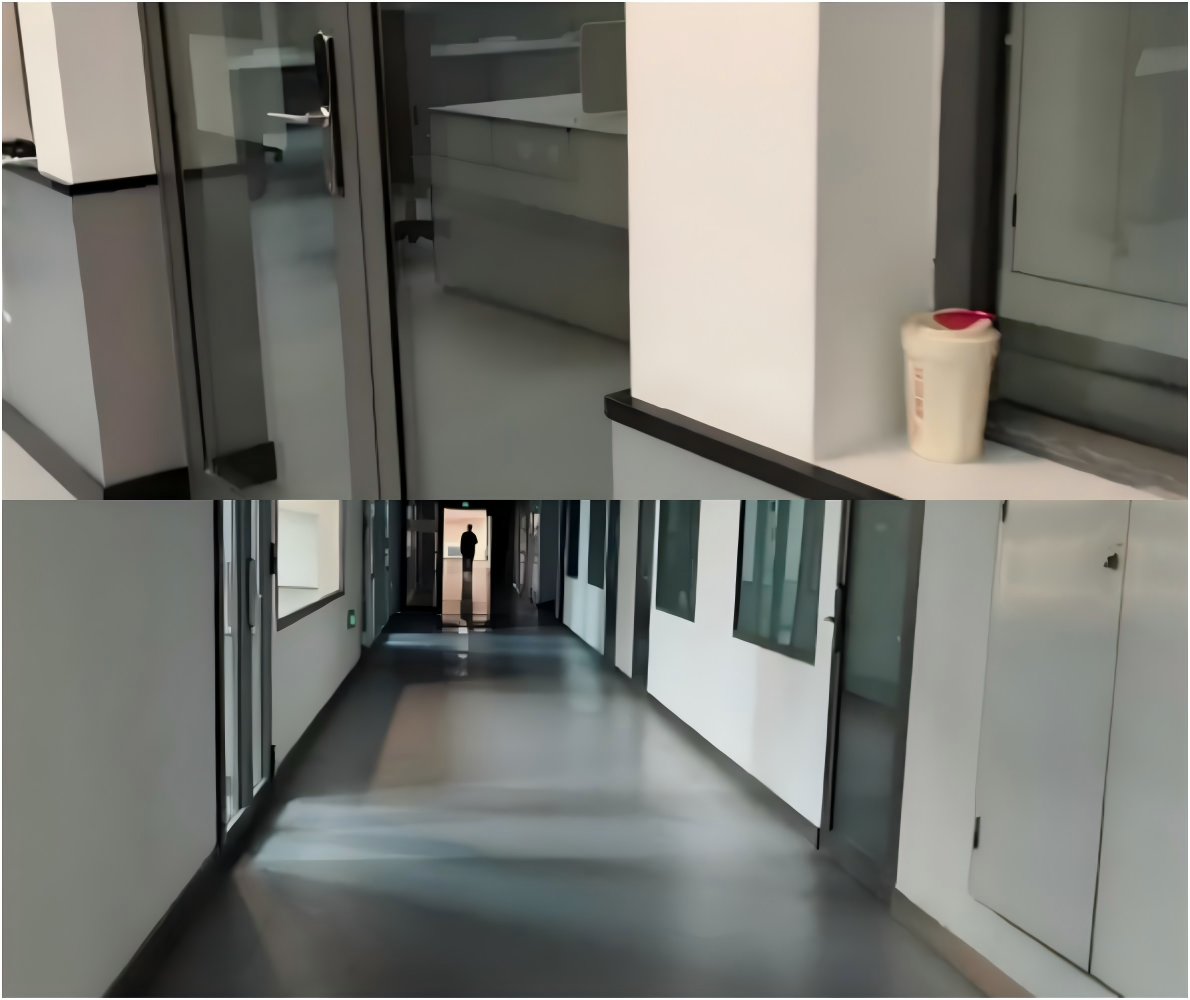}
        \caption{Real view}
        \label{weight-d}
    \end{subfigure}
    \caption{\textbf{Effect of adaptive state update weighting.} We evaluate the impact of adaptive update weighting in a real-world indoor video sequence. The red bounding box highlights a keyframe region where the camera undergoes significant pose variation. Ideally, the trajectory following this region should remain approximately parallel. However, both CUT3R \cite{wang2025continuous} and TTT3R \cite{anonymous2026tttr} exhibit noticeable drift from the real trajectory, while the adaptive weighting strategy (TTTWL denotes PAS3R with adaptive update only) helps maintain trajectory continuity under abrupt viewpoint changes.}
    \label{weight}
    \vspace{-5mm}
\end{figure}

In conventional streaming reconstruction frameworks, the update magnitude is typically fixed or determined by internal attention mechanisms. However, such approaches do not explicitly account for the degree of geometric novelty introduced by each frame. As a result, frames with significant viewpoint changes may not sufficiently influence the state update, while frames with minimal motion may unnecessarily overwrite historical information. An example is shown in \cref{weight}.
To address this limitation, we introduce pose-adaptive state update modulation, which dynamically adjusts the update magnitude based on two complementary cues:
Inter-frame camera motion, which reflects viewpoint change;
Image structural richness, which indicates the availability of geometric information.

Inter-frame camera motion encompasses both translation and rotation. An increase in translation distance indicates higher camera velocity, while an increase in rotation angle signifies a trajectory inflection point or abrupt rotational jitter. Therefore, we calculate the weighted sum of the two to obtain the result shown in Eq. \eqref{new rot weight}:
\begin{equation}
\label{new rot weight}
s_1 = w_1 \Delta x + w_2 \Delta q
\end{equation}
\noindent where $\Delta x$ represents the magnitude of translation displacement and $\Delta q$ denotes the magnitude of rotation change, both of which are positive values; $w_1$ and $w_2$ are their respective weights, and $s_1$ serves as the comprehensive score for the camera pose displacement. 
 
Furthermore, we employ the Fourier transform to conduct a more rigorous evaluation of the current frame's image quality. To minimize computational overhead and eliminate the interference of color information on structural features, the current frame is first converted to a grayscale image $I_{gray}$. A Discrete Fourier Transform (DFT) is then performed on $I_{gray}$ and centered as follows in Eq. \eqref{DFT}:
\begin{equation}
\label{DFT}
F(u, v) = \text{shift}(\|\mathcal{F}\{I_{gray}(x, y)\}\|)
\end{equation}
\noindent where $\mathcal{F}\{I_{gray}(x, y)\}$ denotes the 2D Discrete Fourier Transform of the grayscale image. The $\text{shift}$ function is responsible for moving the zero-frequency low-frequency components—representing the overall smooth brightness and background—to the geometric center $(cx, cy)$ of the matrix. $F(u, v)$ represents the resulting centered magnitude spectrum matrix. To filter out low-frequency signals and preserve high-frequency information, we construct a high-pass filter matrix $M(u, v)$, whose mathematical form is defined in Eq. \eqref{high-pass filter matrix}:
\begin{equation}
\label{high-pass filter matrix}
M(u, v) = \begin{cases} 1, & u^2 + v^2 > r^2 \\ 0, & \text{otherwise} \end{cases}
\end{equation}
\noindent where $r$ is the radius of a circle centered at the origin of the coordinate axes; signals outside this radius are categorized as high-frequency components and are retained. Consequently, we can determine the proportion of high-frequency signals $R$ and calculate the final image quality score $s_2$, as shown in Eq. \eqref{high-frequency signals} and Eq. \eqref{quality score}:
\begin{equation}
\label{high-frequency signals}
R = \frac{\sum_{u, v} [F(u, v) \cdot M(u, v)]}{\sum_{u, v} F(u, v) + \epsilon}
\end{equation}
\begin{equation}
\label{quality score}
s_2 = \frac{1}{1 + e^{-20.0 \times (R - 0.1)}}
\end{equation}
\noindent By integrating the camera pose displacement score $s_1$ and the image quality score $s_2$, we derive the total score $s$ for the current frame according to Eq.\eqref{total score}:
\begin{equation}
\label{total score}
s = s_1 \cdot s_2
\end{equation}
\noindent To prevent excessive weight magnitude, we clip the maximum value of $s$ at 1.0. This restricted total score is then utilized as the final learning rate weight. Through this adaptive learning rate weighting strategy, our method maintains trajectory continuity during smooth camera motion while remaining responsive to abrupt camera jitter and extends the model's reconstruction and prediction performance across extended sequences.

\subsection{Trajectory-Consistent Model Optimization} \label{ssec:fine-tune}

\begin{figure}[t]
    \captionsetup[subfigure]{justification=centering}
    \centering
    \begin{subfigure}[b]{0.32\textwidth}
        \centering
        \includegraphics[width=\textwidth]{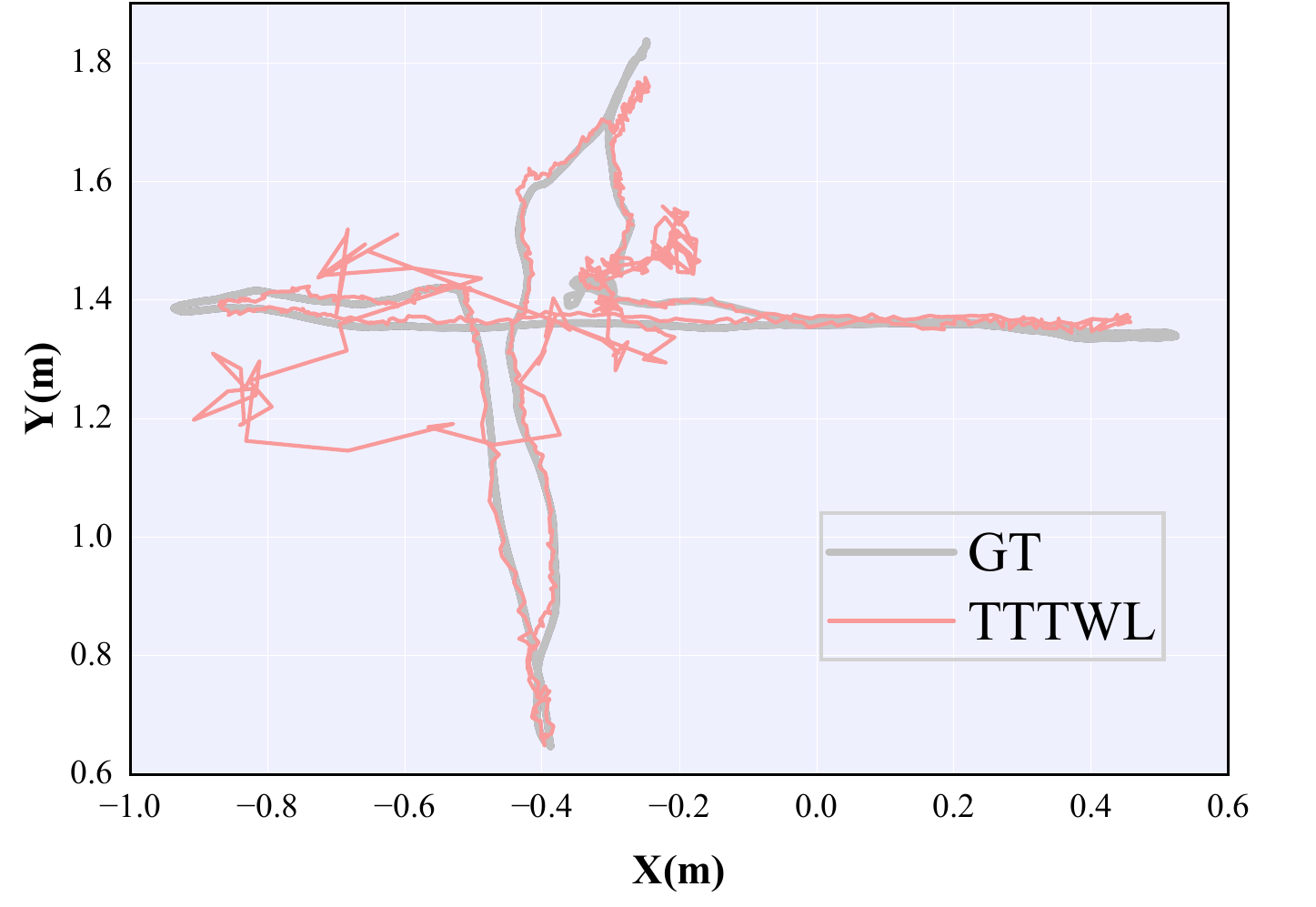}
    \end{subfigure}
    \hspace{2pt}
    \begin{subfigure}[b]{0.32\textwidth}
        \centering
        \includegraphics[width=\textwidth]{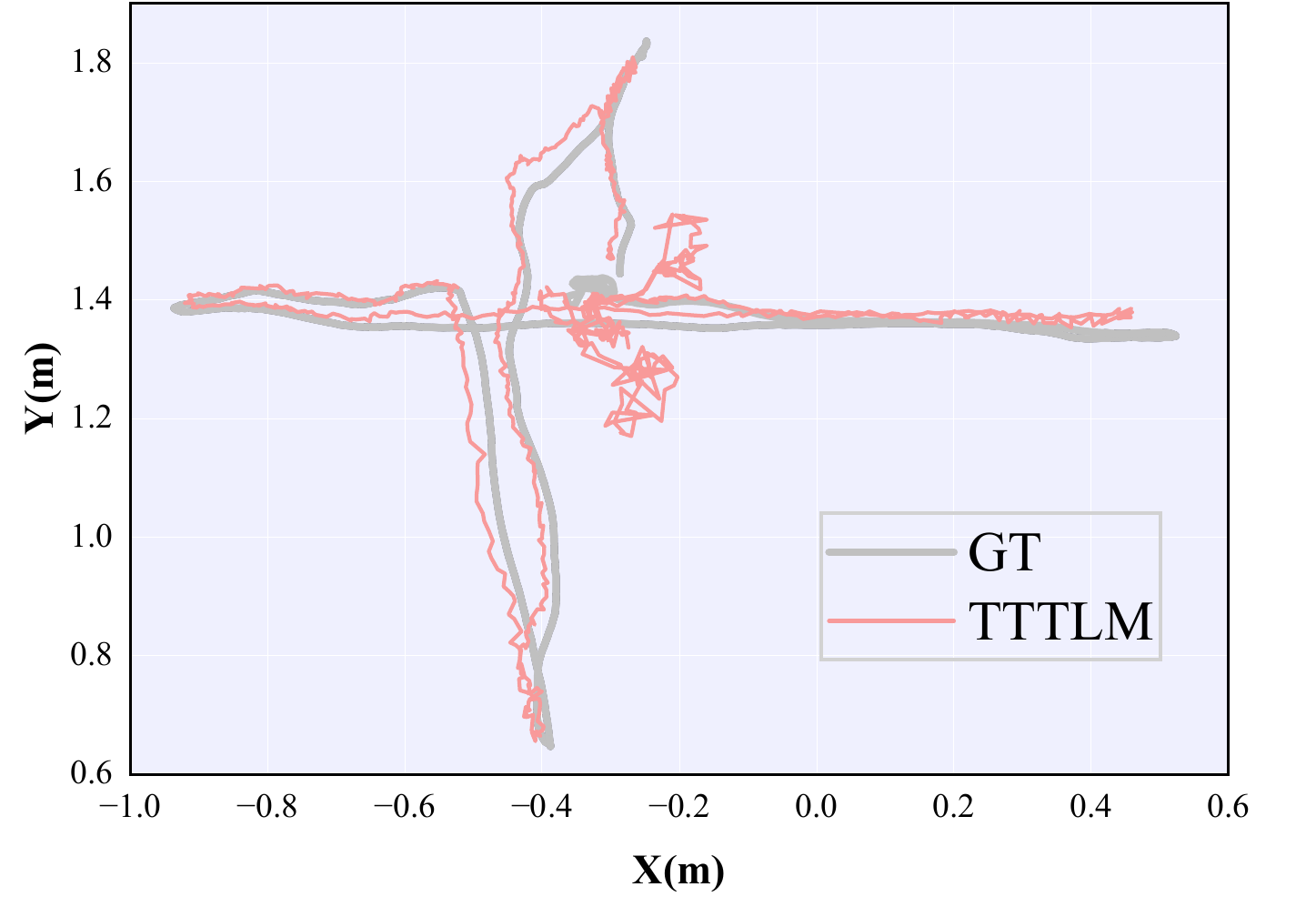}
    \end{subfigure}
    \caption{\textbf{Effect of trajectory-consistent training.} When large camera motion occurs after relatively stable frames, adaptive updates alone may produce abrupt trajectory variations. The proposed loss formulation reduces these discontinuities and produces smoother and more physically consistent camera trajectories. TTTWL denotes PAS3R with adaptive update only (Section~\ref{ssec:pose-adaptive}), while TTTLM denotes PAS3R with adaptive update and trajectory-consistent training (Section~\ref{ssec:fine-tune}).}
    \label{tuning}
    \vspace{-5mm}
\end{figure}

While the pose-adaptive state update mechanism enhances overall trajectory continuity and local variability, a potential issue arises if the model maintains a small learning rate weight when the preceding video sequence is relatively stable; if the camera then undergoes a sudden, large-scale displacement, the model will assign a significantly higher weight. Although this allows the model to track the camera's motion, it can manifest as a sharp, unnatural mutation in the trajectory (an occurrence that is physically improbable in real-world filming scenarios), as shown in \cref{tuning}.

To address this phenomenon, we introduce a trajectory-consistent training objective that regularizes the camera motion predicted by the model. The key idea is to explicitly enforce both local pose consistency and physically plausible motion dynamics during training, enabling the model to produce more stable trajectories when combined with the adaptive state updates of Section~\ref{ssec:pose-adaptive}.

Our framework is compatible with streaming reconstruction architectures \cite{wang2025continuous, anonymous2026tttr}, and the total loss consists of three components: the confidence-aware regression loss $\mathcal{L}_{conf}$, the camera pose loss $\mathcal{L}_{pose}$, and the RGB loss $\mathcal{L}_{rgb}$. 

$\mathcal{L}_{conf}$ and  $\mathcal{L}_{rgb}$ follow the definitions used in CUT3R \cite{wang2025continuous} and are written as:
\begin{equation}
\label{3D regression loss}
\mathcal{L}_{conf} = \sum_{(\bm{\hat{p}}, c) \in (\bm{\hat{I}_p}, C)} \left( c \cdot \left\| \frac{\bm{\hat{p}}}{\hat{s}} - \frac{\bm{p}}{s} \right\|_2 - \alpha \log c \right)
\end{equation}
\begin{equation}
\label{RGB loss}
\mathcal{L}_{rgb} = \left\| \bm{\hat{I}_r} - \bm{I_r} \right\|_2^2
\end{equation}
\noindent where $\hat{s}$ and $s$ represent the scale normalization factors for the predicted point cloud $\bm{\hat{I}_p}$ and the ground truth point cloud $\bm{I_p}$, respectively. $\bm{I_r}$ denotes the ground truth RGB image pixels, while $\bm{\hat{I}_r}$ denotes the model predicted RGB pixels.

\textbf{Pose-adaptive Loss.} We introduce an enhanced pose loss that combines global trajectory accuracy with local motion consistency. Specifically, the pose loss consists of three complementary components: The first is the Absolute Trajectory Error (ATE), defined as follows in Eq.\eqref{ATE loss}:
\begin{equation}
\label{ATE loss}
\mathcal{L}_{ATE} = \frac{1}{N} \sum_{t=1}^{N} \left( \left\| \frac{\bm{\hat{x}_t}}{\hat{s}} - \frac{\bm{x_t}}{s} \right\|_2 + (1 - |\bm{\hat{q}_t} \cdot \bm{q_t}|) \right)
\end{equation}
\noindent where $\bm{\hat{x}_t}$ represents the predicted translation vector, $\bm{x_t}$ represents the ground truth translation vector, $\bm{\hat{q}_t}$ is the predicted rotation quaternion, and $\bm{q_t}$ is the ground truth rotation quaternion. The second component is the Relative Pose Error (RPE), whose mathematical expression is defined in Eq.\eqref{RPE loss}:
\begin{equation}
\label{RPE loss}
\mathcal{L}_{RPE} = \frac{1}{N-1} \sum_{t=2}^{N} \left( \left\| \Delta \bm{\hat{x}_t} - \Delta \bm{x_t} \right\|_2 + \left\| \Delta \bm{\hat{q}_t} - \Delta \bm{q_t} \right\|_2 \right)
\end{equation}
\noindent here, $\Delta$ denotes the difference between the current frame and the preceding frame; thus, the solution can only be computed when $N > 1$. The final term is the acceleration stabilization constraint, which penalizes trajectory acceleration to reduce jitter, as defined in Eq.\eqref{smooth loss}:
\begin{equation}
\label{smooth loss}
\mathcal{L}_{acc} = \frac{1}{N-2} \sum_{t=3}^{N} \left( \left\| \Delta^2 \bm{\hat{x}_t} \right\|_2 + \left\| \Delta^2 \bm{\hat{q}_t} \right\|_2 \right)
\end{equation}
\noindent The final Pose-loss is a weighted sum of these three terms, as shown in Eq.\eqref{pose loss}:
\begin{equation}
\label{pose loss}
\mathcal{L}_{pose} = w_a \cdot \mathcal{L}_{ATE} + w_r \cdot \mathcal{L}_{RPE} + w_s \cdot \mathcal{L}_{acc}
\end{equation}
\noindent The Total loss is the weighted sum of the 3D regression loss, RGB loss, and Pose-loss, as defined in Eq.\eqref{sum loss}:
\begin{equation}
\label{sum loss}
\mathcal{L} = \lambda_1 \cdot \mathcal{L}_{conf} + \lambda_2 \cdot \mathcal{L}_{rgb} + \lambda_3 \cdot \mathcal{L}_{pose}
\end{equation}

\subsection{Online Spatiotemporal Stabilization} \label{ssec:st-stabilization}

\begin{figure}[t]
    \captionsetup[subfigure]{justification=centering}
    \centering
    \begin{subfigure}[b]{0.32\textwidth}
        \centering
        \includegraphics[width=\textwidth]{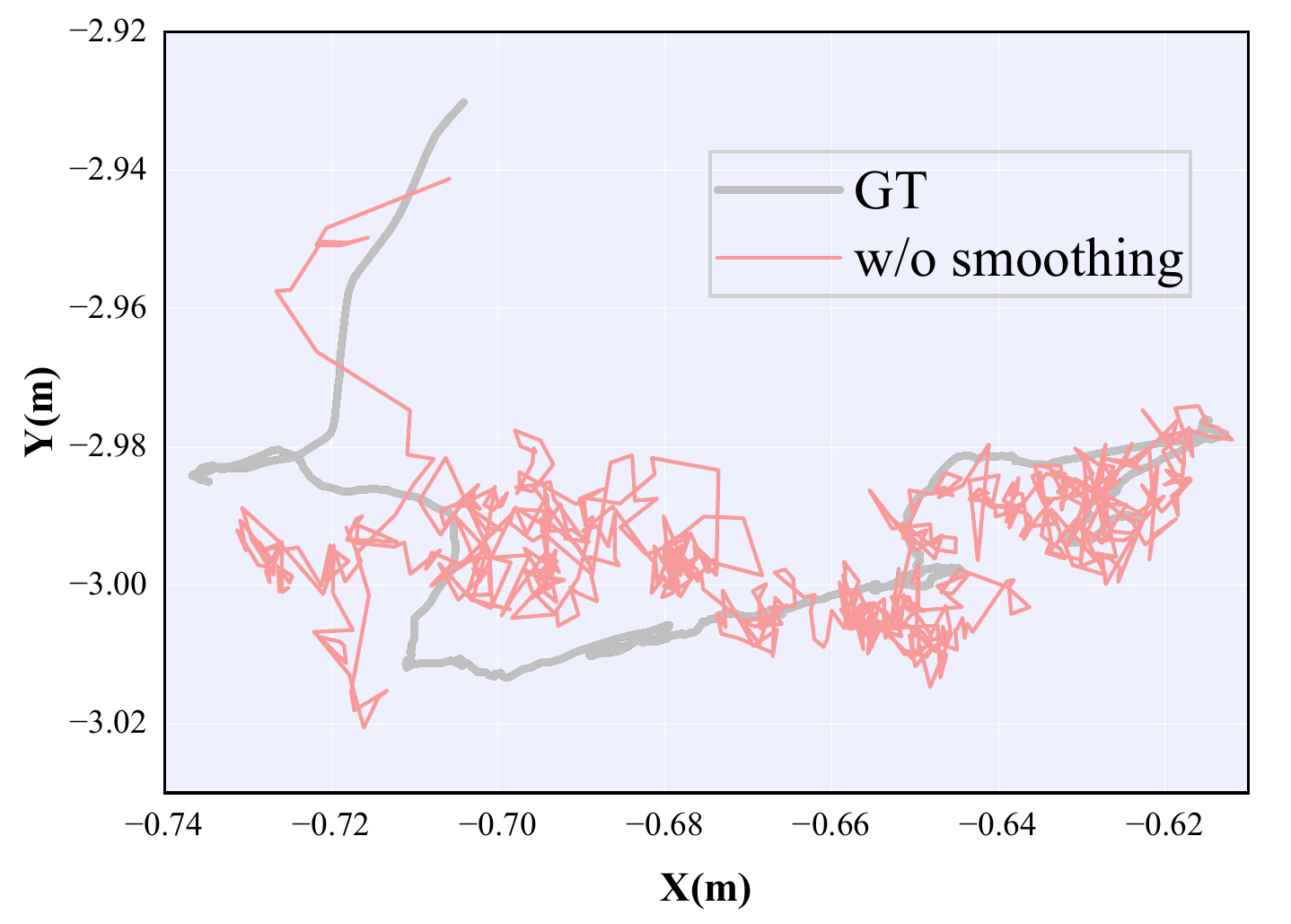}
        \caption{w/o temporal stabilization}
        \label{smoothing-a}
    \end{subfigure}
    \hspace{2pt}
    \begin{subfigure}[b]{0.32\textwidth}
        \centering
        \includegraphics[width=\textwidth]{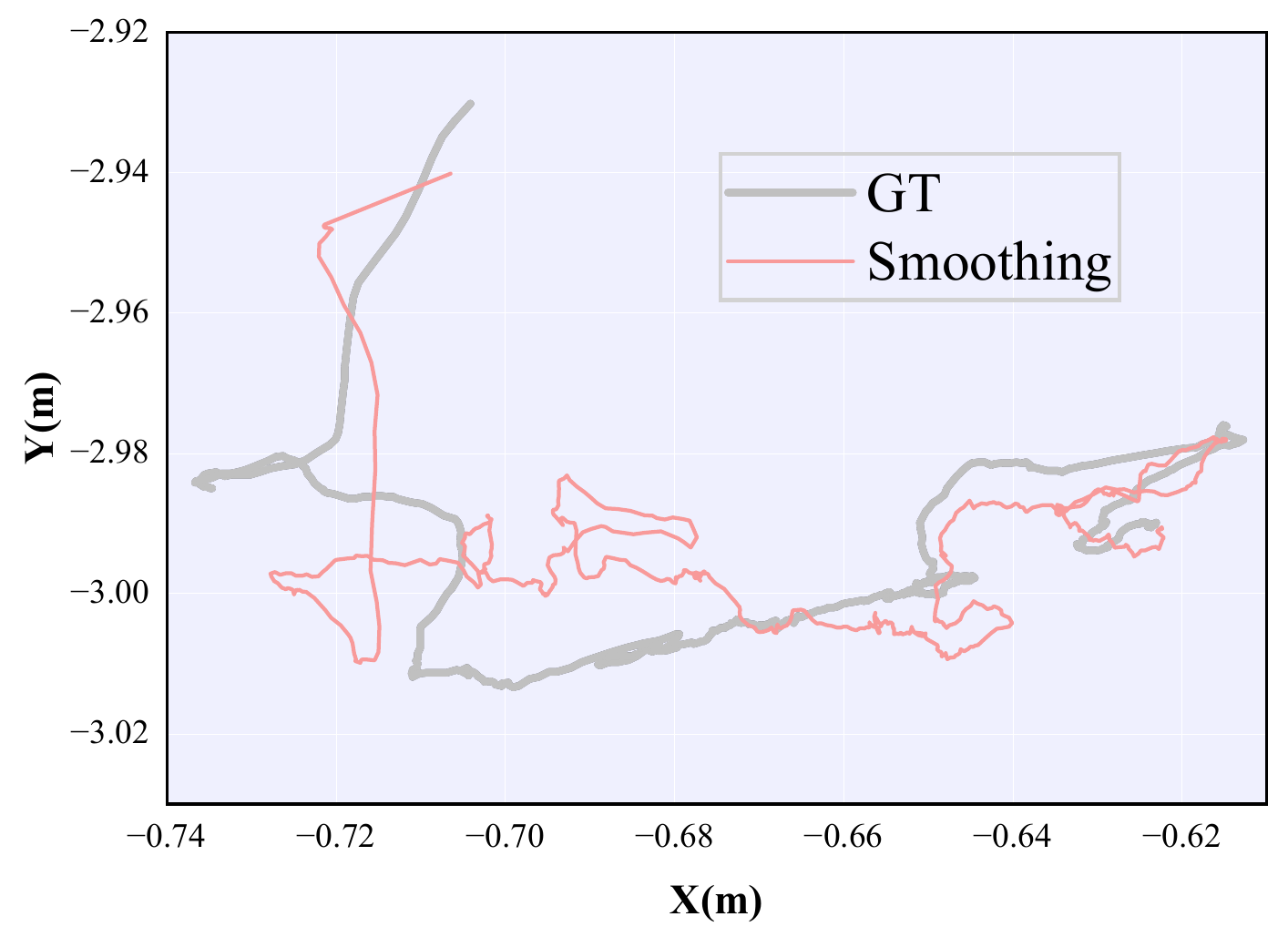}
        \caption{Full method}
        \label{smoothing-b}
    \end{subfigure}
    \caption{Effect of online trajectory stabilization.}
    \label{smoothing}
    \vspace{-5mm}
\end{figure}

\begin{figure}[t]
  \captionsetup[subfigure]{justification=centering}
  \centering
  \begin{subfigure}{0.32\linewidth}
    \centering
    \includegraphics[width=\linewidth]{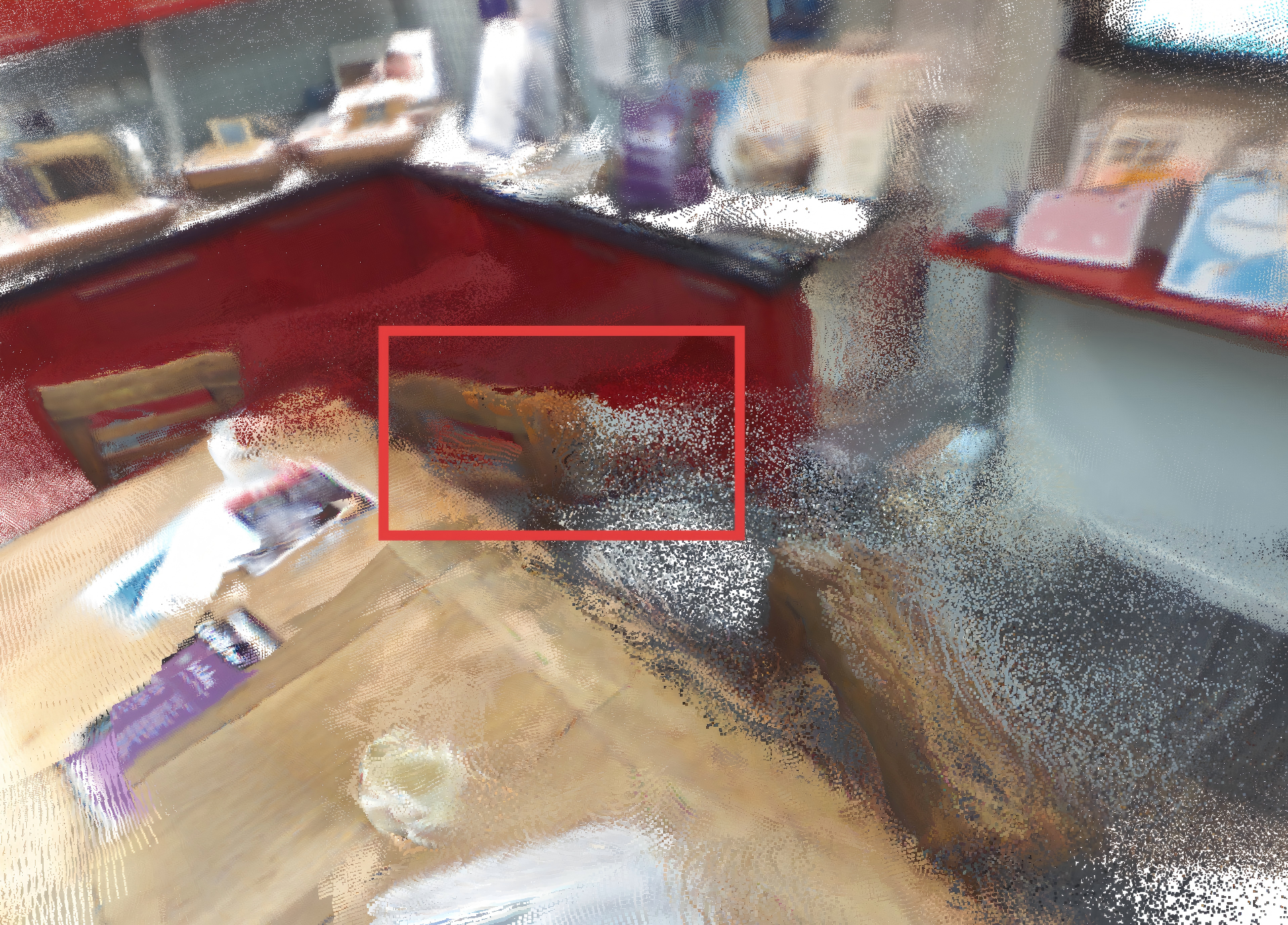}
    \caption{w/o spatial refinement}
    \label{bilateral-a}
  \end{subfigure}
  \hfill 
  \begin{subfigure}{0.32\linewidth}
    \centering
    \includegraphics[width=\linewidth]{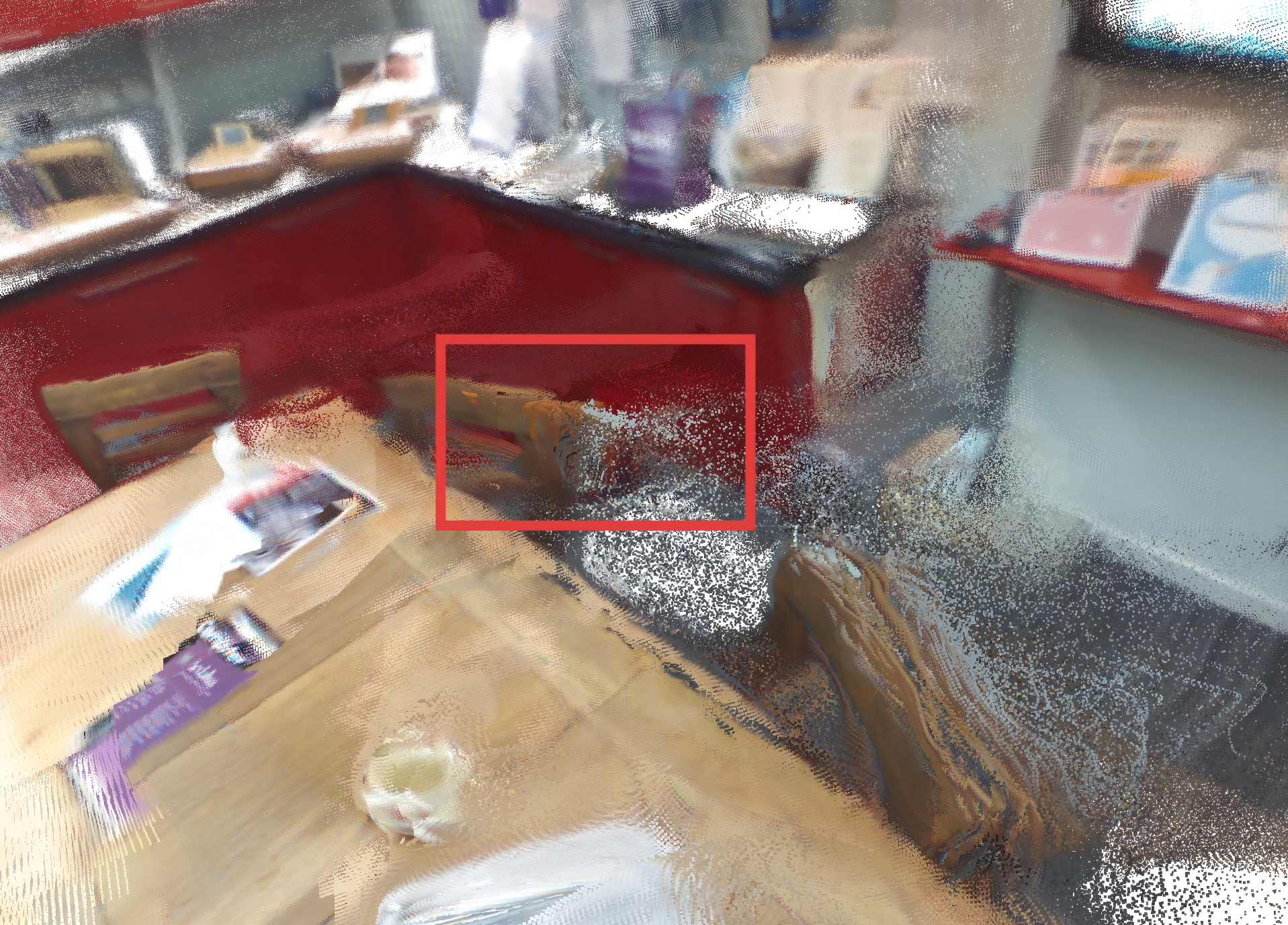}
    \caption{Full method}
    \label{bilateral-b}
  \end{subfigure}
  \hfill 
  \begin{subfigure}{0.32\linewidth}
    \centering
    \includegraphics[width=\linewidth]{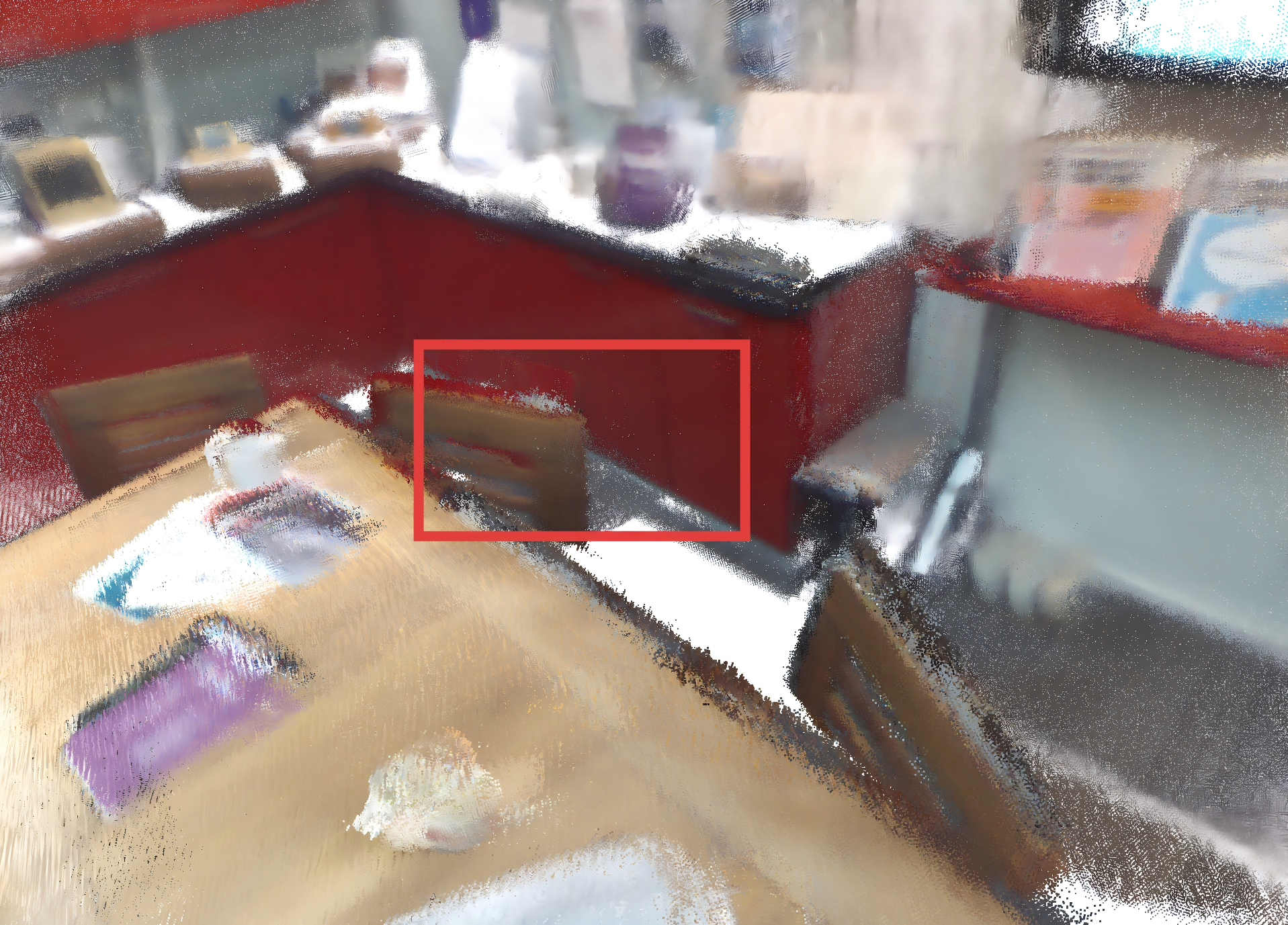}
    \caption{GT}
    \label{bilateral-c}
  \end{subfigure}
  \caption{Effect of online spatial geometry refinement.}
  \label{bilateral}
  \vspace{-5mm}
\end{figure}

Despite improvements introduced by pose-adaptive state updates and trajectory-consistent training, minor prediction noise may still accumulate during sequential inference. This noise often manifests as small trajectory jitter \cref{smoothing} or geometric artifacts in reconstructed point clouds \cref{bilateral}. To further improve robustness without introducing additional model complexity, we incorporate a lightweight online stabilization module that operates during inference.

\noindent \textbf{Temporal Trajectory Stabilization.} 
Camera trajectory jitter often arises from small frame-level prediction fluctuations. To reduce this effect, we apply online stabilization to the predicted camera movement. Given an input sequence of image frames $\bm{I^{n \times H \times W}}$, let the $i$-th frame be $\bm{I_i^{H \times W}}$ and its preceding frame be $\bm{I_{i-1}^{H \times W}}$. Let $\bm{\hat{x}_i} = \bm{X}\{\bm{I_i^{H \times W}}\}$, $\bm{\hat{x}_{i-1}} = \bm{X}\{\bm{I_{i-1}^{H \times W}}\}$, $\bm{\hat{q}_i} = \bm{Q}\{\bm{I_i^{H \times W}}\}$, and $\bm{\hat{q}_{i-1}} = \bm{Q}\{\bm{I_{i-1}^{H \times W}}\}$ represent the camera translation and rotation quantities predicted by the model in the world coordinate system for frames $i$ and $i-1$, respectively. We first apply One Euro filtering \cite{casiez20121} to the predicted camera translation, a process that can be formulated as follows Eq. \eqref{filtering}:
\begin{equation}
    \label{filtering}
    \begin{array}{cc}
          \bm{x_{i-1}} = \alpha_{i-2} \cdot \bm{\hat{x}_{i-1}} + (1 - \alpha_{i-2}) \cdot \bm{x_{i-2}} \\
          \bm{x_i} = \alpha_{i-1} \cdot \bm{\hat{x_i}} + (1 - \alpha_{i-1}) \cdot \bm{x_{i-1}}
    \end{array}
\end{equation}
\noindent In Eq. \eqref{filtering}, $\bm{x_{i-2}}$, $\bm{x_{i-1}}$, and $\bm{x_i}$ denote the smoothed translation vectors for the $(i-2)$-th, $(i-1)$-th, and $i$-th frames, respectively. $\alpha_{i-2}$ represents the smoothing factor between the $(i-1)$-th and $(i-2)$-th frames, and $\alpha_{i-1}$ represents the smoothing factor between the $i$-th and $(i-1)$-th frames. For the rotation quaternions, we first normalize the predicted quaternion $\bm{\hat{q}_i}$ of the $i$-th frame and the smoothed quaternion $\bm{q_{i-1}}$ of the $(i-1)$-th frame. Subsequently, we apply the Spherical Linear Interpolation (Slerp) function to perform temporal smoothing. The mathematical expression is as follows:
\begin{equation}
\label{smoothed quaternions}
    \begin{array}{cc}
    \bm{q_{i-1}} = \text{Slerp}(\bm{q_{i-2}}, \bm{\hat{q}_{i-1}}; \alpha_{i-2}) \\
    \bm{q_i} = \text{Slerp}(\bm{q_{i-1}}, \bm{\hat{q}_i}; \alpha_{i-1})
    \end{array}
\end{equation}
\textbf{Spatial Geometry Refinement.} Experimental observations revealed that during the final visualization stage, point cloud data predicted by the model often exhibited artifacts at object boundaries or unnatural protrusions on smooth surfaces, as shown in \cref{bilateral}. 
To resolve these issues, we apply online bilateral spatial filtering \cite{tomasi1998bilateral} to the point clouds during inference. This method effectively prevents mutual interference between data on different sides of an edge, thereby achieving smoothness in flat regions while preserving the sharpness of object contours.

\section{Experiments}

We evaluate PAS3R through a comprehensive experimental study covering camera pose estimation (Sec.~\ref{ssec:cam-pose}), depth prediction (Sec.~\ref{ssec:depth}), dense 3D reconstruction (Sec.~\ref{ssec:3d-recon}), and ablation study (Sec.~\ref{ssec:ablation}). These experiments aim to assess two key aspects of streaming reconstruction: long-horizon stability and geometric fidelity under continuous viewpoint changes.

Unlike offline reconstruction methods that process entire image sets simultaneously, streaming systems must update a persistent internal state as frames arrive sequentially. As a result, performance degradation often becomes more pronounced as the sequence length increases. Our evaluation therefore focuses not only on final reconstruction accuracy but also on how reconstruction quality evolves over time as the number of processed frames grows.

\noindent\textbf{Baselines.} We compare PAS3R with several recent state-of-the-art online reconstruction frameworks, including CUT3R \cite{wang2025continuous}, TTT3R \cite{anonymous2026tttr}, IVGGT \cite{yuan2026infinitevggt}, and Mem4D \cite{cai2025mem4d}. All models follow the same preprocessing pipeline as used in CUT3R and TTT3R to ensure fair comparison.
Experiments are conducted on a server equipped with 8 NVIDIA RTX 4090 GPUs. Importantly, PAS3R introduces only lightweight computations and maintains the constant-memory property required for streaming reconstruction, making the improvements applicable in practical long-video scenarios.

\subsection{Camera Pose Estimation} \label{ssec:cam-pose}

We first evaluate the quality of camera trajectory estimation on ScanNet \cite{dai2017scannet}, Sintel \cite{2012naturalistic} and Tum dynamic \cite{2012benchmark} datasets. Absolute Trajectory Error (ATE) and Relative Pose Error (RPE) are used as evaluation metrics.

\begin{figure}[tb]
  \centering
  \begin{subfigure}{0.32\linewidth}
    \centering
    \includegraphics[width=\linewidth]{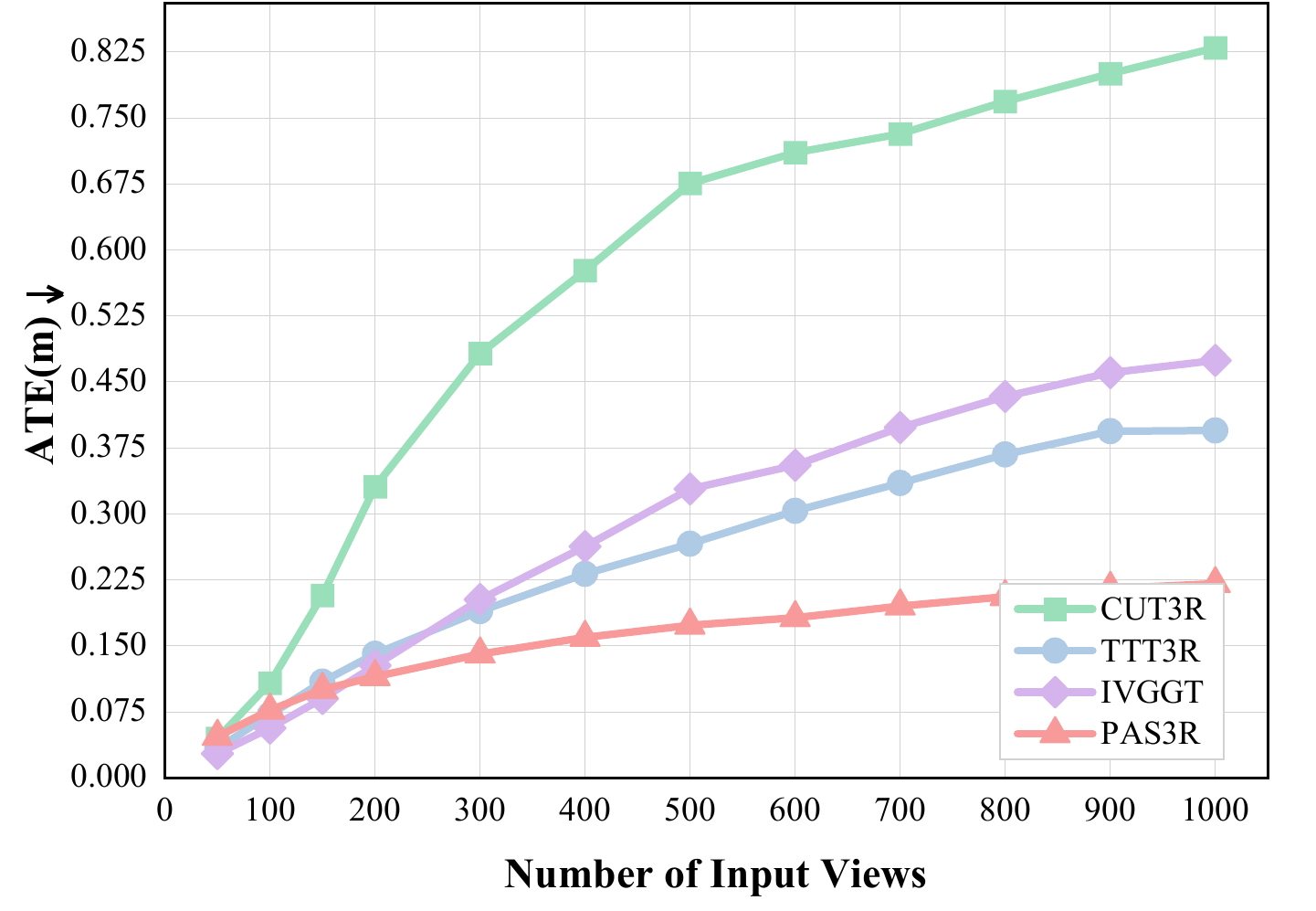}
    \caption{ATE on Scannet}
    \label{scannet-cp-a}
  \end{subfigure}
  \hfill 
  \begin{subfigure}{0.32\linewidth}
    \centering
    \includegraphics[width=\linewidth]{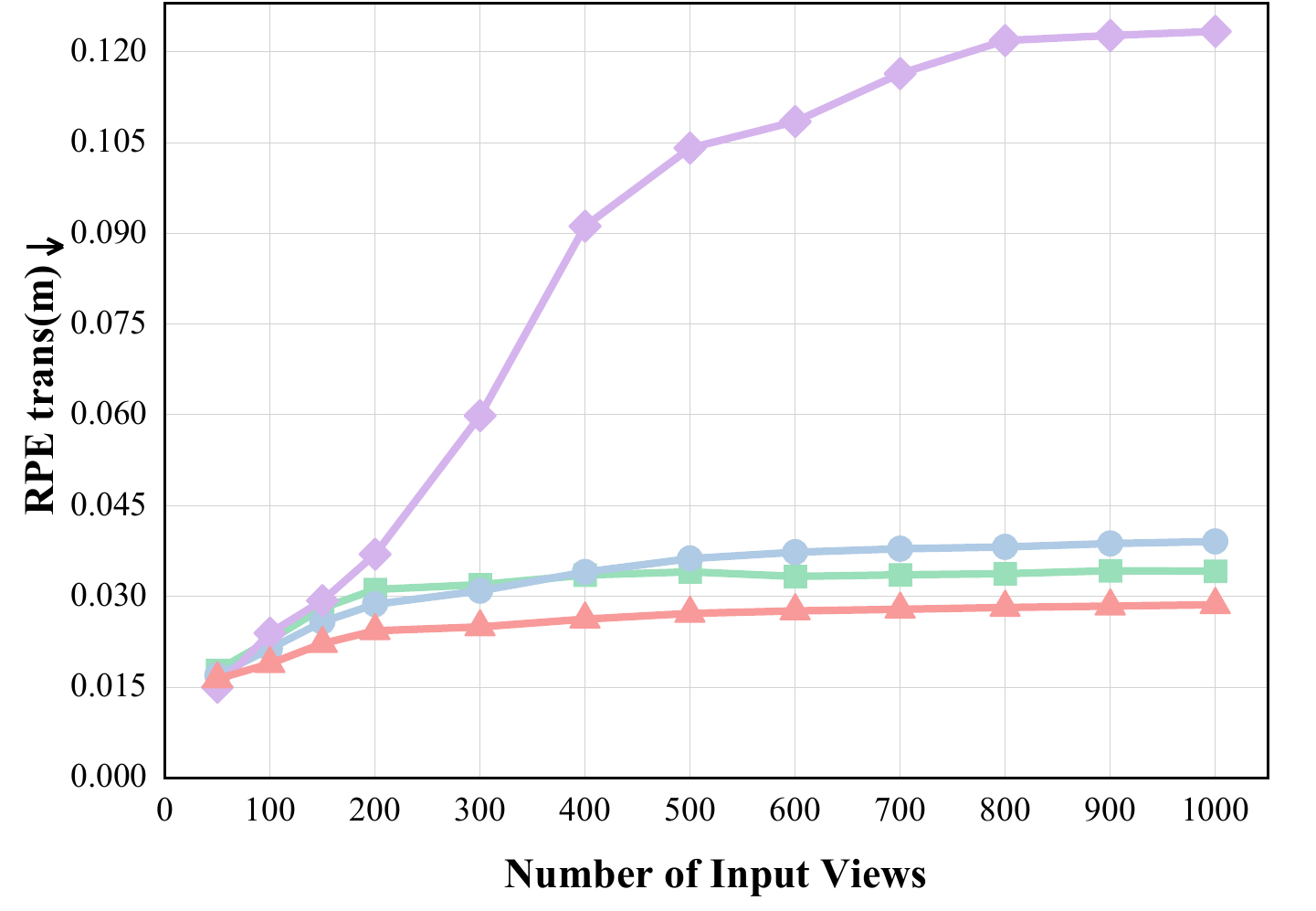}
    \caption{RPE trans on Scannet}
    \label{scannet-cp-b}
  \end{subfigure}
  \hfill 
  \begin{subfigure}{0.32\linewidth}
    \centering
    \includegraphics[width=\linewidth]{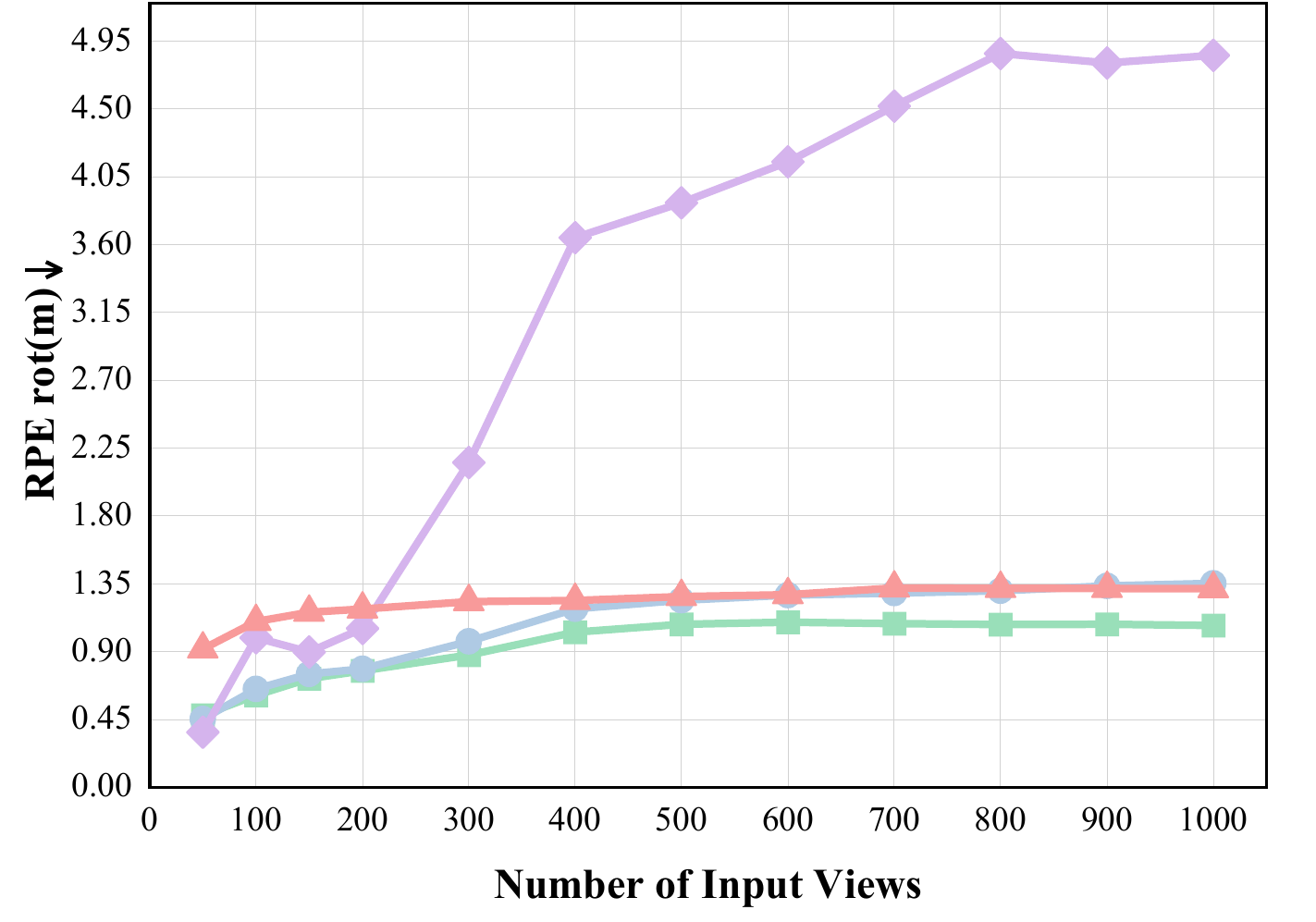}
    \caption{RPE rot on Scannet}
    \label{scannet-cp-c}
  \end{subfigure}
  \caption{Comparison of camera pose estimation on Scannet \cite{dai2017scannet}}
  \label{scannet-cp}
  \vspace{-5mm}
\end{figure}

We first conduct our evaluations on sequences ranging from the first 50 to 1000 frames for each scene on Scannet. 
\cref{scannet-cp} analyzes how pose accuracy evolves as the number of input frames increases on ScanNet sequences. This experiment reveals a key property of PAS3R: trajectory stability improves relative to competing methods as the sequence length grows. While most methods exhibit gradually increasing drift, the trajectory error of PAS3R grows much more slowly, indicating that the proposed pose-adaptive update mechanism effectively mitigates long-term error accumulation.

 Similarly to the experimental configuration of CUT3R, \cref{tab-comparison} further evaluates performance on shorter sequences from the Sintel and TUM datasets. PAS3R achieves the best translational RPE and remains competitive in ATE across both datasets. Although the rotational RPE is slightly higher than the best-performing baseline in some cases, the overall trajectory remains more stable over long horizons.
 
These results suggest that the primary advantage of PAS3R lies not only in local pose accuracy but also in maintaining consistent trajectory estimation throughout extended sequences, which is essential for reliable streaming reconstruction.

\begin{table}[tb]
    \centering
    \caption{Quantitative comparison on Sintel \cite{2012naturalistic} and Tum \cite{2012benchmark} datasets. \textbf{Bold} indicates the best performance, and \underline{underlined} indicates the second best. $\downarrow$ indicates lower is better.}
    \label{tab-comparison}
    \renewcommand{\arraystretch}{1.2}
    \setlength{\tabcolsep}{6pt}
    \begin{tabular}{lcccccc}
        \toprule
        \multirow{2}{*}{\textbf{Method}} & \multicolumn{3}{c}{\textbf{Sintel}} & \multicolumn{3}{c}{\textbf{Tum}} \\
        \cmidrule(lr){2-4} \cmidrule(lr){5-7}
         & ATE $\downarrow$ & RPE trans $\downarrow$ & RPE rot $\downarrow$ & ATE $\downarrow$ & RPE trans $\downarrow$ & RPE rot $\downarrow$ \\
        \midrule
        Mem4D \cite{cai2025mem4d} & 0.263 & 0.091 & 0.812 & 0.061 & 0.020 & 0.517 \\
        CUT3R \cite{wang2025continuous} & \textbf{0.209} & \underline{0.071} & \textbf{0.632} & 0.047 & 0.015 & 0.448 \\
        TTT3R \cite{anonymous2026tttr} & \underline{0.210} & 0.090 & 0.730 & \underline{0.029} & 0.013 & \underline{0.379} \\
        IVGGT \cite{yuan2026infinitevggt} & 0.237 & 0.096 & 0.806 & \textbf{0.027} & \underline{0.012} & \textbf{0.313} \\
        Ours  & 0.211 & \textbf{0.053} & \underline{0.688} & \textbf{0.027} & \textbf{0.011} & 0.475 \\
        \bottomrule
    \end{tabular}
    \vspace{-5mm}
\end{table}

\subsection{DEPTH ESTIMATION} \label{ssec:depth}

We next evaluate depth prediction accuracy using the Bonn dataset. Two standard metrics are used: Absolute Relative Error (Abs Rel) and $\delta < 1.25$ (the percentage of predicted depths within a $1.25$ factor of the ground truth). To ensure a comprehensive evaluation, we report depth estimation results under three different alignment settings: original (direct prediction), scale-aligned, and scale-and-shift-aligned. The sequence lengths for evaluation range . We initially evaluate the depth prediction performance from 50 to 500 frames on Bonn dataset \cite{2019refusion}.

\begin{figure}[tb]
  \captionsetup[subfigure]{justification=centering}
  \centering
  \begin{subfigure}{0.32\linewidth}
    \centering
    \includegraphics[width=\linewidth]{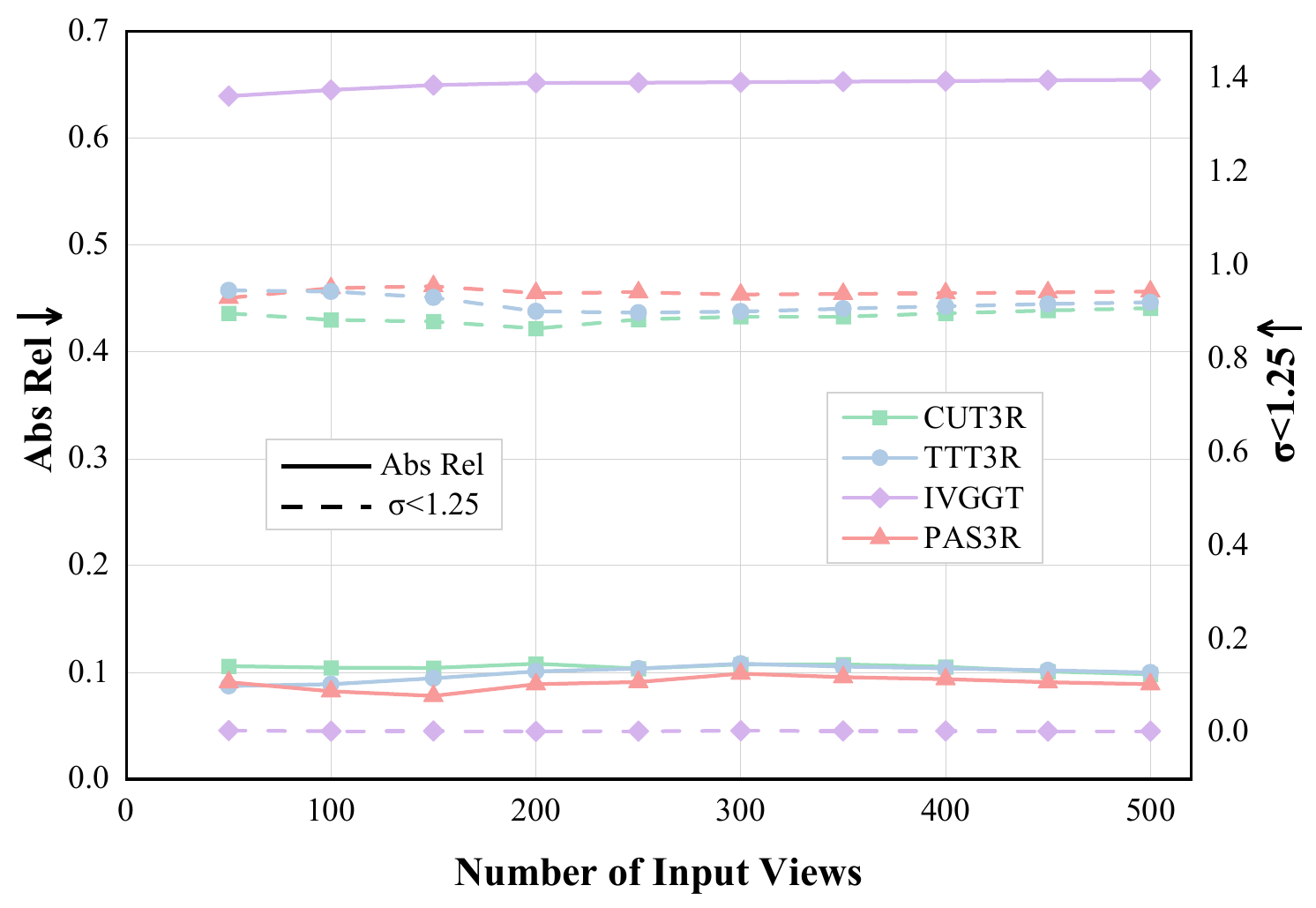}
    \caption{Depth evaluation on Bonn(oringinal)}
    \label{bonn-depth-a}
  \end{subfigure}
  \hfill 
  \begin{subfigure}{0.32\linewidth}
    \centering
    \includegraphics[width=\linewidth]{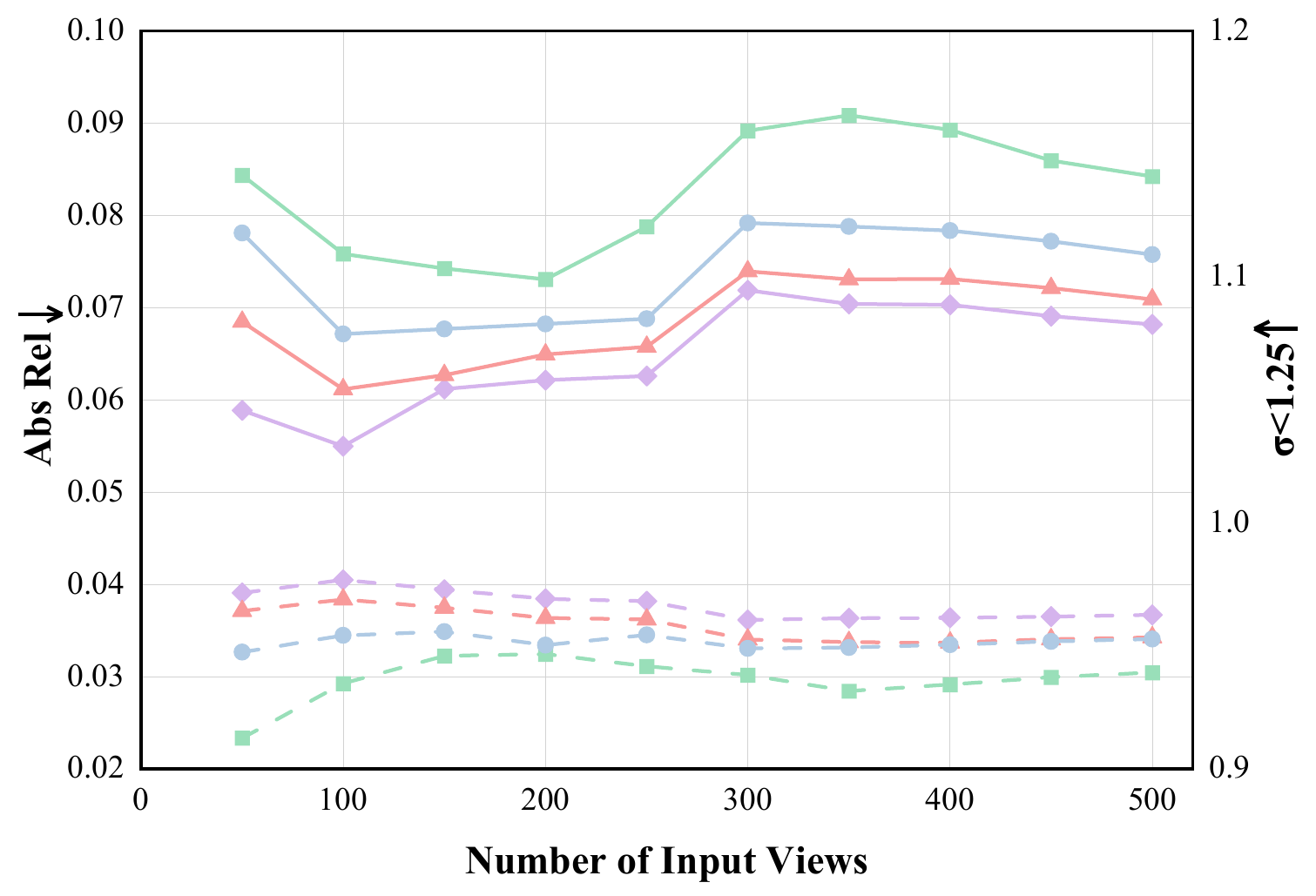}
    \caption{Depth evaluation on Bonn(Scale)}
    \label{bonn-depth-b}
  \end{subfigure}
  \hfill 
  \begin{subfigure}{0.32\linewidth}
    \centering
    \includegraphics[width=\linewidth]{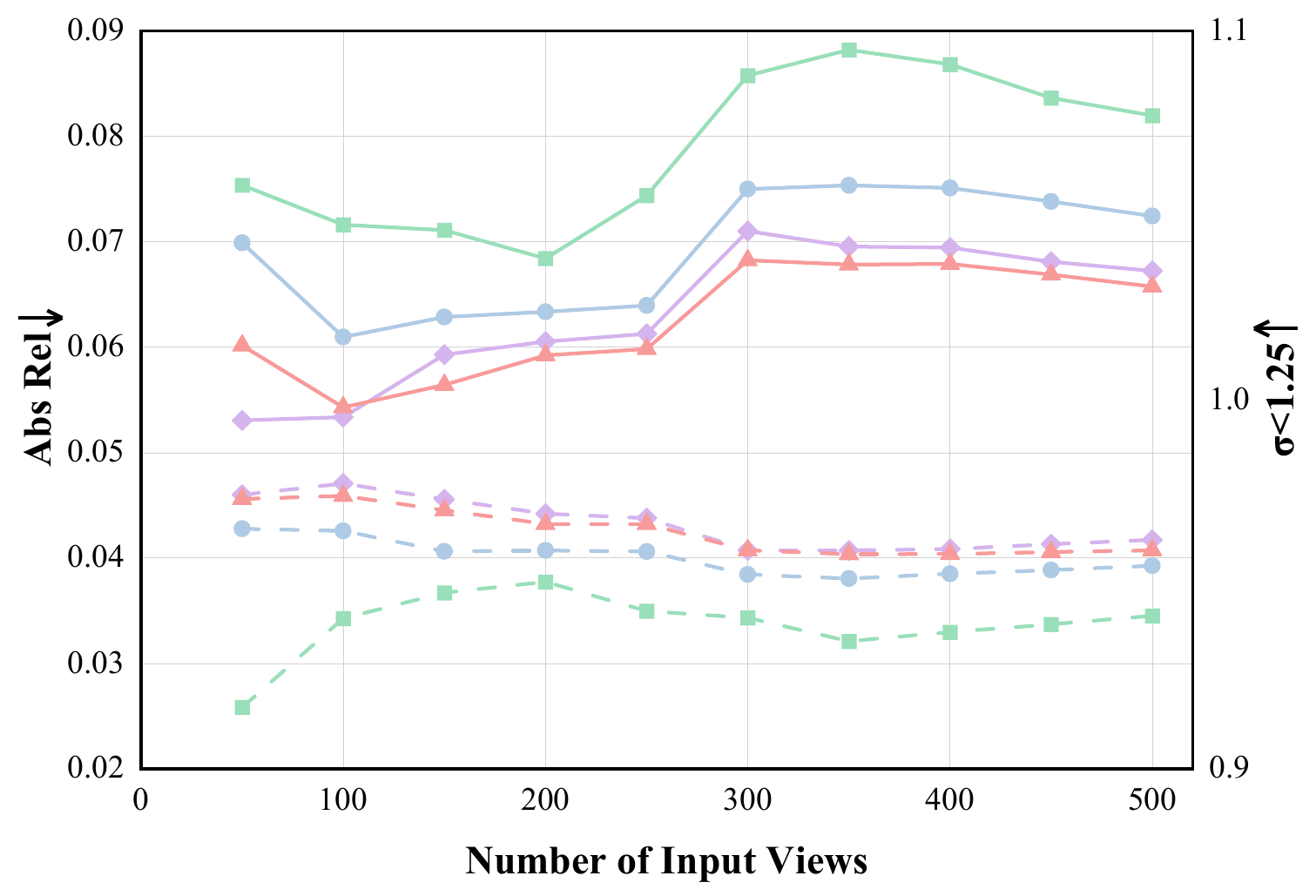}
    \caption{Depth evaluation on Bonn(Scale\&Shift)}
    \label{bonn-depth-c}
  \end{subfigure}
  \caption{Comparison of depth on Bonn \cite{2019refusion}}
  \label{bonn-depth}
  \vspace{-2mm}
\end{figure}

The results shown in \cref{bonn-depth} reveal an interesting pattern. PAS3R achieves the strongest performance in the original depth prediction setting, indicating that the predicted geometry is already well aligned with the ground truth without requiring post-hoc scaling. In contrast, some competing methods exhibit large errors in the original evaluation but improve significantly after scale alignment, suggesting that their predictions rely more heavily on global scaling correction.
This behavior reflects an important property of pose-adaptive updates. By adjusting the influence of incoming frames according to camera motion and scene structure, PAS3R maintains a more coherent internal representation of scene geometry. As a result, the predicted depth remains consistent even without additional alignment.

\cref{tab-comparison_multi_dataset} further reports results on short sequences from Sintel \cite{2012naturalistic}, Bonn \cite{2019refusion}, and KITTI \cite{2013vision} datasets. PAS3R maintains performance comparable to state-of-the-art online reconstruction methods across all datasets, demonstrating that the improvements in long-sequence stability do not compromise performance in short sequences.
\begin{table}[tb]
    \centering
    \caption{Quantitative comparison on Sintel, Bonn \cite{2019refusion}, and Kitti \cite{2013vision} datasets. \textbf{Bold} indicates the best performance, and \underline{underlined} indicates the second best. $\downarrow$ indicates lower is better, while $\uparrow$ indicates higher is better.}
    \label{tab-comparison_multi_dataset}
    \renewcommand{\arraystretch}{1.2}
    \setlength{\tabcolsep}{5pt}
    \begin{tabular}{lcccccc}
        \toprule
        \multirow{2}{*}{\textbf{Method}} & \multicolumn{2}{c}{\textbf{Sintel}} & \multicolumn{2}{c}{\textbf{Bonn}} & \multicolumn{2}{c}{\textbf{Kitti}} \\
        \cmidrule(lr){2-3} \cmidrule(lr){4-5} \cmidrule(lr){6-7}
         & Abs Rel $\downarrow$ & $\sigma < 1.25 \uparrow$ & Abs Rel $\downarrow$ & $\sigma < 1.25 \uparrow$ & Abs Rel $\downarrow$ & $\sigma < 1.25 \uparrow$ \\
        \midrule
        Mem4D \cite{cai2025mem4d} & 0.520 & 43.1 & 0.072 & 95.7 & 0.140 & 82.0 \\
        CUT3R \cite{wang2025continuous} & 0.432 & 47.0 & 0.077 & 93.8 & 0.122 & 87.6 \\
        TTT3R \cite{anonymous2026tttr} & \underline{0.406} & \underline{48.8} & 0.069 & 95.4 & \textbf{0.114} & \underline{90.6} \\
        IVGGT \cite{yuan2026infinitevggt} & \textbf{0.329} & \textbf{66.0} & \textbf{0.059} & \textbf{97.4} & 0.185 & 69.8 \\
        Ours  & 0.407 & 48.7 & \underline{0.064} & \underline{96.6} & \underline{0.115} & \textbf{91.4} \\
        \bottomrule
    \end{tabular}
    \vspace{-5mm}
\end{table}

\subsection{3D RECONSTRUCTION} \label{ssec:3d-recon}

To evaluate the quality of reconstructed geometry, we conduct experiments on the 7-Scenes dataset using Accuracy (Acc), Completeness (Comp), and Normal Consistency (NC) as evaluation metrics. The evaluation is conducted on video sequences ranging from 50 to 400 frames. 
\begin{figure}[tb]
  \captionsetup[subfigure]{justification=centering}
  \centering
  \begin{subfigure}{0.32\linewidth}
    \centering
    \includegraphics[width=\linewidth]{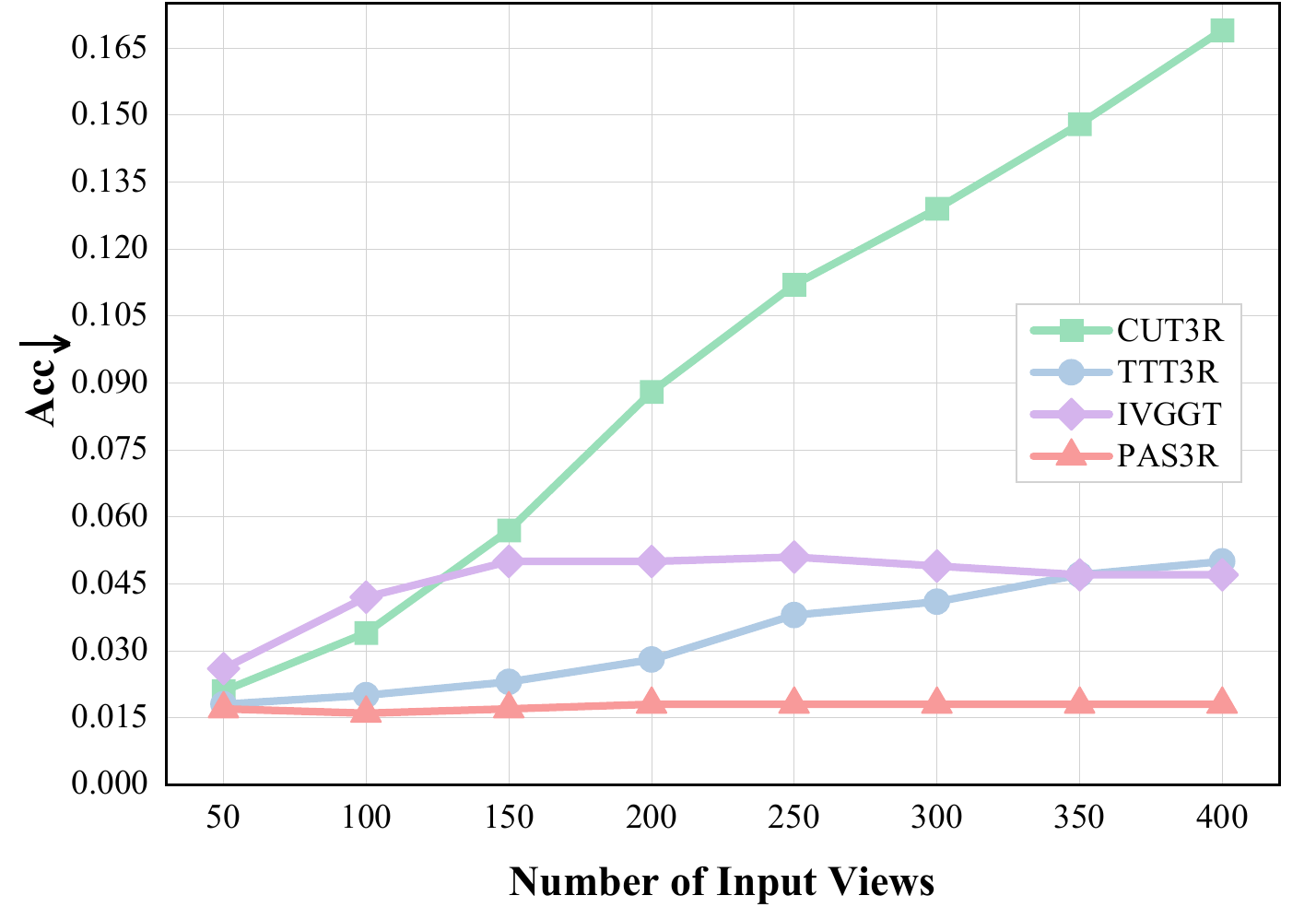}
    \caption{Acc on 7scene}
    \label{7scene-3D-a}
  \end{subfigure}
  \hfill 
  \begin{subfigure}{0.32\linewidth}
    \centering
    \includegraphics[width=\linewidth]{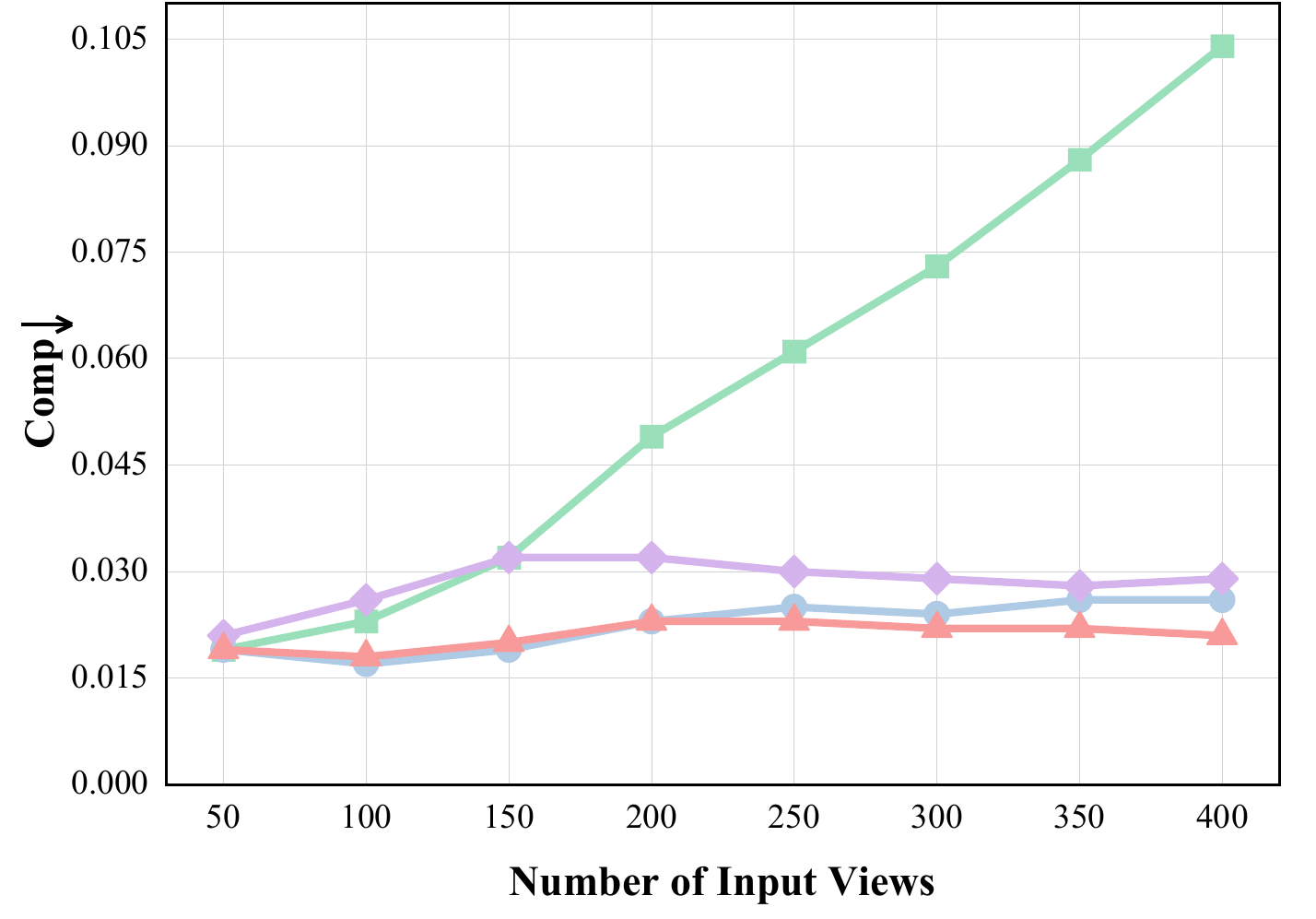}
    \caption{Comp on 7scene}
    \label{7scene-3D-b}
  \end{subfigure}
  \hfill 
  \begin{subfigure}{0.32\linewidth}
    \centering
    \includegraphics[width=\linewidth]{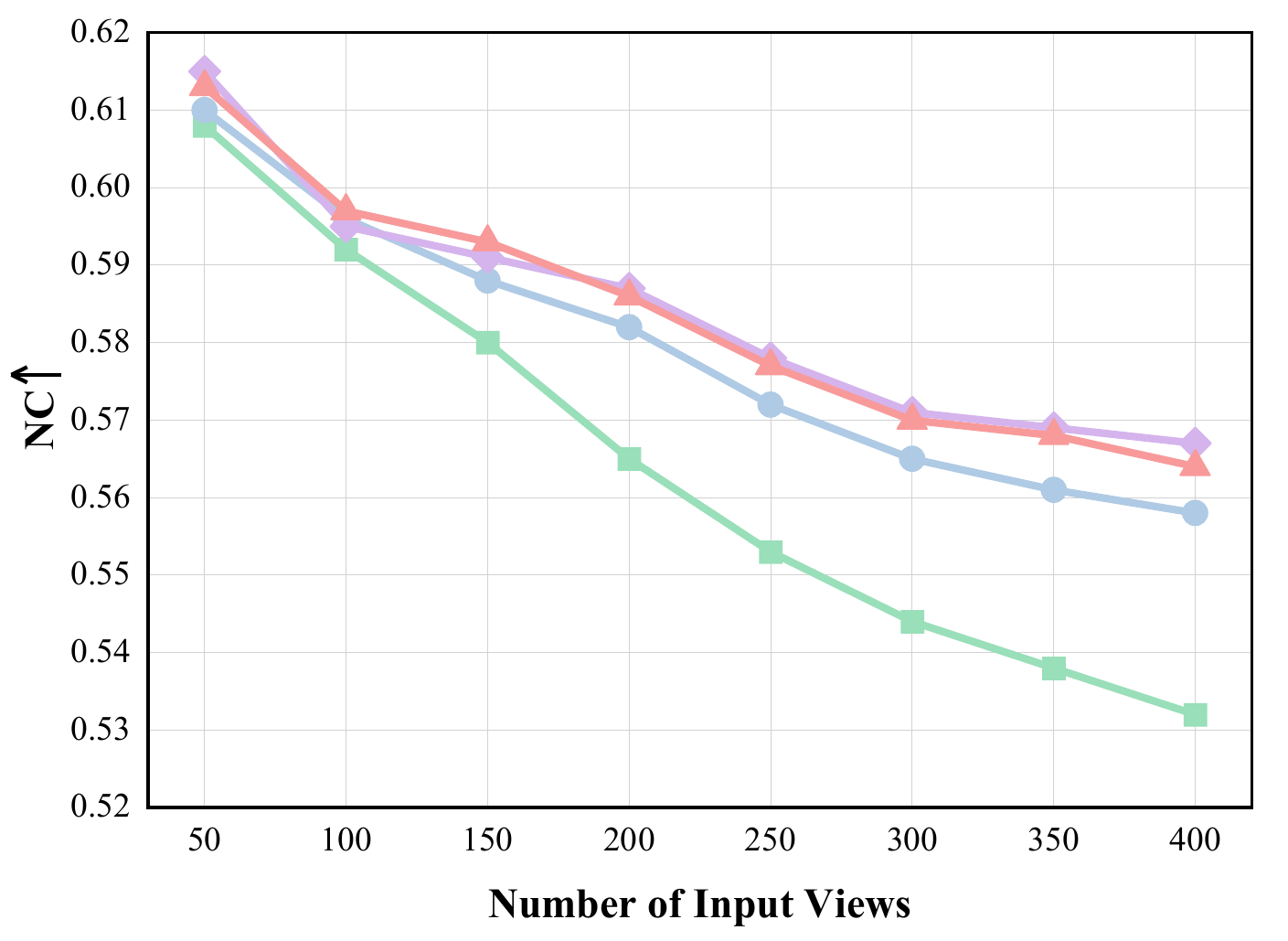}
    \caption{NC on 7scene}
    \label{7scene-3D-c}
  \end{subfigure}
  \caption{Comparison of 3D Reconstruction on 7scene \cite{2013scene}}
  \label{7scene-3D}
  \vspace{-5mm}
\end{figure}
The quantitative comparison results in \cref{7scene-3D} illustrate how reconstruction quality evolves as the number of processed frames increases. A clear trend emerges: while competing methods gradually degrade as the sequence length grows, PAS3R maintains relatively stable reconstruction metrics across the entire sequence. This stability highlights the advantage of the pose-adaptive update strategy, which prevents new frames from excessively overwriting previously accumulated geometry. Quantitatively, PAS3R achieves the best performance in both Accuracy and Completeness while maintaining competitive Normal Consistency. These results indicate that the proposed framework improves both global geometric alignment and local surface reconstruction quality.

\begin{figure*}[t]
    \centering
    \begin{tikzpicture}
        \node[anchor=south west, inner sep=0] (pic) at (0,0) {\includegraphics[width=0.9\linewidth]{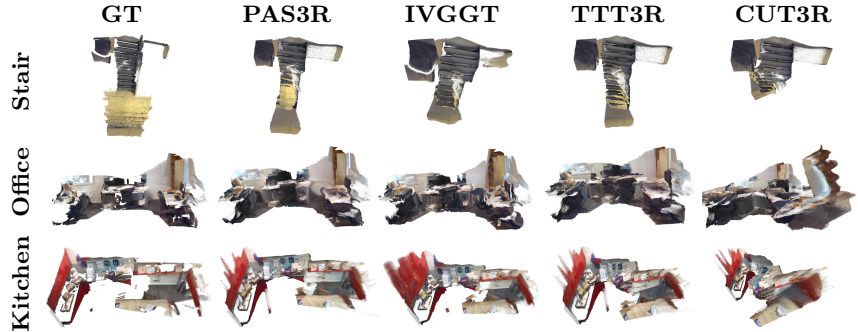}};

        \begin{scope}[x={(pic.south east)}, y={(pic.north west)}]
            \tikzset{header/.style={
                font=\small\bfseries, 
                inner sep=0pt, 
                text height=1.5ex, 
                text depth=0.25ex
            }}

            \node[header, anchor=south] at (0.10, 1.01) {GT};
            \node[header, anchor=south] at (0.30, 1.01) {PAS3R};
            \node[header, anchor=south] at (0.50, 1.01) {IVGGT};
            \node[header, anchor=south] at (0.70, 1.01) {TTT3R};
            \node[header, anchor=south] at (0.90, 1.01) {CUT3R};

            \node[header, rotate=90, anchor=center] at (-0.02, 0.83) {Stair};
            \node[header, rotate=90, anchor=center] at (-0.02, 0.50) {Office};
            \node[header, rotate=90, anchor=center] at (-0.02, 0.17) {Kitchen};
        \end{scope}
    \end{tikzpicture}
    \vspace{-2mm}
    \caption{Qualitative comparison of 3D reconstruction results. In the Stair scene, methods such as IVGGT and CUT3R suffer from large spatial drift. In the Office and Kitchen scene, IVGGT, TTT3R, and CUT3R all experience varying degrees of structural collapse at wall corners, whereas PAS3R remains well aligned with the GT.}
    \label{point_cloud}
    \vspace{-5mm}
\end{figure*}

\cref{point_cloud} provides qualitative comparisons across three representative scenes (Stair, Office, and Kitchen). Competing methods often exhibit spatial drift or structural distortion, particularly in regions with large viewpoint changes. In contrast, PAS3R reconstructs more coherent scene geometry with fewer structural artifacts. This improvement is especially noticeable in complex indoor environments where repeated viewpoint changes occur.

\subsection{Ablation Study} \label{ssec:ablation}

We validate the design of our proposed components through a series of ablation experiments, as summarized in \cref{ablation_study}. 
 We evaluate camera pose estimation, depth estimation, and 3D point cloud reconstruction on the TUM (1,000 frames), Bonn (500 frames), and 7-Scene (400 frames) datasets, respectively. According to the results of the ablation study, the pose-adaptive state update modulation significantly improves camera pose estimation, depth prediction, and point cloud reconstruction quality. Trajectory-consistent training yields consistent improvements across all metrics, thereby enhancing the overall performance of the framework. Finally, removing the online spatiotemporal stabilization module slightly improves certain individual metrics but degrades overall trajectory stability. This highlights the complementary role of the stabilization module: while it may introduce minor smoothing effects in isolated metrics, it significantly improves overall reconstruction robustness in streaming scenarios.
\begin{table*}[t]
\centering
\caption{Quantitative ablation study results on Tum1000 \cite{2012benchmark}, Bonn500 \cite{2019refusion}, and 7Scene400 \cite{2013scene} datasets. The best results are highlighted in \textbf{bold}, and the second-best results are \underline{underlined}. $\downarrow$ indicates lower is better, while $\uparrow$ indicates higher is better.}
\label{ablation_study}
\resizebox{\textwidth}{!}{
\begin{tabular}{lccccccccc}
\toprule
\multirow{2}{*}{Method} & \multicolumn{3}{c}{Tum1000} & \multicolumn{2}{c}{Bonn500} & \multicolumn{3}{c}{7Scene400} \\
\cmidrule(lr){2-4} \cmidrule(lr){5-6} \cmidrule(lr){7-9}
& ATE $\downarrow$ & RPE trans $\downarrow$ & RPE rot $\downarrow$ & Abs Rel $\downarrow$ & $\delta < 1.25 \uparrow$ & Acc $\downarrow$ & Comp $\downarrow$ & NC $\uparrow$ \\
\midrule
w/o pose-adaptive state update    & 0.10854 & \underline{0.00374} & \textbf{0.46871} & 0.0757 & 95.24 & 0.048 & 0.028 & \underline{0.563} \\
w/o trajectory-consistent training   & 0.05616 & 0.00459 & 0.59215 & \underline{0.0710} & \underline{95.35} & 0.023 & \underline{0.021} & 0.562 \\
w/o spatiotemporal stabilization   & \textbf{0.04980} & 0.01138 & 0.61098 & \underline{0.0710} & \textbf{95.36} & \underline{0.019} & \textbf{0.019} & 0.561 \\
Full Method             & \underline{0.05214} & \textbf{0.00354} & \underline{0.52595} & \textbf{0.0709} & \underline{95.35} & \textbf{0.018} & \underline{0.021} & \textbf{0.564} \\
\bottomrule
\end{tabular}
}
\vspace{-5mm}
\end{table*}

\section{Conclusion}
We presented PAS3R, a pose-adaptive framework for streaming monocular 3D reconstruction that addresses the stability–adaptation dilemma in long video sequences. By dynamically regulating frame influence according to geometric novelty, PAS3R enables rapid adaptation to new viewpoints while preserving previously accumulated scene structure. Combined with trajectory-consistent training and lightweight online stabilization, the framework significantly improves camera pose estimation, depth prediction, and reconstruction quality across multiple benchmarks, particularly on long sequences. These results highlight the importance of pose-aware update strategies for robust streaming reconstruction.

\noindent\textbf{Limitations.} Despite its effectiveness, PAS3R still has limitations. Current benchmarks provide limited coverage of diverse long video streams, and rotational trajectory accuracy can be further improved.

\section*{Acknowledgements}
This paper is supported by NSFC under grant No. 62495092, 62125305 and Natural Science Basic Research Plan in Shaanxi Province of China (No. 2025SYS-SYSZD-023). Cheng Wang is funded by Royal Society ISPF International Collaboration Award (ICA\textbackslash R2\textbackslash252130).

%
%
\bibliographystyle{splncs04}
\bibliography{main}

@String(PAMI  = {IEEE Trans. Pattern Anal. Mach. Intell.})

@String(IJCV  = {Int. J. Comput. Vis.})

@String(CVPR  = {IEEE Conf. Comput. Vis. Pattern Recog.})

@String(ICCV  = {Int. Conf. Comput. Vis.})

@String(ECCV  = {Eur. Conf. Comput. Vis.})

@String(NeurIPS = {Adv. Neural Inform. Process. Syst.})

@String(ICML  = {Int. Conf. Mach. Learn.})

@String(ICLR  = {Int. Conf. Learn. Represent.})

@String(ACCV  = {Asian Conf. Comput. Vis.})

@String(TOG   = {ACM Trans. Graph.})

@String(TIP   = {IEEE Trans. Image Process.})

@String(TMM   = {IEEE Trans. Multimedia})

@String(PAMI  = {IEEE TPAMI})

@String(IJCV  = {IJCV})

@String(CVPR  = {CVPR})

@String(ICCV  = {ICCV})

@String(ECCV  = {ECCV})

@String(NeurIPS = {NeurIPS})

@String(ICML  = {ICML})

@String(ICLR  = {ICLR})

@String(ACCV  = {ACCV})

@String(TOG   = {ACM TOG})

@String(TIP   = {IEEE TIP})

@String(TMM   =	{IEEE TMM})

@inproceedings{qi2017pointnet,
  title = {Pointnet: Deep learning on point sets for 3d classification and segmentation},
  author = {Qi, Charles R and Su, Hao and Mo, Kaichun and Guibas, Leonidas J},
  booktitle = CVPR,
  pages = {652--660},
  year = 2017
}

@article{qi2017pointnet++,
  title = {Pointnet++: Deep hierarchical feature learning on point sets in a metric space},
  author = {Qi, Charles Ruizhongtai and Yi, Li and Su, Hao and Guibas, Leonidas J},
  journal = NeurIPS,
  volume = {30},
  year = 2017
}

@inproceedings{wang2018pixel2mesh,
  title = {Pixel2mesh: Generating 3d mesh models from single rgb images},
  author = {Wang, Nanyang and Zhang, Yinda and Li, Zhuwen and Fu, Yanwei and Liu, Wei and Jiang, Yu-Gang},
  booktitle = ECCV,
  pages = {52--67},
  year = 2018
}

@inproceedings{groueix2018papier,
  title = {A papier-m{\^a}ch{\'e} approach to learning 3d surface generation},
  author = {Groueix, Thibault and Fisher, Matthew and Kim, Vladimir G and Russell, Bryan C and Aubry, Mathieu},
  booktitle = CVPR,
  pages = {216--224},
  year = 2018
}

@inproceedings{choy20163d,
  title = {3d-r2n2: A unified approach for single and multi-view 3d object reconstruction},
  author = {Choy, Christopher B and Xu, Danfei and Gwak, JunYoung and Chen, Kevin and Savarese, Silvio},
  booktitle = ECCV,
  pages = {628--644},
  year = 2016,
  organization ={Springer}
}

@article{wu2016learning,
  title = {Learning a probabilistic latent space of object shapes via 3d generative-adversarial modeling},
  author = {Wu, Jiajun and Zhang, Chengkai and Xue, Tianfan and Freeman, Bill and Tenenbaum, Josh},
  journal = NeurIPS,
  volume = 29,
  year = 2016
}

@article{liu2020dlgan,
  title = {DLGAN: Depth-preserving latent generative adversarial network for 3D reconstruction},
  author = {Liu, Caixia and Kong, Dehui and Wang, Shaofan and Li, Jinghua and Yin, Baocai},
  journal = TMM,
  volume = 23,
  pages = {2843--2856},
  year = 2020,
  publisher = {IEEE}
}

@article{kerbl20233d,
  title = {3D Gaussian splatting for real-time radiance field rendering.},
  author = {Kerbl, Bernhard and Kopanas, Georgios and Leimk{\"u}hler, Thomas and Drettakis, George},
  journal = TOG,
  volume = 42,
  number = 4,
  pages = {139--1},
  year = 2023
}

@inproceedings{wu20244d,
  title = {4d gaussian splatting for real-time dynamic scene rendering},
  author = {Wu, Guanjun and Yi, Taoran and Fang, Jiemin and Xie, Lingxi and Zhang, Xiaopeng and Wei, Wei and Liu, Wenyu and Tian, Qi and Wang, Xinggang},
  booktitle = CVPR,
  pages = {20310--20320},
  year = 2024
}

@inproceedings{lu2024scaffold,
  title = {Scaffold-gs: Structured 3d gaussians for view-adaptive rendering},
  author = {Lu, Tao and Yu, Mulin and Xu, Linning and Xiangli, Yuanbo and Wang, Limin and Lin, Dahua and Dai, Bo},
  booktitle = CVPR,
  pages = {20654--20664},
  year = 2024
}

@inproceedings{guo2024tetsphere,
 author = {Guo, Minghao and Wang, Bohan and He, Kaiming and Matusik, Wojciech},
 booktitle = ICLR,
 editor = {Y. Yue and A. Garg and N. Peng and F. Sha and R. Yu},
 pages = {30290--30314},
 title = {TetSphere Splatting: Representing High-Quality Geometry with Lagrangian Volumetric Meshes},
 url = {https://proceedings.iclr.cc/paper_files/paper/2025/file/4ae7d78ebbe48f772e31c5c3fcc04c43-Paper-Conference.pdf},
 volume = 2025,
 year = 2025
}

@inproceedings{mildenhall2020nerf,
  title = {NeRF: Representing scenes as neural radiance fields for view synthesis},
  author = {Mildenhall, Ben and Srinivasan, Pratul P and Tancik, Matthew and Barron, Jonathan T and Ramamoorthi, Ravi and Ng, Ren},
  booktitle = ECCV,
  pages = {405--421},
  year = 2020,
  organization = {Springer}
}

@inproceedings{park2019deepsdf,
  title = {Deepsdf: Learning continuous signed distance functions for shape representation},
  author = {Park, Jeong Joon and Florence, Peter and Straub, Julian and Newcombe, Richard and Lovegrove, Steven},
  booktitle = CVPR,
  pages = {165--174},
  year = 2019
}

@article{muller2022instant,
  title = {Instant neural graphics primitives with a multiresolution hash encoding},
  author = {M{\"u}ller, Thomas and Evans, Alex and Schied, Christoph and Keller, Alexander},
  journal = TOG,
  volume = 41,
  number = 4,
  pages = {1--15},
  year = 2022,
  publisher = {ACM New York, NY, USA}
}

@inproceedings{wang2021neus,
  title = {NeuS: Learning Neural Implicit Surfaces by Volume Rendering for Multi-view Reconstruction},
  author = {Wang, P and Liu, L and Liu, Y and Theobalt, C and Komura, T and Wang, WP},
  booktitle = NeurIPS,
  year = 2021,
  organization = {Curran Associates.}
}

@inproceedings{schonberger2016structure,
  title = {Structure-from-motion revisited},
  author = {Schonberger, Johannes L and Frahm, Jan-Michael},
  booktitle = CVPR,
  pages = {4104--4113},
  year = 2016
}

@inproceedings{yao2018mvsnet,
  title = {Mvsnet: Depth inference for unstructured multi-view stereo},
  author = {Yao, Yao and Luo, Zixin and Li, Shiwei and Fang, Tian and Quan, Long},
  booktitle = ECCV,
  pages = {767--783},
  year = 2018
}

@article{barnes2009patchmatch,
  title={PatchMatch: A randomized correspondence algorithm for structural image editing},
  author={Barnes, Connelly and Shechtman, Eli and Finkelstein, Adam and Goldman, Dan B},
  journal = TOG,
  volume= 28,
  number= 3,
  pages= 24,
  year= 2009
}

@inproceedings{leroy2024grounding,
  title = {Grounding image matching in 3d with mast3r},
  author = {Leroy, Vincent and Cabon, Yohann and Revaud, J{\'e}r{\^o}me},
  booktitle = ECCV,
  pages = {71--91},
  year = 2024,
  organization = {Springer}
}

@article{teed2021droid,
  title = {Droid-slam: Deep visual slam for monocular, stereo, and rgb-d cameras},
  author = {Teed, Zachary and Deng, Jia},
  journal = NeurIPS,
  volume = 34,
  pages = {16558--16569},
  year = 2021
}

@inproceedings{wang2024dust3r,
  title = {Dust3r: Geometric 3d vision made easy},
  author = {Wang, Shuzhe and Leroy, Vincent and Cabon, Yohann and Chidlovskii, Boris and Revaud, Jerome},
  booktitle = CVPR,
  pages = {20697--20709},
  year = 2024
}

@inproceedings{wang2025continuous,
  title = {Continuous 3d perception model with persistent state},
  author = {Wang, Qianqian and Zhang, Yifei and Holynski, Aleksander and Efros, Alexei A and Kanazawa, Angjoo},
  booktitle = CVPR,
  pages = {10510--10522},
  year = 2025
}

@inproceedings{anonymous2026tttr,
title = {TTT3R: 3D Reconstruction as Test-Time Training},
author = {Chen, Xingyu and Chen, Yue and Xiu, Yuliang and Geiger, Andreas and Chen, Anpei},
booktitle = ICLR,
year = 2026,
url = {https://openreview.net/forum?id=aMs6FtNaY5}
}

@article{yuan2026infinitevggt,
  title = {InfiniteVGGT: Visual Geometry Grounded Transformer for Endless Streams},
  author = {Yuan, Shuai and Yang, Yantai and Yang, Xiaotian and Zhang, Xupeng and Zhao, Zhonghao and Zhang, Lingming and Zhang, Zhipeng},
  journal = {arXiv preprint arXiv:2601.02281},
  year = 2026
}

@inproceedings{wang2025vggt,
  title = {Vggt: Visual geometry grounded transformer},
  author = {Wang, Jianyuan and Chen, Minghao and Karaev, Nikita and Vedaldi, Andrea and Rupprecht, Christian and Novotny, David},
  booktitle = CVPR,
  pages = {5294--5306},
  year = 2025
}

@inproceedings{yang2025fast3r,
  title = {Fast3r: Towards 3d reconstruction of 1000+ images in one forward pass},
  author = {Yang, Jianing and Sax, Alexander and Liang, Kevin J and Henaff, Mikael and Tang, Hao and Cao, Ang and Chai, Joyce and Meier, Franziska and Feiszli, Matt},
  booktitle = CVPR,
  pages = {21924--21935},
  year = 2025
}

@article{zhuo2025streaming,
  title = {Streaming 4d visual geometry transformer},
  author = {Zhuo, Dong and Zheng, Wenzhao and Guo, Jiahe and Wu, Yuqi and Zhou, Jie and Lu, Jiwen},
  journal = {arXiv preprint arXiv:2507.11539},
  year = 2025
}

@article{zhang2025efficiently,
  title = {Efficiently Reconstructing Dynamic Scenes One D4RT at a Time},
  author = {Zhang, Chuhan and Moing, Guillaume Le and Koppula, Skanda and Rocco, Ignacio and Momeni, Liliane and Xie, Junyu and Sun, Shuyang and Sukthankar, Rahul and Barral, Jo{\"e}lle K and Hadsell, Raia and others},
  journal = {arXiv preprint arXiv:2512.08924},
  year = 2025
}

@article{vaswani2017attention,
  title = {Attention is all you need},
  author = {Vaswani, Ashish and Shazeer, Noam and Parmar, Niki and Uszkoreit, Jakob and Jones, Llion and Gomez, Aidan N and Kaiser, {\L}ukasz and Polosukhin, Illia},
  journal = NeurIPS,
  volume = 30,
  year = 2017
}

@inproceedings{casiez20121,
  title = {1€ filter: a simple speed-based low-pass filter for noisy input in interactive systems},
  author = {Casiez, G{\'e}ry and Roussel, Nicolas and Vogel, Daniel},
  booktitle = {Proceedings of the SIGCHI Conference on Human Factors in Computing Systems},
  pages = {2527--2530},
  year = 2012
}

@inproceedings{tomasi1998bilateral,
  title={Bilateral filtering for gray and color images},
  author={Tomasi, Carlo and Manduchi, Roberto},
  booktitle = ICCV,
  pages = {839--846},
  year = 1998,
  organization={IEEE}
}

@book{hartley2003multiple,
  title = {Multiple view geometry in computer vision},
  author = {Hartley, Richard and Zisserman, Andrew},
  year = 2003,
  publisher = {Cambridge university press}
}

@book{faugeras1993three,
  title = {Three-dimensional computer vision: a geometric viewpoint},
  author = {Faugeras, Olivier},
  year = 1993,
  publisher = {MIT press}
}

@book{szeliski2022computer,
  title = {Computer vision: algorithms and applications},
  author = {Szeliski, Richard},
  year = 2022,
  publisher = {Springer Nature}
}

@inproceedings{schonberger2016pixelwise,
  title = {Pixelwise view selection for unstructured multi-view stereo},
  author = {Sch{\"o}nberger, Johannes L and Zheng, Enliang and Frahm, Jan-Michael and Pollefeys, Marc},
  booktitle = ECCV,
  pages = {501--518},
  year = 2016,
  organization = {Springer}
}

@inproceedings{schonberger2016vote,
  title = {A vote-and-verify strategy for fast spatial verification in image retrieval},
  author = {Sch{\"o}nberger, Johannes L and Price, True and Sattler, Torsten and Frahm, Jan-Michael and Pollefeys, Marc},
  booktitle = ACCV,
  pages = {321--337},
  year = 2016,
  organization = {Springer}
}

@article{lowe2004distinctive,
  title = {Distinctive image features from scale-invariant keypoints},
  author = {Lowe, David G},
  journal = IJCV,
  volume = 60,
  number = 2,
  pages = {91--110},
  year = 2004,
  publisher = {Springer}
}

@inproceedings{rublee2011orb,
  title = {ORB: An efficient alternative to SIFT or SURF},
  author = {Rublee, Ethan and Rabaud, Vincent and Konolige, Kurt and Bradski, Gary},
  booktitle = ICCV,
  pages = {2564--2571},
  year = 2011,
  organization = {IEEE}
}

@inproceedings{detone2018superpoint,
  title = {Superpoint: Self-supervised interest point detection and description},
  author = {DeTone, Daniel and Malisiewicz, Tomasz and Rabinovich, Andrew},
  booktitle = CVPR,
  pages = {224--236},
  year = 2018
}

@inproceedings{sarlin2020superglue,
  title = {Superglue: Learning feature matching with graph neural networks},
  author = {Sarlin, Paul-Edouard and DeTone, Daniel and Malisiewicz, Tomasz and Rabinovich, Andrew},
  booktitle = CVPR,
  pages = {4938--4947},
  year = 2020
}

@inproceedings{triggs1999bundle,
  title = {Bundle adjustment—a modern synthesis},
  author = {Triggs, Bill and McLauchlan, Philip F and Hartley, Richard I and Fitzgibbon, Andrew W},
  booktitle = {International workshop on vision algorithms},
  pages = {298--372},
  year = 1999,
  organization = {Springer}
}

@inproceedings{agarwal2010bundle,
  title = {Bundle adjustment in the large},
  author = {Agarwal, Sameer and Snavely, Noah and Seitz, Steven M and Szeliski, Richard},
  booktitle = ECCV,
  pages = {29--42},
  year = 2010,
  organization = {Springer}
}

@article{mur2017orb,
  title = {Orb-slam2: An open-source slam system for monocular, stereo, and rgb-d cameras},
  author = {Mur-Artal, Raul and Tard{\'o}s, Juan D},
  journal = {IEEE Transactions on robotics},
  volume = 33,
  number = 5,
  pages = {1255--1262},
  year = 2017,
  publisher = {IEEE}
}

@article{engel2017direct,
  title = {Direct sparse odometry},
  author = {Engel, Jakob and Koltun, Vladlen and Cremers, Daniel},
  journal = PAMI,
  volume = 40,
  number = 3,
  pages = {611--625},
  year = 2017,
  publisher = {IEEE}
}

@inproceedings{newcombe2011dtam,
  title = {DTAM: Dense tracking and mapping in real-time},
  author = {Newcombe, Richard A and Lovegrove, Steven J and Davison, Andrew J},
  booktitle = ICCV,
  pages = {2320--2327},
  year = 2011,
  organization = {IEEE}
}

@inproceedings{NEURIPS2023,
 author = {Teed, Zachary and Lipson, Lahav and Deng, Jia},
 booktitle = NeurIPS,
 editor = {A. Oh and T. Naumann and A. Globerson and K. Saenko and M. Hardt and S. Levine},
 pages = {39033--39051},
 publisher = {Curran Associates, Inc.},
 title = {Deep Patch Visual Odometry},
 url = {https://proceedings.neurips.cc/paper_files/paper/2023/file/7ac484b0f1a1719ad5be9aa8c8455fbb-Paper-Conference.pdf},
 volume = 36,
 year = 2023
}

@article{cadena2017past,
  title = {Past, present, and future of simultaneous localization and mapping: Toward the robust-perception age},
  author = {Cadena, Cesar and Carlone, Luca and Carrillo, Henry and Latif, Yasir and Scaramuzza, Davide and Neira, Jos{\'e} and Reid, Ian and Leonard, John J},
  journal = {IEEE Transactions on robotics},
  volume = 32,
  number = 6,
  pages = {1309--1332},
  year = 2017,
  publisher = {IEEE}
}

@inproceedings{strasdat2011double,
  title = {Double window optimisation for constant time visual SLAM},
  author = {Strasdat, Hauke and Davison, Andrew J and Montiel, JM Mart{\`\i}nez and Konolige, Kurt},
  booktitle = ICCV,
  pages = {2352--2359},
  year = 2011,
  organization = {IEEE}
}

@inproceedings{zhu2022nice,
  title = {Nice-slam: Neural implicit scalable encoding for slam},
  author = {Zhu, Zihan and Peng, Songyou and Larsson, Viktor and Xu, Weiwei and Bao, Hujun and Cui, Zhaopeng and Oswald, Martin R and Pollefeys, Marc},
  booktitle = CVPR,
  pages = {12786--12796},
  year = 2022
}

@article{forster2016svo,
  title = {SVO: Semidirect visual odometry for monocular and multicamera systems},
  author = {Forster, Christian and Zhang, Zichao and Gassner, Michael and Werlberger, Manuel and Scaramuzza, Davide},
  journal = {IEEE Transactions on Robotics},
  volume = 33,
  number = 2,
  pages = {249--265},
  year = 2016,
  publisher = {IEEE}
}

@inproceedings{brachmann2017dsac,
  title = {Dsac-differentiable ransac for camera localization},
  author = {Brachmann, Eric and Krull, Alexander and Nowozin, Sebastian and Shotton, Jamie and Michel, Frank and Gumhold, Stefan and Rother, Carsten},
  booktitle = CVPR,
  pages = {6684--6692},
  year = 2017
}

@article{zhang2024monst3r,
  title = {Monst3r: A simple approach for estimating geometry in the presence of motion},
  author = {Zhang, Junyi and Herrmann, Charles and Hur, Junhwa and Jampani, Varun and Darrell, Trevor and Cole, Forrester and Sun, Deqing and Yang, Ming-Hsuan},
  journal = ICLR,
  year = 2024
}

@article{chen2025easi3r,
  title = {Easi3r: Estimating disentangled motion from dust3r without training},
  author = {Chen, Xingyu and Chen, Yue and Xiu, Yuliang and Geiger, Andreas and Chen, Anpei},
  journal = {arXiv preprint arXiv:2503.24391},
  year = 2025
}

@inproceedings{cabon2025must3r,
  title = {Must3r: Multi-view network for stereo 3d reconstruction},
  author = {Cabon, Yohann and Stoffl, Lucas and Antsfeld, Leonid and Csurka, Gabriela and Chidlovskii, Boris and Revaud, Jerome and Leroy, Vincent},
  booktitle = CVPR,
  pages = {1050--1060},
  year = 2025
}

@article{wang20243d,
  title = {3d reconstruction with spatial memory},
  author = {Wang, Hengyi and Agapito, Lourdes},
  journal = {arXiv preprint arXiv:2408.16061},
  year = 2024
}

@article{wu2025point3r,
  title = {Point3R: Streaming 3D Reconstruction with Explicit Spatial Pointer Memory},
  author = {Wu, Yuqi and Zheng, Wenzhao and Zhou, Jie and Lu, Jiwen},
  journal = {arXiv preprint arXiv:2507.02863},
  year = 2025
}

@article{shen2025mut3r,
  title = {MUT3R: Motion-aware Updating Transformer for Dynamic 3D Reconstruction},
  author = {Shen, Guole and Deng, Tianchen and Qin, Xingrui and Wang, Nailin and Wang, Jianyu and Wang, Yanbo and Chen, Yongtao and Wang, Hesheng and Wang, Jingchuan},
  journal = {arXiv preprint arXiv:2512.03939},
  year = 2025
}

@inproceedings{sun2020test,
  title = {Test-time training with self-supervision for generalization under distribution shifts},
  author = {Sun, Yu and Wang, Xiaolong and Liu, Zhuang and Miller, John and Efros, Alexei and Hardt, Moritz},
  booktitle = ICML,
  pages = {9229--9248},
  year = 2020,
  organization = {PMLR}
}

@article{behrouz2024titans,
  title = {Titans: Learning to memorize at test time},
  author = {Behrouz, Ali and Zhong, Peilin and Mirrokni, Vahab},
  journal = {arXiv preprint arXiv:2501.00663},
  year = 2024
}

@inproceedings{gu2024mamba,
  title = {Mamba: Linear-time sequence modeling with selective state spaces},
  author = {Gu, Albert and Dao, Tri},
  booktitle = {First conference on language modeling},
  year = 2024
}

@article{su2024roformer,
  title = {Roformer: Enhanced transformer with rotary position embedding},
  author = {Su, Jianlin and Ahmed, Murtadha and Lu, Yu and Pan, Shengfeng and Bo, Wen and Liu, Yunfeng},
  journal = {Neurocomputing},
  volume = 568,
  pages = {127063},
  year = 2024,
  publisher = {Elsevier}
}

@article{zhang2023h2o,
  title = {H2o: Heavy-hitter oracle for efficient generative inference of large language models},
  author = {Zhang, Zhenyu and Sheng, Ying and Zhou, Tianyi and Chen, Tianlong and Zheng, Lianmin and Cai, Ruisi and Song, Zhao and Tian, Yuandong and R{\'e}, Christopher and Barrett, Clark and others},
  journal = NeurIPS,
  volume = 36,
  pages = {34661--34710},
  year = 2023
}

@article{weinzaepfel2022croco,
  title = {Croco: Self-supervised pre-training for 3d vision tasks by cross-view completion},
  author = {Weinzaepfel, Philippe and Leroy, Vincent and Lucas, Thomas and Br{\'e}gier, Romain and Cabon, Yohann and Arora, Vaibhav and Antsfeld, Leonid and Chidlovskii, Boris and Csurka, Gabriela and Revaud, J{\'e}r{\^o}me},
  journal = NeurIPS,
  volume = 35,
  pages = {3502--3516},
  year = 2022
}

@InProceedings{Yin_2019_ICCV,
author = {Yin, Wei and Liu, Yifan and Shen, Chunhua and Yan, Youliang},
title = {Enforcing Geometric Constraints of Virtual Normal for Depth Prediction},
booktitle = ICCV,
month = {October},
year = 2019
}

@article{bochkovskii2410depth,
  title = {Depth Pro: Sharp monocular metric depth in less than a second.},
  author = {Bochkovskii, Aleksei and Delaunoy, Ama{\"e}l and Germain, Hugo and Santos, Marcel and Zhou, Yichao and Richter, Stephan R and Koltun, Vladlen},
  journal = {arXiv preprint arXiv:2410.02073}
}

@inproceedings{kopf2021robust,
  title = {Robust consistent video depth estimation},
  author = {Kopf, Johannes and Rong, Xuejian and Huang, Jia-Bin},
  booktitle = CVPR,
  pages = {1611--1621},
  year = 2021
}

@inproceedings{wang2023neural,
  title = {Neural video depth stabilizer},
  author = {Wang, Yiran and Shi, Min and Li, Jiaqi and Huang, Zihao and Cao, Zhiguo and Zhang, Jianming and Xian, Ke and Lin, Guosheng},
  booktitle = ICCV,
  pages = {9466--9476},
  year = 2023
}

@inproceedings{gedraite2011investigation,
  title = {Investigation on the effect of a Gaussian Blur in image filtering and segmentation},
  author = {Gedraite, Estev{\~a}o S and Hadad, Murielle},
  booktitle = {Proceedings ELMAR-2011},
  pages = {393--396},
  year = 2011,
  organization = {IEEE}
}

@article{perona2002scale,
  title = {Scale-space and edge detection using anisotropic diffusion},
  author = {Perona, Pietro and Malik, Jitendra},
  journal = PAMI,
  volume = 12,
  number = 7,
  pages = {629--639},
  year = 2002,
  publisher = {IEEE}
}

@article{rudin1992nonlinear,
  title={Nonlinear total variation based noise removal algorithms},
  author={Rudin, Leonid I and Osher, Stanley and Fatemi, Emad},
  journal={Physica D: nonlinear phenomena},
  volume = 60,
  number = {1-4},
  pages = {259--268},
  year = 1992,
  publisher = {Elsevier}
}

@inproceedings{paris2006fast,
  title = {A fast approximation of the bilateral filter using a signal processing approach},
  author = {Paris, Sylvain and Durand, Fr{\'e}do},
  booktitle = ECCV,
  pages = {568--580},
  year = 2006,
  organization = {Springer}
}

@article{digne2017bilateral,
  title = {The bilateral filter for point clouds},
  author = {Digne, Julie and De Franchis, Carlo},
  journal = {Image Processing On Line},
  volume = 7,
  pages = {278--287},
  year = 2017
}

@incollection{fleishman2003bilateral,
  title = {Bilateral mesh denoising},
  author = {Fleishman, Shachar and Drori, Iddo and Cohen-Or, Daniel},
  booktitle = {ACM SIGGRAPH 2003 Papers},
  pages = {950--953},
  year = 2003
}

@inproceedings{barron2016fast,
  title = {The fast bilateral solver},
  author = {Barron, Jonathan T and Poole, Ben},
  booktitle = ECCV,
  pages = {617--632},
  year = 2016,
  organization = {Springer}
}

@article{zhang2021bilateral,
  title = {Bilateral attention network for RGB-D salient object detection},
  author = {Zhang, Zhao and Lin, Zheng and Xu, Jun and Jin, Wen-Da and Lu, Shao-Ping and Fan, Deng-Ping},
  journal = TIP,
  volume = 30,
  pages = {1949--1961},
  year = 2021,
  publisher = {IEEE}
}

@inproceedings{yeshwanth2023scannet++,
  title={Scannet++: A high-fidelity dataset of 3d indoor scenes},
  author={Yeshwanth, Chandan and Liu, Yueh-Cheng and Nie{\ss}ner, Matthias and Dai, Angela},
  booktitle = CVPR,
  pages = {12--22},
  year = 2023
}

@inproceedings{wang2020tartanair,
  title={Tartanair: A dataset to push the limits of visual slam},
  author={Wang, Wenshan and Zhu, Delong and Wang, Xiangwei and Hu, Yaoyu and Qiu, Yuheng and Wang, Chen and Hu, Yafei and Kapoor, Ashish and Scherer, Sebastian},
  booktitle={2020 IEEE/RSJ International Conference on Intelligent Robots and Systems (IROS)},
  pages={4909--4916},
  year={2020},
  organization={IEEE}
}

@inproceedings{sun2020scalability,
  title = {Scalability in perception for autonomous driving: Waymo open dataset},
  author = {Sun, Pei and Kretzschmar, Henrik and Dotiwalla, Xerxes and Chouard, Aurelien and Patnaik, Vijaysai and Tsui, Paul and Guo, James and Zhou, Yin and Chai, Yuning and Caine, Benjamin and others},
  booktitle = CVPR,
  pages = {2446--2454},
  year = 2020
}

@article{sun2024learning,
  title = {Learning to (learn at test time): Rnns with expressive hidden states},
  author = {Sun, Yu and Li, Xinhao and Dalal, Karan and Xu, Jiarui and Vikram, Arjun and Zhang, Genghan and Dubois, Yann and Chen, Xinlei and Wang, Xiaolong and Koyejo, Sanmi and others},
  journal = {arXiv preprint arXiv:2407.04620},
  year = 2024
}

@article{cai2025mem4d,
  title = {Mem4d: Decoupling static and dynamic memory for dynamic scene reconstruction},
  author = {Cai, Xudong and Wang, Shuo and Wang, Peng and Wang, Yongcai and Fan, Zhaoxin and Li, Wanting and Zhang, Tianbao and Tao, Jianrong and Jin, Yeying and Li, Deying},
  journal = {arXiv preprint arXiv:2508.07908},
  year = 2025
}

@inproceedings{dai2017scannet,
  title = {Scannet: Richly-annotated 3d reconstructions of indoor scenes},
  author = {Dai, Angela and Chang, Angel X and Savva, Manolis and Halber, Maciej and Funkhouser, Thomas and Nie{\ss}ner, Matthias},
  booktitle = CVPR,
  pages = {5828--5839},
  year = 2017
}

@inproceedings{2012naturalistic,
  title = {A naturalistic open source movie for optical flow evaluation},
  author = {Butler, Daniel J and Wulff, Jonas and Stanley, Garrett B and Black, Michael J},
  booktitle = ECCV,
  pages = {611--625},
  year = 2012,
  organization={Springer}
}

@inproceedings{2012benchmark,
  title = {A benchmark for the evaluation of RGB-D SLAM systems},
  author = {Sturm, J{\"u}rgen and Engelhard, Nikolas and Endres, Felix and Burgard, Wolfram and Cremers, Daniel},
  booktitle = {2012 IEEE/RSJ international conference on intelligent robots and systems (IROS)},
  pages = {573--580},
  year = 2012,
  organization={IEEE}
}

@inproceedings{2019refusion,
  title = {ReFusion: 3D reconstruction in dynamic environments for RGB-D cameras exploiting residuals},
  author = {Palazzolo, Emanuele and Behley, Jens and Lottes, Philipp and Giguere, Philippe and Stachniss, Cyrill},
  booktitle = {2019 IEEE/RSJ International Conference on Intelligent Robots and Systems (IROS)},
  pages = {7855--7862},
  year = 2019,
  organization = {IEEE}
}

@article{2013vision,
  title = {Vision meets robotics: The kitti dataset},
  author = {Geiger, Andreas and Lenz, Philip and Stiller, Christoph and Urtasun, Raquel},
  journal = {The international journal of robotics research},
  volume = 32,
  number = 11,
  pages={1231--1237},
  year=2013,
  publisher={Sage Publications Sage UK: London, England}
}

@inproceedings{2013scene,
  title = {Scene coordinate regression forests for camera relocalization in RGB-D images},
  author = {Shotton, Jamie and Glocker, Ben and Zach, Christopher and Izadi, Shahram and Criminisi, Antonio and Fitzgibbon, Andrew},
  booktitle = CVPR,
  pages = {2930--2937},
  year = 2013
}

@article{loshchilov2017decoupled,
  title={Decoupled weight decay regularization},
  author={Loshchilov, Ilya and Hutter, Frank},
  journal={arXiv preprint arXiv:1711.05101},
  year = 2017
}

\appendix

\section{Implementation details of Approach}

\subsection{More details on trajectory-consistent model optimization}

During trajectory-consistent model optimization, we utilized a mixture of three datasets  to cover both indoor and outdoor scenarios: the Scannet++ dataset \cite{yeshwanth2023scannet++},
the Waymo dataset \cite{sun2020scalability}, and the TartanAir dataset \cite{wang2020tartanair}, with a mixing ratio of 5:3:2. There is no overlap between the fine-tuning data and the test sets used in our experiments. The image resolution was set to 512 × 384, and the maximum number 
of frames per sample was 64. The network parameters were initialized using 
the ViT-Large model with pre-trained CUT3R \cite{wang2025continuous} weights and optimized using the 
AdamW optimizer \cite{loshchilov2017decoupled}. The initial learning rate was set to 1e-7 with a maximum 
learning rate of 1e-6, employing a linear warm-up and cosine decay strategy. 
The entire training process was conducted on 8 RTX 4090 GPUs.

\subsection{More details on online stabilization module}

\noindent This section provides comprehensive implementation details for the online stabilization module introduced in \secref{ssec:st-stabilization} of the main paper. To ensure the reproducibility of our research, we elaborate on the formal mathematical derivations and specific parameter configurations of this module. Specifically, the smoothing factor $\alpha_{i-1}$ in \cref{filtering} of the main paper is formulated as follows: 
\begin{equation}
\label{smoothing factor}
\alpha_{i-1} = \frac{2\pi \cdot f_{i-1} \cdot t'_{i-1}}{2\pi \cdot f_{i-1} \cdot t'_{i-1} + 1}
\end{equation}
\noindent where $t'_{i-1}$ represents the time interval between the $i$-th frame timestamp $t_i$ and the $(i-1)$-th frame timestamp $t_{i-1}$. $f_{i-1}$ denotes the cutoff frequency that dictates the magnitude of the smoothing factor for the $(i-1)$-th frame, which is mathematically expressed by:
\begin{equation}
\label{frame}
f_{i-1} = f_{\min} + \beta \cdot |v_{i-1}|
\end{equation}
\noindent where $f_{\min}$ is the minimum cutoff frequency. A smaller value of $f_{\min}$ results in a more stable trajectory when the camera is stationary but introduces higher latency during rapid camera movement; conversely, a larger value reduces latency at the cost of increased trajectory jitter. $\beta$ serves as the dynamic gain for the cutoff frequency $f_{i-1}$, and its impact on the trajectory and latency is similar to that of $f_{\min}$. $v_{i-1}$ denotes the instantaneous velocity from frame $i-1$ to frame $i$, calculated as the difference between the predicted translation of frame $i$ ($\bm{\hat{x}}_i$) and the processed translation of frame $i-1$ ($\bm{x}_{i-1}$), divided by the time interval $t'_{i-1}$. Additionally, for the Slerp function employed in \cref{smoothed quaternions} of the main paper, its formal definition is given by:
\begin{equation}
\label{Spherical Linear Interpolation}
    \begin{array}{cc}
    Slerp(\mathbf{a}, \mathbf{b}; \gamma) = \frac{\sin((1-\gamma)\theta)}{\sin \theta} \mathbf{a} + \frac{\sin(\gamma \theta)}{\sin \theta} \mathbf{b} \\
    \cos \theta = \mathbf{a} \cdot \mathbf{b}
    \end{array}
\end{equation}
\noindent where the parameter $\gamma$ represents the interpolation factor, ranging from 0 to 1. In addition, the specific process of spatial geometry refinement is as follows: For any predicted point $\hat{p}$ within the point cloud, its processed position $p$ is obtained through a weighted average of points $\hat{q}$ within its neighborhood $S$:
\begin{equation}
\label{smoothed position}
    p = \frac{1}{W_p} \sum_{\hat{q} \in S} G_{\sigma_s} (\|\hat{p} - \hat{q}\|) \cdot G_{\sigma_r} (|D_p - D_q|) \cdot D_p
\end{equation}
\noindent $G_{\sigma_s}$ represents the spatial proximity factor, where a smaller distance between $\hat{p}$ and $\hat{q}$ results in a higher weight. $G_{\sigma_r}$ acts as the geometric similarity factor, measuring the depth difference between $\hat{p}$ and $\hat{q}$. When points within the neighborhood lie on opposite sides of an edge, the weights decrease rapidly toward zero. This mechanism effectively prevents mutual interference between data on different sides of an edge.

\section{More Results and Comparison} 
To further verify the reliability of our proposed method, we provide additional experimental results on camera pose estimation (\secref{Camera Pose Estimation}), depth evaluation (\secref{DEPTH ESTIMATION}), and 3D reconstruction quality (\secref{3D RECONSTRUCTION}). Furthermore, we present a comprehensive comparison between our method and state-of-the-art online reconstruction approaches in terms of GPU memory consumption and frame rate (FPS) (\secref{GPU Usage And FPS Estimation}). In addition, it should be noted that the code of Mem4D \cite{cai2025mem4d} has not been made public yet, so all the results of Mem4D mentioned come from its original paper. Since CUT3R samples the TUM dataset\cite{2012benchmark} at intervals, we slightly adjusted the time interval parameter of PAS3R when evaluating its camera pose performance under CUT3R settings in main paper.

\subsection{Camera Pose Estimation} \label{Camera Pose Estimation}

\begin{figure}[tb]
  \centering
  \begin{subfigure}{0.32\linewidth}
    \centering
    \includegraphics[width=\linewidth]{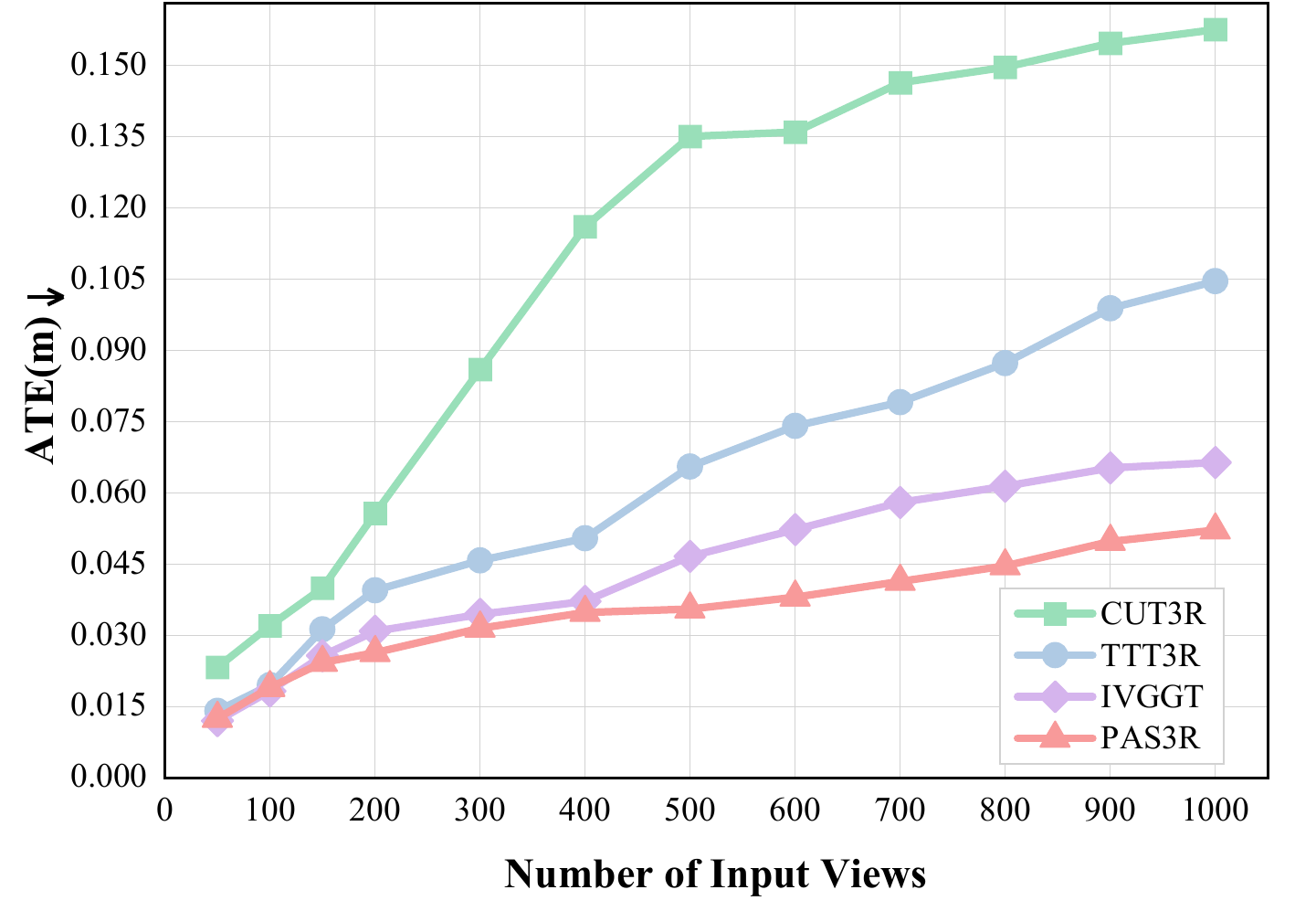}
    \caption{ATE on TUM}
    \label{tum-cp-a}
  \end{subfigure}
  \hfill 
  \begin{subfigure}{0.32\linewidth}
    \centering
    \includegraphics[width=\linewidth]{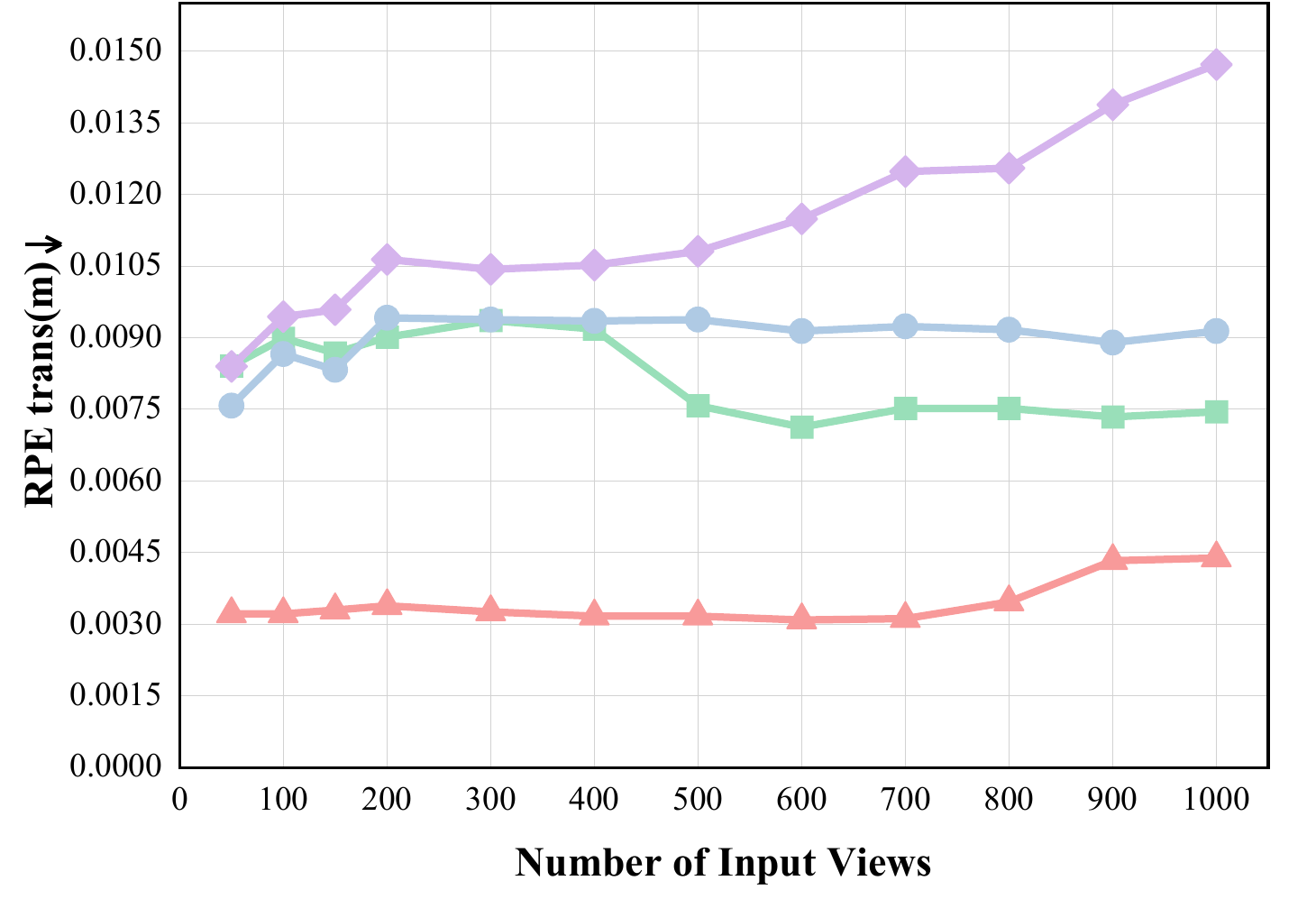}
    \caption{RPE trans on TUM}
    \label{tum-cp-b}
  \end{subfigure}
  \hfill 
  \begin{subfigure}{0.32\linewidth}
    \centering
    \includegraphics[width=\linewidth]{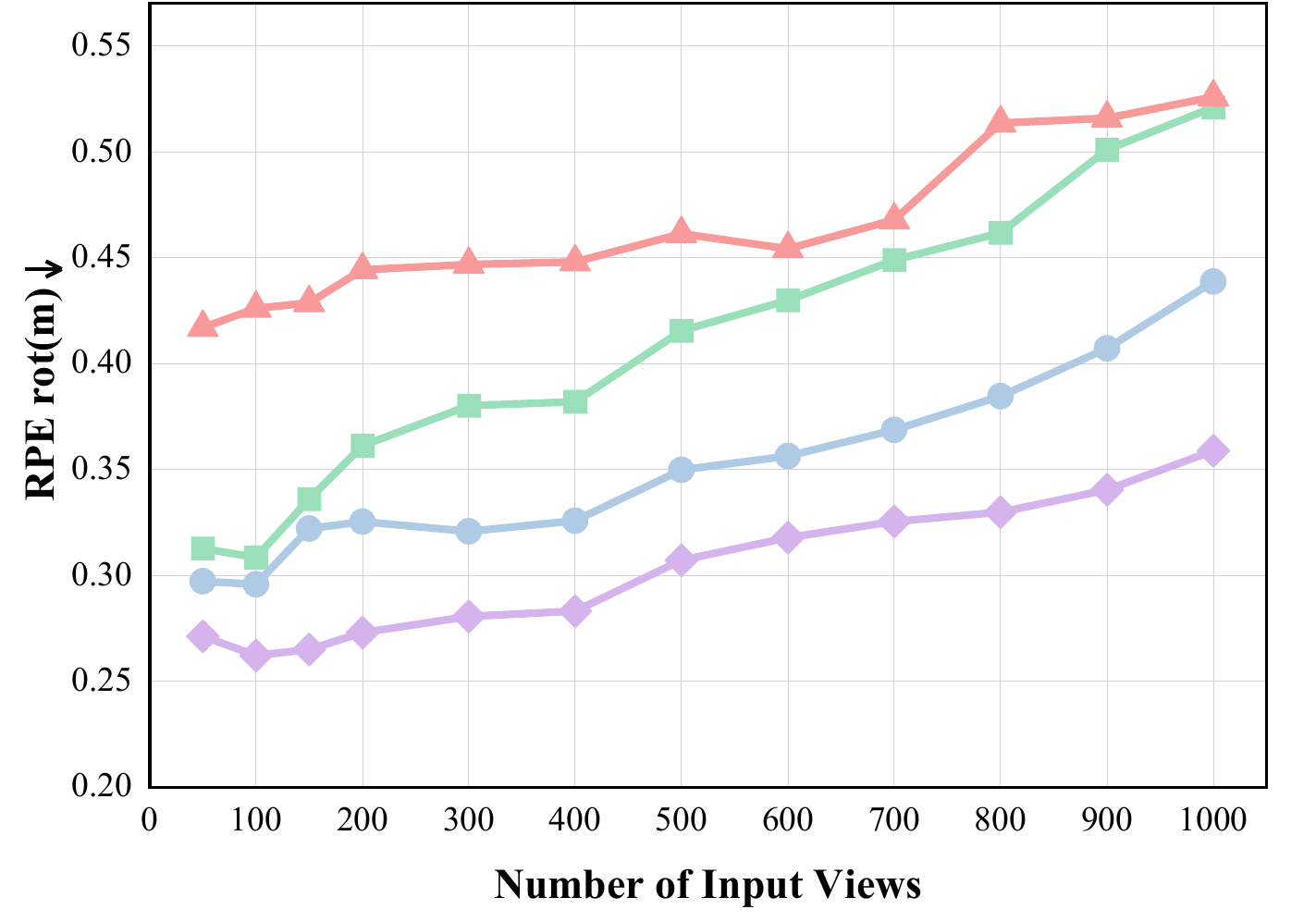}
    \caption{RPE rot on TUM}
    \label{tum-cp-c}
  \end{subfigure}
  \caption{Comparison of camera pose estimation on TUM \cite{2012benchmark}}
  \label{tum-cp}
\end{figure}

We also evaluate camera pose estimation performance on TUM, whose sequences range from the first 50 to 1000 frames for each scene. The quantitative results are shown in \cref{tum-cp}.
Experimental results demonstrate that though the RPE rot of PAS3R is slightly lower than that of competing methods on TUM, it exhibits a substantial, near-categorical lead over CUT3R, TTT3R \cite{anonymous2026tttr}, and Infinite VGGT (IVGGT)\cite{yuan2026infinitevggt} in terms of ATE and RPE trans. This performance gap is particularly pronounced in long video sequences. 

\begin{figure}[t]
    \captionsetup[subfigure]{justification=centering}
    \centering
    \begin{subfigure}[b]{0.24\textwidth}
        \centering
        \includegraphics[width=\textwidth]{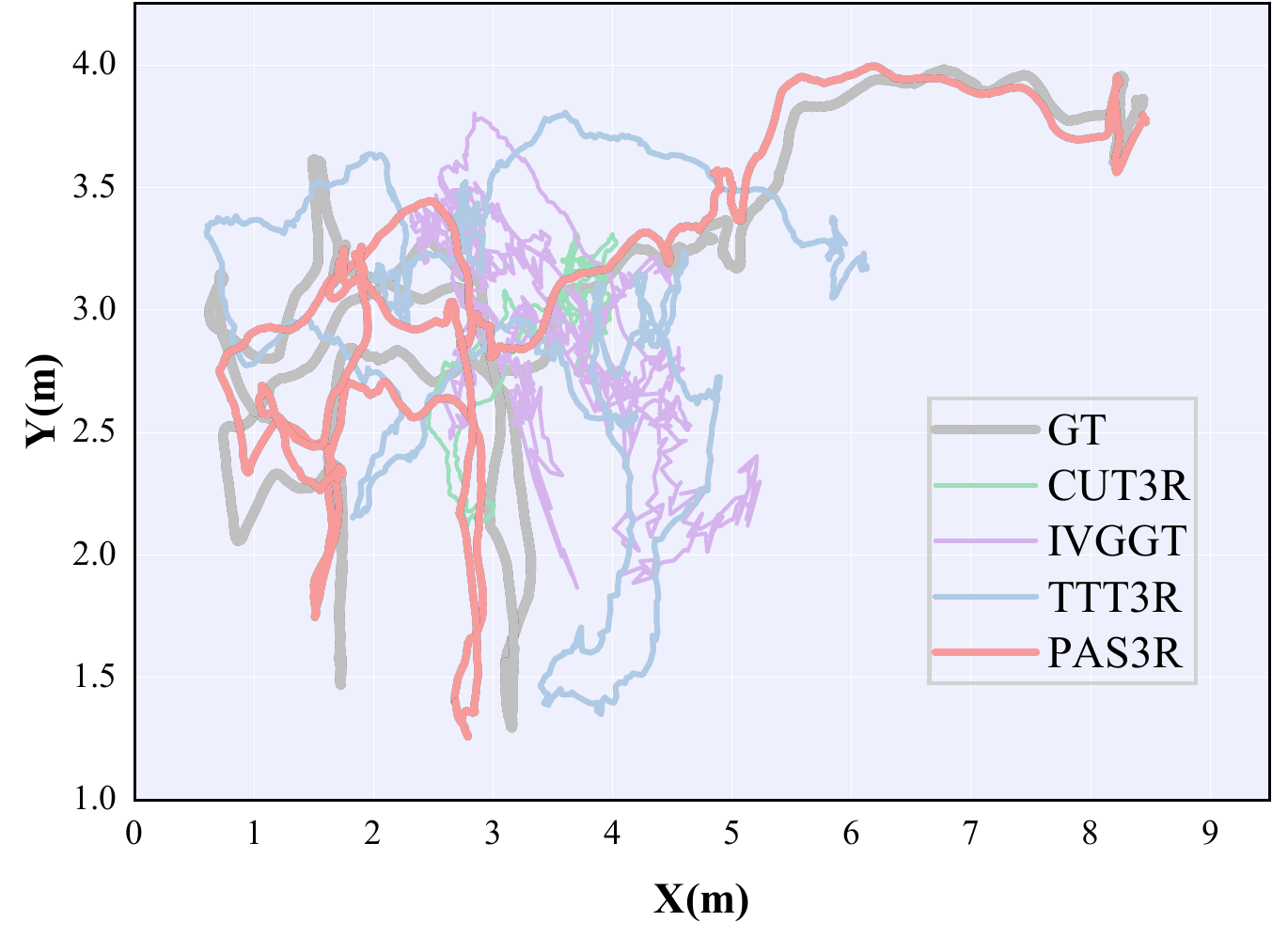}
        \caption{Scene 721 on Scannet}
        \label{traj-a}
    \end{subfigure}
    \hfill
    \begin{subfigure}[b]{0.24\textwidth}
        \centering
        \includegraphics[width=\textwidth]{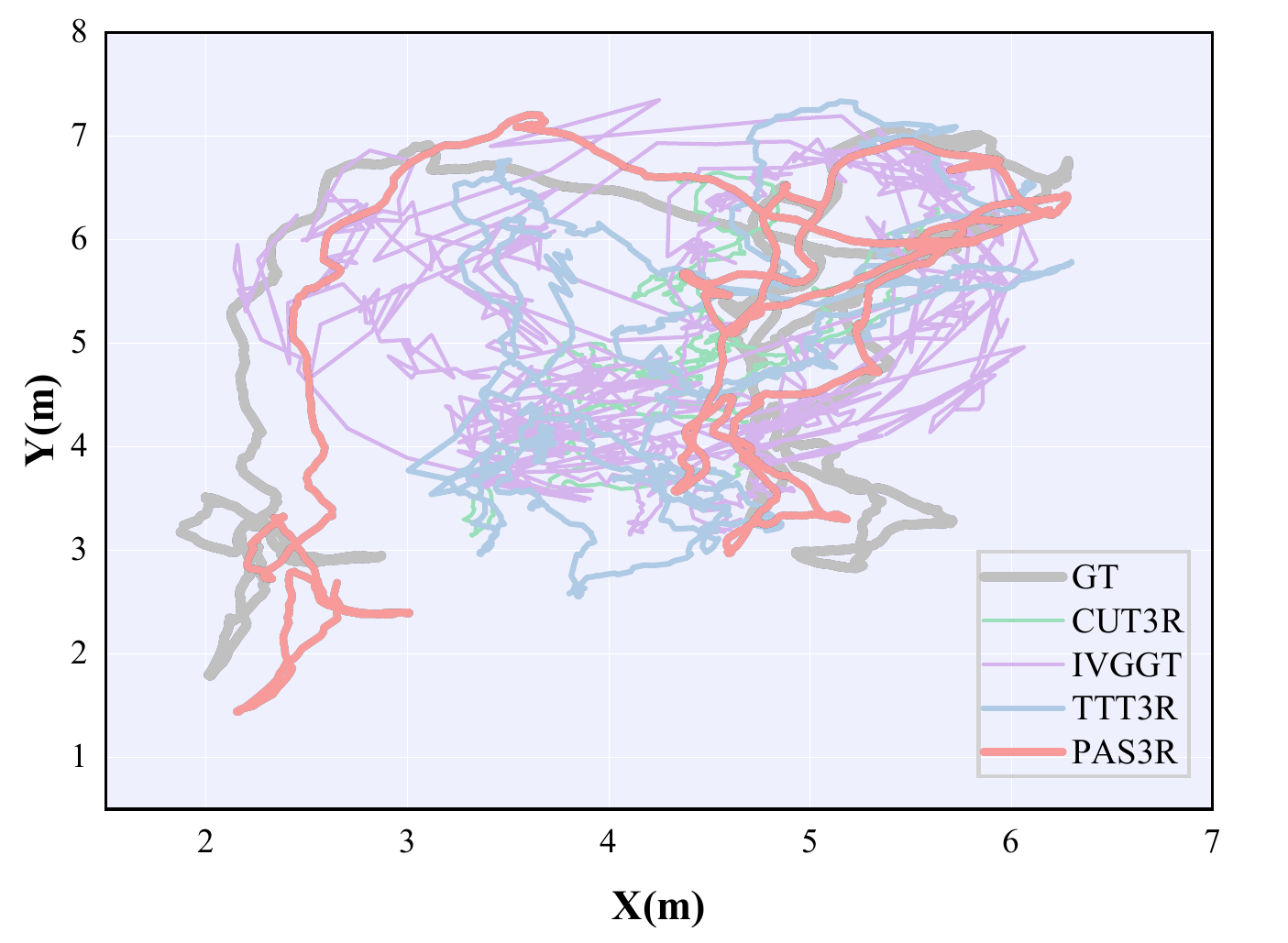}
        \caption{Scene 793 on Scannet}
        \label{traj-b}
    \end{subfigure}
    \begin{subfigure}[b]{0.24\textwidth}
        \centering
        \includegraphics[width=\textwidth]{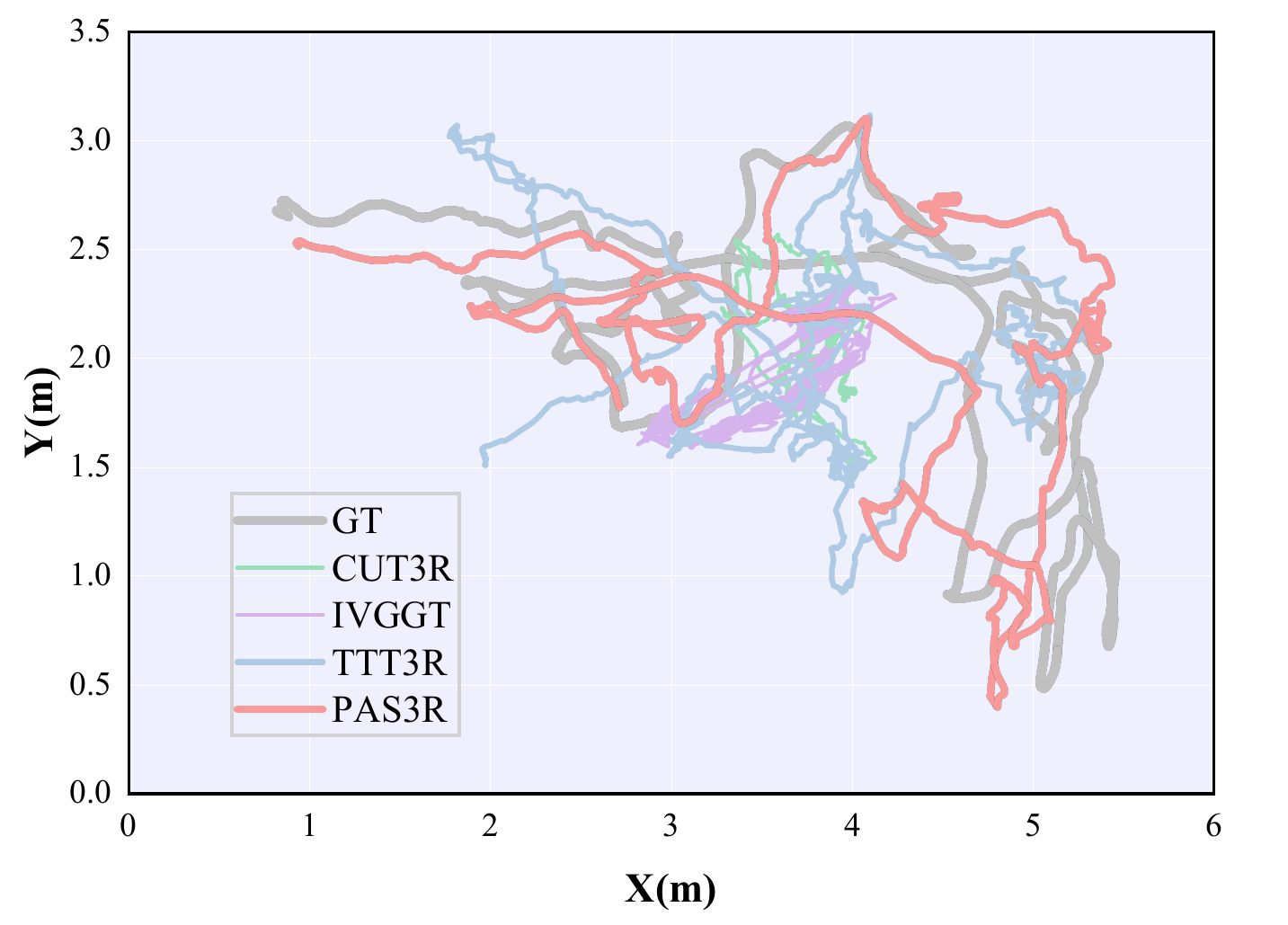}
        \caption{Scene 766 on Scannet}
        \label{traj-c}
    \end{subfigure}
    \hfill
    \begin{subfigure}[b]{0.23\textwidth}
        \centering
        \includegraphics[width=\textwidth]{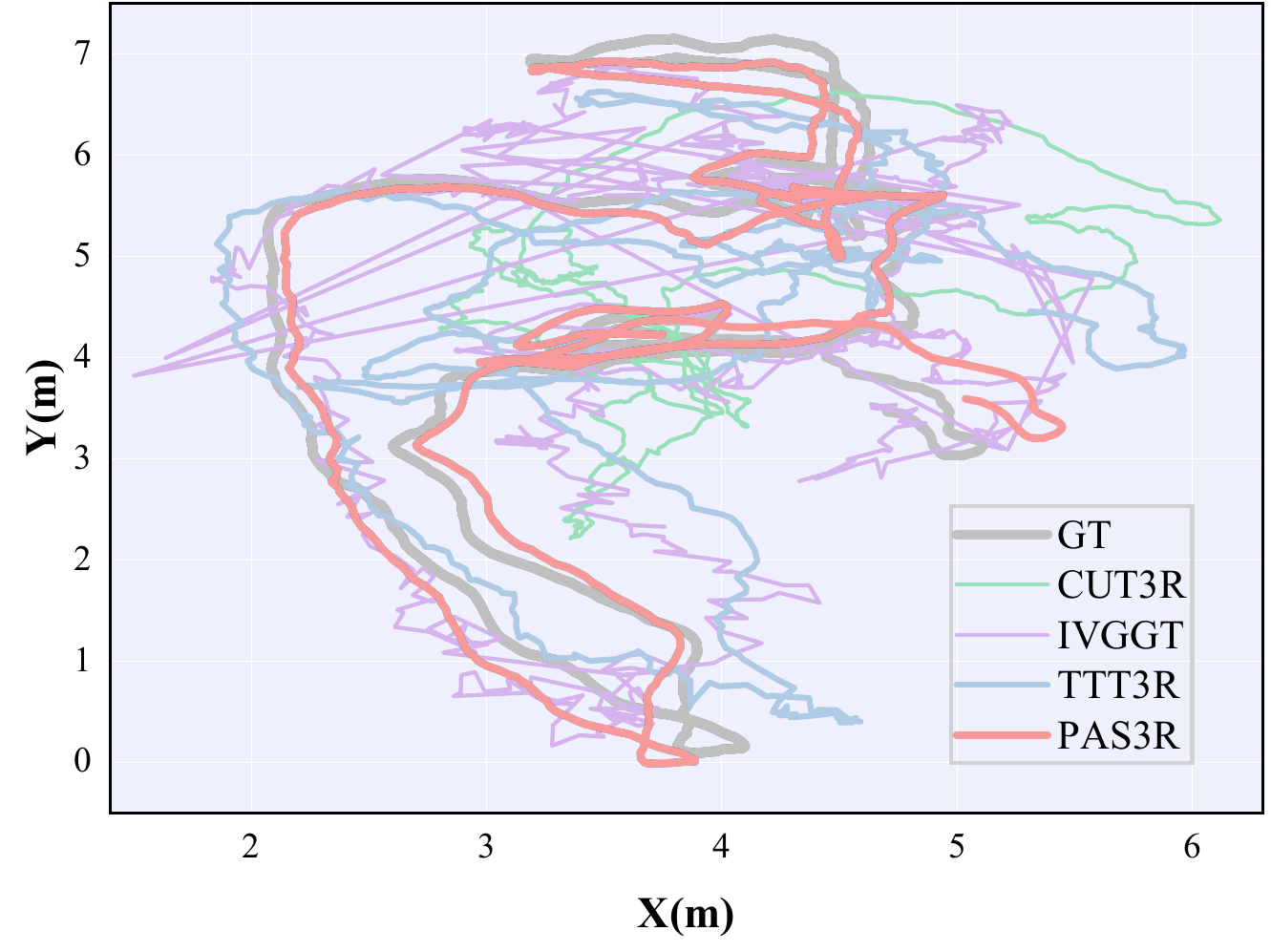}
        \caption{Scene 781 on Scannet}
        \label{traj-d}
    \end{subfigure}
    \hfill
    \caption{Comparison of camera trajectories on Scannet \cite{dai2017scannet}}
    \label{traj}
    \vspace{-5mm}
\end{figure}

In addition, we perform visualization of the alignment between the predicted camera trajectories and the ground-truth trajectories on Scannet dataset \cite{dai2017scannet}. The results validate that PAS3R achieves the optimal trajectory fitting performance in most scenarios, and we present partial representative results in \cref{traj}.

\subsection{DEPTH ESTIMATION} \label{DEPTH ESTIMATION}

\begin{figure}[tb]
  \captionsetup[subfigure]{justification=centering}
  \centering
  \begin{subfigure}{0.32\linewidth}
    \centering
    \includegraphics[width=\linewidth]{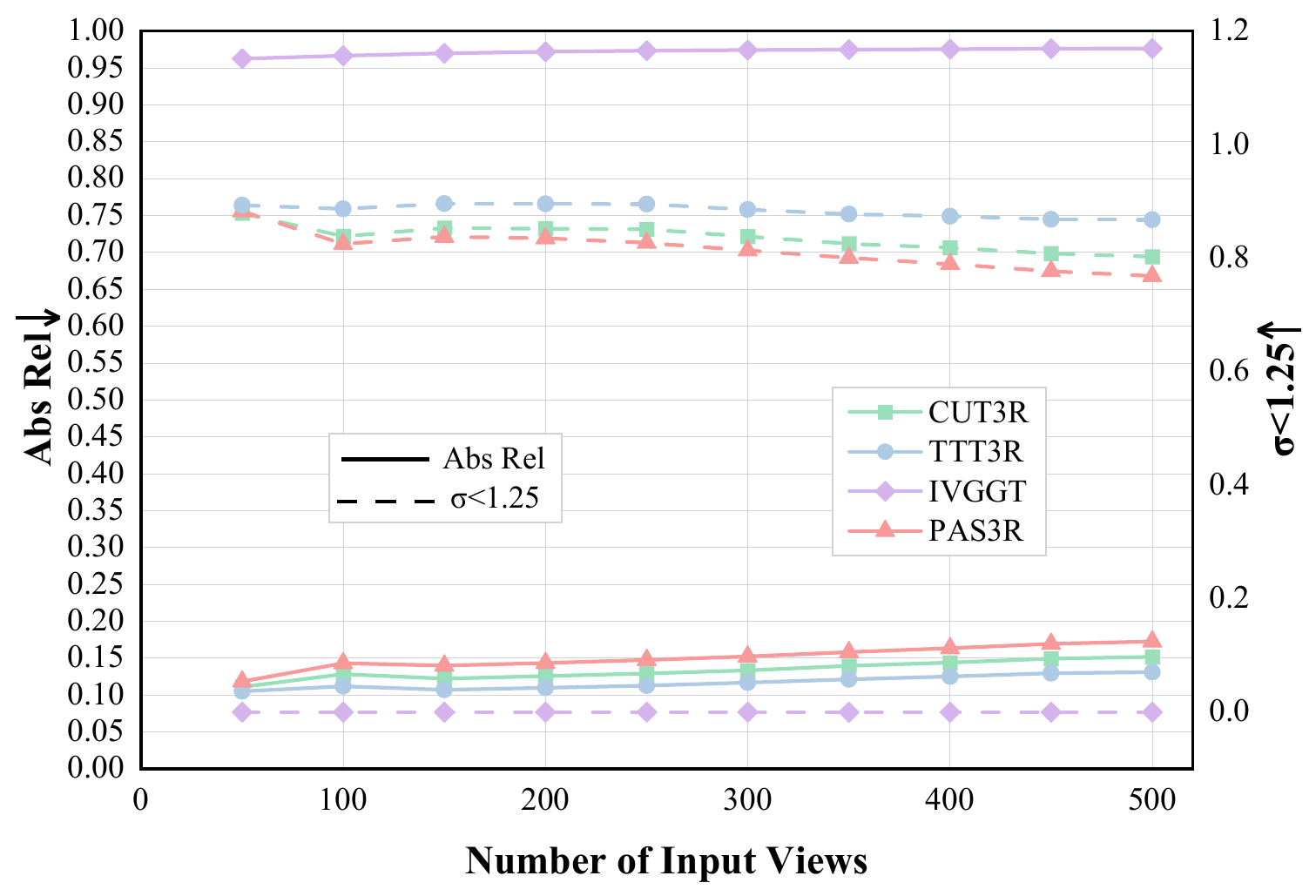}
    \caption{Depth evaluation on Kitti(oringinal)}
    \label{kitti-depth-a}
  \end{subfigure}
  \hfill 
  \begin{subfigure}{0.32\linewidth}
    \centering
    \includegraphics[width=\linewidth]{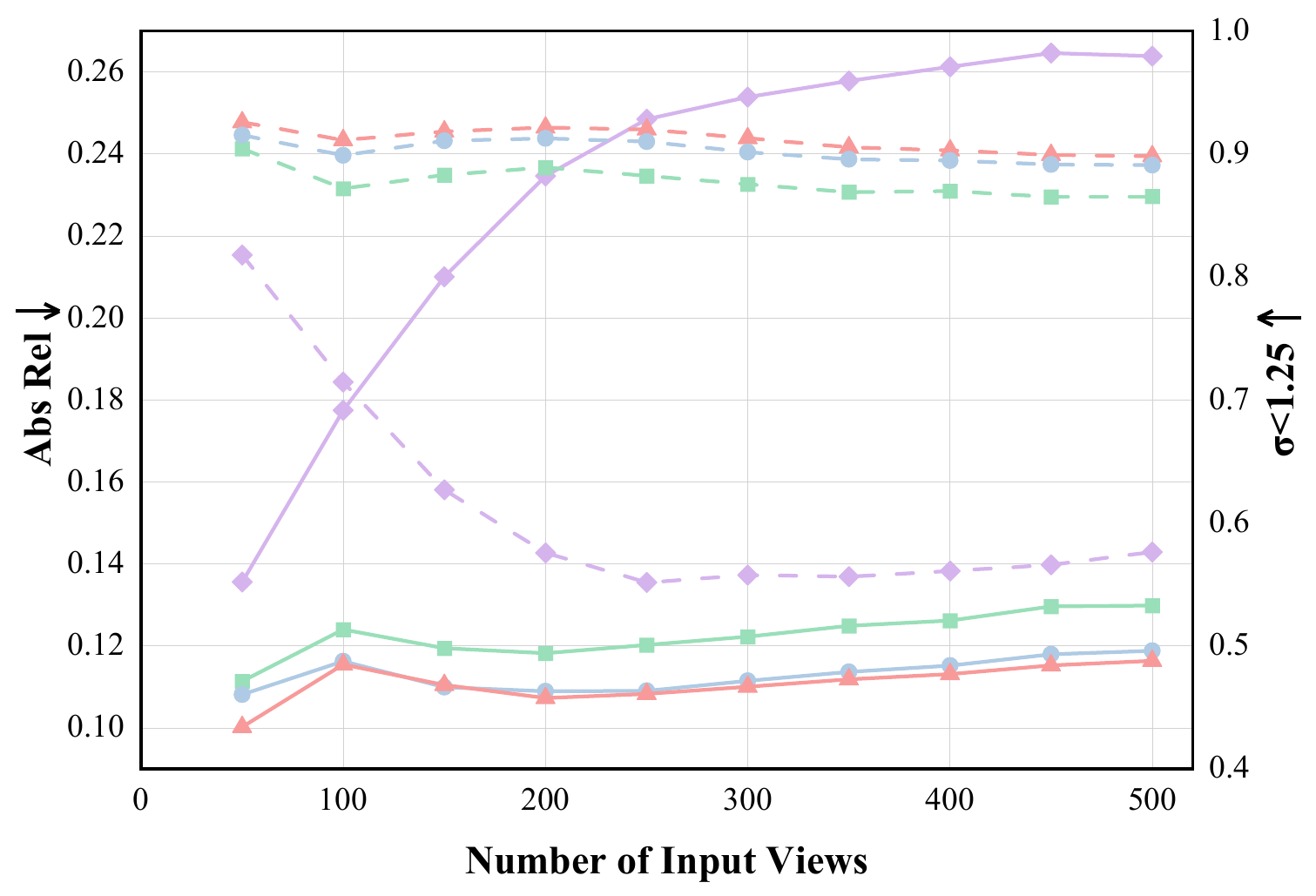}
    \caption{Depth evaluation on Kitti(Scale)}
    \label{kitti-depth-b}
  \end{subfigure}
  \hfill 
  \begin{subfigure}{0.32\linewidth}
    \centering
    \includegraphics[width=\linewidth]{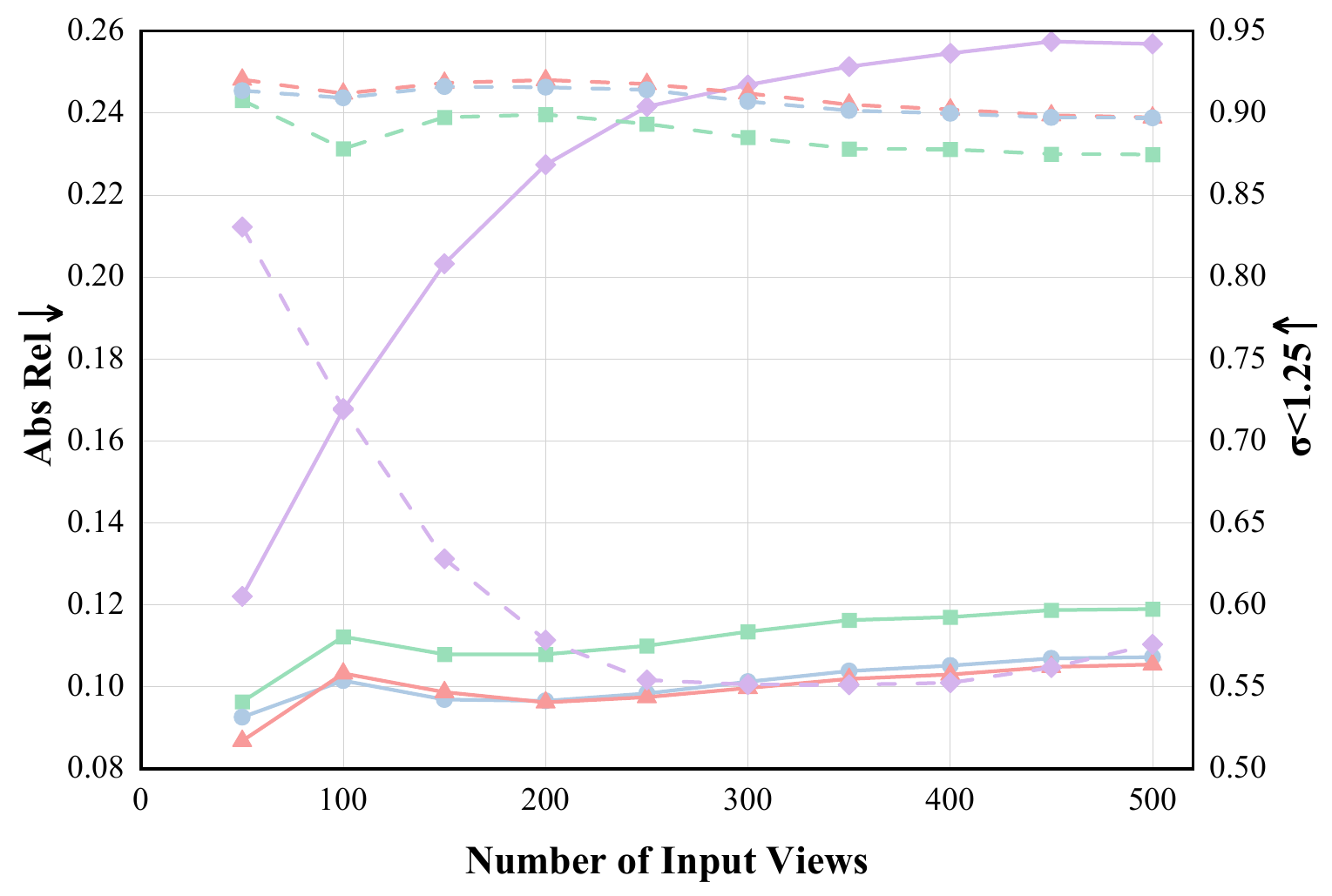}
    \caption{Depth evaluation on Kitti(Scale\&Shift)}
    \label{kitti-depth-c}
  \end{subfigure}
  \caption{Comparison of depth on Kitti \cite{2013vision}}
  \label{kitti-depth}
\end{figure}

We further evaluate the depth prediction accuracy on Kitti \cite{2013vision}, with sequence lengths from 50 to 500 frames, as shown in \cref{kitti-depth}.
Regarding the evaluation on the Kitti dataset, PAS3R shows slightly lower performance than CUT3R and TTT3R under the original setting. However, it consistently achieves the best performance under both scale-aligned and scale-and-shift-aligned evaluations. In comparison, IVGGT demonstrates relatively lower performance across these evaluation settings on this dataset. Overall, these results suggest that PAS3R maintains competitive and stable depth estimation performance compared with current state-of-the-art online reconstruction methods.

\subsection{3D RECONSTRUCTION} \label{3D RECONSTRUCTION}

While our model is mainly tailored for long sequences, we also seek to validate its competitive performance on short sequences. To this end, following the experimental setup of CUT3R, we evaluate the 3D point cloud reconstruction quality of our model on two relatively short-sequence benchmarks, namely the 7Scene dataset \cite{2013scene} and the NRGBD dataset \cite{2013scene}, and conduct a comprehensive comparison with state-of-the-art online methods. The quantitative comparison results are summarized in \cref{comparison_7scenes_nrgbd}.
\begin{table}[htbp]
    \centering
    \caption{Quantitative comparison on 7-Scenes and NRGBD datasets \cite{2013scene}. \textbf{Bold} indicates the best performance, and \underline{underlined} indicates the second best. Acc (Accuracy) and Comp (Completeness) are error metrics, where $\downarrow$ indicates lower is better.}
    \label{comparison_7scenes_nrgbd}
    \renewcommand{\arraystretch}{1.2}
    \setlength{\tabcolsep}{6pt}
    \begin{tabular}{lcccc}
        \toprule
        \multirow{2}{*}{\textbf{Method}} & \multicolumn{2}{c}{\textbf{7-Scenes} \cite{2013scene}} & \multicolumn{2}{c}{\textbf{NRGBD} \cite{2013scene}} \\
        \cmidrule(lr){2-3} \cmidrule(lr){4-5}
         & Acc $\downarrow$ & Comp $\downarrow$ & Acc $\downarrow$ & Comp $\downarrow$ \\
        \midrule
        Mem4D\cite{cai2025mem4d} & 0.185 & 0.178 & 0.271 & 0.212 \\
        CUT3R\cite{wang2025continuous} & \textbf{0.120} & \underline{0.156} & \textbf{0.100} & \textbf{0.078} \\
        TTT3R\cite{anonymous2026tttr} & 0.156 & 0.196 & 0.105 & \underline{0.080} \\
        IVGGT\cite{yuan2026infinitevggt} & \underline{0.124} & \textbf{0.106} & 0.110 & 0.109 \\
        Ours  & 0.151 & 0.182 & \underline{0.103} & 0.087 \\
        \bottomrule
    \end{tabular}
\end{table}

\begin{figure*}[t]
    \centering
    \begin{tikzpicture}
        \node[anchor=south west, inner sep=0] (pic) at (0,0) {\includegraphics[width=0.9\linewidth]{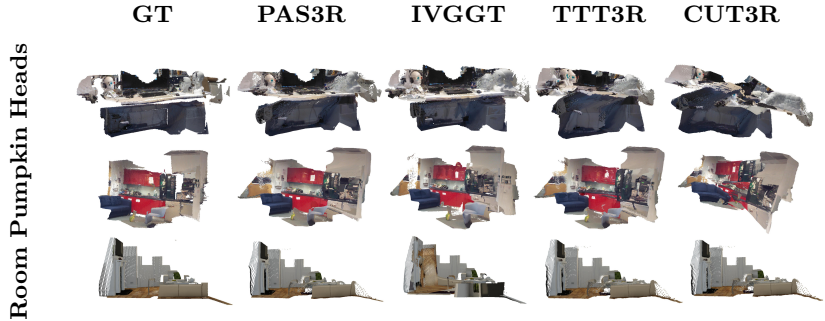}};
        \begin{scope}[x={(pic.south east)}, y={(pic.north west)}]
            \tikzset{header/.style={
                font=\small\bfseries, 
                inner sep=0pt, 
                text height=1.5ex, 
                text depth=0.25ex
            }}

            \node[header, anchor=south] at (0.14, 1.01) {GT};
            \node[header, anchor=south] at (0.32, 1.01) {PAS3R};
            \node[header, anchor=south] at (0.51, 1.01) {IVGGT};
            \node[header, anchor=south] at (0.68, 1.01) {TTT3R};
            \node[header, anchor=south] at (0.84, 1.01) {CUT3R};

            \node[header, rotate=90, anchor=center] at (-0.02, 0.83) {Heads};
            \node[header, rotate=90, anchor=center] at (-0.02, 0.50) {Pumpkin};
            \node[header, rotate=90, anchor=center] at (-0.02, 0.17) {Room};
        \end{scope}
    \end{tikzpicture}
    \vspace{-2mm}
    \caption{More qualitative comparison of 3D reconstruction results}
    \label{sup_cloud}
    \vspace{-5mm}
\end{figure*}

As summarized in \cref{comparison_7scenes_nrgbd}, the evaluation results on short sequences from the 7-Scene and NRGBD datasets demonstrate that PAS3R continues to demonstrate strong competitiveness relative to current state-of-the-art methods. Furthermore, to visually demonstrate the discrepancies in point cloud reconstruction quality more intuitively, we present additional qualitative results on the 7Scene dataset with an image sequence length of 400 and the NRGBD dataset with short sequence, as illustrated in \cref{sup_cloud}. The experimental results show that PAS3R not only has leading performance for long sequences but is also highly competitive in short sequence point cloud reconstruction.

\subsection{GPU Usage And FPS Estimation} \label{GPU Usage And FPS Estimation}

\begin{figure}[tb]
    \captionsetup[subfigure]{justification=centering}
    \centering
    \begin{subfigure}[b]{0.32\textwidth}
        \centering
        \includegraphics[width=\textwidth]{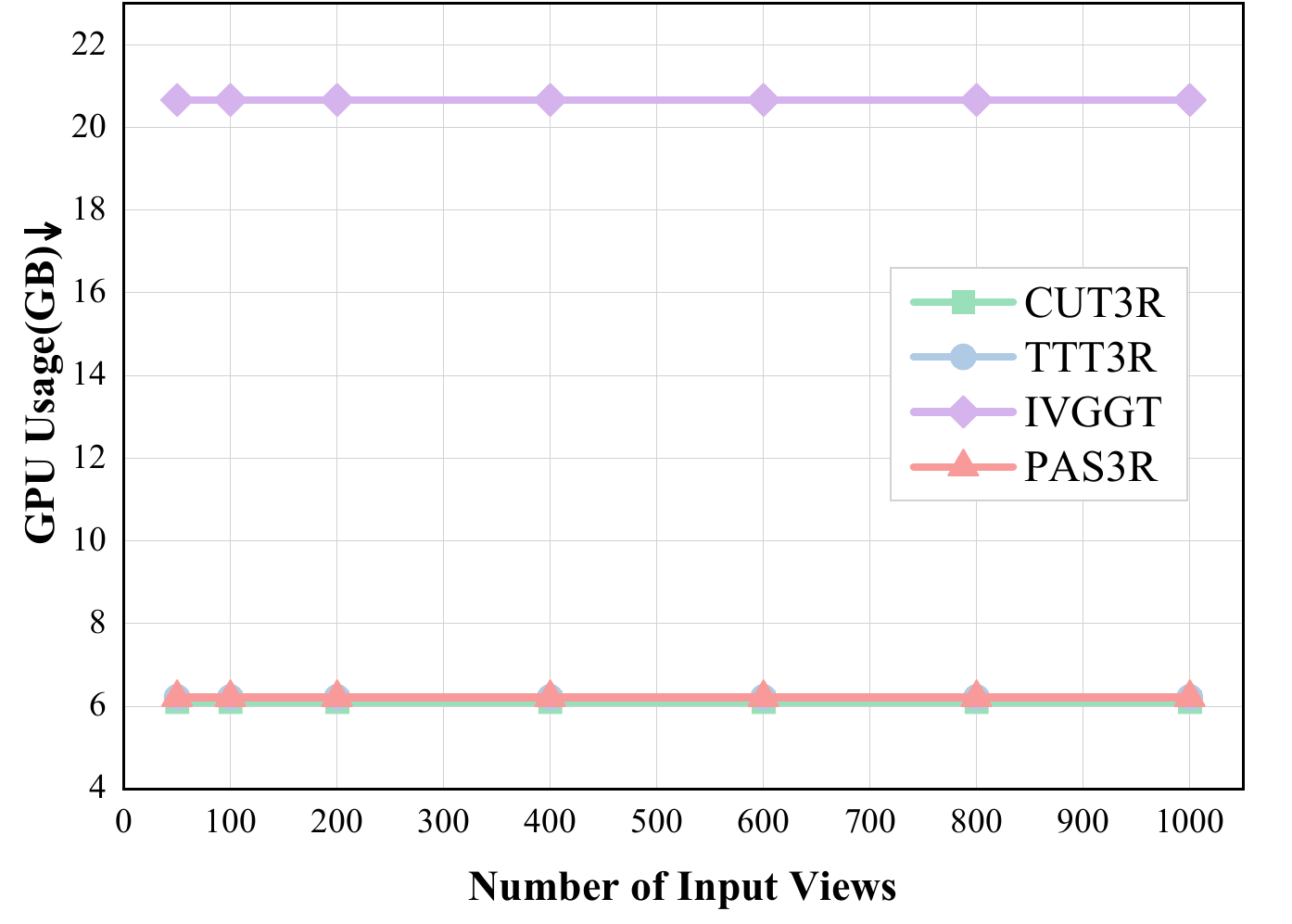}
        \caption{Comparison of GPU usage}
        \label{fg-a}
    \end{subfigure}
    \hspace{2pt}
    \begin{subfigure}[b]{0.32\textwidth}
        \centering
        \includegraphics[width=\textwidth]{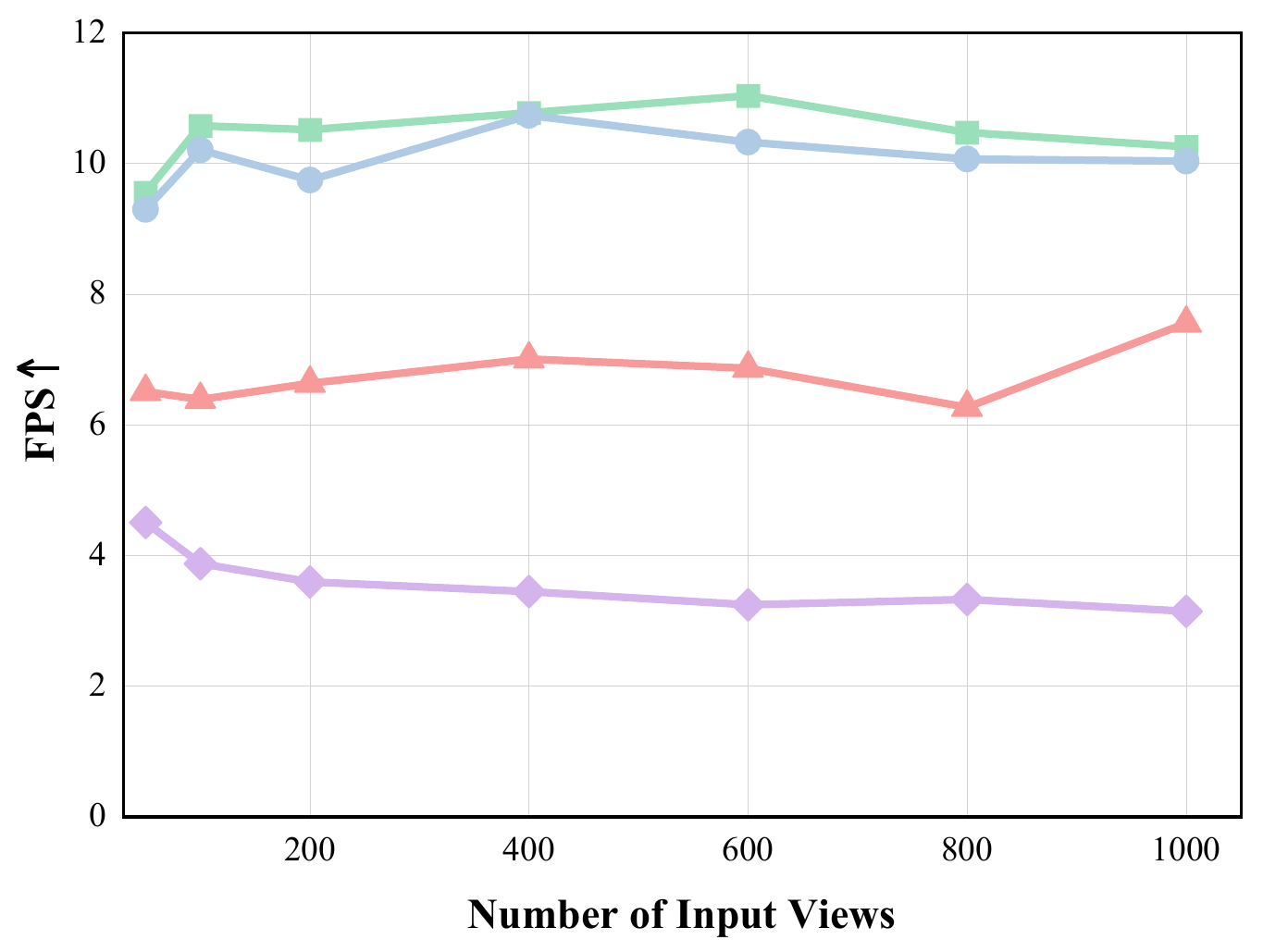}
        \caption{Comparison of FPS}
        \label{fg-b}
    \end{subfigure}
    \caption{Comparison of GPU usage and FPS on Scannet \cite{dai2017scannet}}
    \label{fg}
    \vspace{-5mm}
\end{figure}

We further evaluate the GPU peak memory usage and frames per second (FPS) of our model on the ScanNet dataset across sequence lengths ranging from 50 to 1000 frames, and compare these metrics with state-of-the-art online reconstruction methods, as depicted in \cref{fg}. The results indicate that PAS3R maintains stable runtime performance while preserving constant memory consumption as the sequence length increases. These results confirm that the proposed framework satisfies the efficiency requirements for online 3D reconstruction and scales well to long video sequences.

\end{document}